\RequirePackage{fix-cm}
\documentclass[smallextended]{svjour3}       % onecolumn (second format)
\smartqed  % flush right qed marks, e.g. at end of proof
\usepackage{graphicx}
\usepackage{lscape}
\usepackage{multirow}
\usepackage{verbatim}
\usepackage{pifont}
\usepackage[table]{xcolor}
\usepackage{tikz}
\usepackage{amssymb}
\usepackage{arydshln}
\usepackage{float}
\usepackage{hyperref}
\usepackage{makecell}
\usepackage[normalem]{ulem}
\usepackage{amsmath}
\usepackage{tablefootnote}
\usepackage{bm}

\usepackage{pifont}% http://ctan.org/pkg/pifont
\usetikzlibrary{plotmarks}

\begin{document}

\title{Sentiment analysis in tweets: an assessment study from classical to modern text representation models
}


\author{Sérgio Barreto \and Ricardo Moura \and Jonnathan Carvalho \and Aline Paes \and Alexandre Plastino}


\institute{Sérgio Barreto \at
             Universidade Federal Fluminense\\
             \email{sergiobarreto@id.uff.br} 
\and
Ricardo Moura \at
             Universidade Federal Fluminense\\
             \email{mouraricardo@id.uff.br} 
\and
            Jonnathan Carvalho \at
              Instituto Federal Fluminense \\
              \email{joncarv@iff.edu.br}           %  \\
\and
           Aline Paes \at
             Universidade Federal Fluminense\\
             \email{alinepaes@ic.uff.br} 
           \and
           Alexandre Plastino \at
             Universidade Federal Fluminense\\
             \email{plastino@ic.uff.br} 
}

\date{Received: date / Accepted: date}

\maketitle

\begin{abstract}
With the growth of social medias, such as Twitter, plenty of user-generated data emerge daily. The short texts published on Twitter -- the tweets -- have earned significant attention as a rich source of information to guide many decision-making processes. However, their inherent characteristics, such as the informal, and noisy linguistic style, remain challenging to many natural language processing (NLP) tasks, including sentiment analysis. Sentiment classification is tackled mainly by machine learning-based classifiers. The literature has adopted word representations from distinct natures to transform tweets to vector-based inputs to feed sentiment classifiers. The representations come from simple count-based methods, such as bag-of-words, to more sophisticated ones, such as BERTweet, built upon the trendy BERT architecture. Nevertheless, most studies mainly focus on evaluating those models using only a small number of datasets. Despite the progress made in recent years in language modelling, there is still a gap regarding a robust evaluation of induced embeddings applied to sentiment analysis on tweets. Furthermore, while fine-tuning the model from downstream tasks is prominent nowadays, less attention has been given to adjustments based on the specific linguistic style of the data. In this context, this study fulfils an assessment of existing language models in distinguishing the sentiment expressed in tweets by using a rich collection of 22 datasets from distinct domains and five classification algorithms. The evaluation includes static and contextualized representations. Contexts are assembled from Transformer-based autoencoder models that are also fine-tuned based on the masked language model task, using a plethora of strategies.

\end{abstract}

\section{Introduction}
\label{sec:intro}

In recent years, the use of social media networks, such as Twitter\footnote{\url{http://www.twitter.com}}, has been growing exponentially. It is estimated that about 500 million tweets -- the short informal messages sent by Twitter users -- are published daily\footnote{\url{https://www.dsayce.com/social-media/tweets-day/}}. Unlike others text style, tweets have an informal linguistic style, misspelled words, the careless use of grammar, URL links, user mentions, hashtags, and more. Due to these inherent characteristics, discovering patterns from tweets represents a challenge and opportunities for machine learning and natural language processing (NLP) tasks, such as sentiment analysis. 

Sentiment analysis is the field of study that analyzes people's opinions, sentiments, appraisals, attitudes, and emotions toward entities and their attributes expressed in written text~\cite{liu2020sentiment}. Usually, one reduces the sentiment analysis task to find out the polarity classification, i.e., whether they carry a positive or negative connotation. One of the biggest challenges concerning the sentiment classification of tweets is that people often express their sentiments and opinions using a casual linguistic style, resulting in the presence of misspelling words and the careless use of grammar. Consequently, the automated analysis of tweets' content requires machines to build a deep understanding of natural text to deal effectively with its informal structure~\cite{Pathak}. However, before discovering patterns from text, it is essential to define a more fundamental step: how automatic methods can numerically represent textual content.

Vector space models (VSMs)~\cite{Salton} are one of the earliest and most common strategies adopted in text classification literature to allow for machines deal with texts and their structures. The VSM represents each document in a corpus as a point in a vector space. Points that are close together in this space are semantically similar, and points that are far apart are semantically distant~\cite{turney2010frequency}. The firsts VSM approaches are count-based methods, such as Bag-of-Words (BoW)~\cite{10.5555/1394399} and BoW with TF-IDF~\cite{10.5555/1394399}. Although VSMs have been extensively used in the literature, they cannot deal with the curse of dimensionality. More clearly, considering the inherent characteristics of tweets, a corpus of tweets may contain different spellings for each unique word leading to an extensive vocabulary, making the vector representation of those tweets very large and often sparse.

To tackle the curse of dimensionality inherent from BOW-based approaches, in the last years it has become a standard practice to learn dense vectors to represent words and texts, the \emph{embeddings}. Methods such as such as Word2Vec~\cite{w2v}, Fast-Text~\cite{fastext}, and others~\cite{agrawal-etal-2018-learning,felbo-etal-2017-using,tang-etal-2014-learning,xu-etal-2018-emo2vec} have been used with relative success to address a plethora of NLP tasks. Nevertheless, in general, the performance of such techniques are still unsatisfactory to solve sentiment analysis from tweets, taking into account the dynamic vocabulary used by Twitter users to express themselves. Specifically, in tweets, the ironic and sarcastic content expressed in a limited space, regularly out of context and informal, makes even more challenging to retrieve meaning from the words. Such attributes may degrade the performance of traditional word embeddings methods if not handled properly. In this context, contextualized word representations have recently emerged in the literature, aiming at allowing the vector representation of words to adapt to the context they appear. Contextual embedding techniques, including ELMo~\cite{peters2018deep} and Transformer-based autoencoder methods, such as BERT~\cite{devlin2018bert}, RoBERTa~\cite{liu2019roberta}, and BERTweet~\cite{nguyen2020bertweet}, capture not only complex characteristics of word usage, such as syntax and semantics, but also how the word usage vary across linguistic contexts. Those methods have achieved state-of-the-art results on various NLP tasks, including sentiment analysis~\cite{akkalyoncu-yilmaz-etal-2019-applying,chaybouti2021efficientqa,8864964,abs-1904-08398}.

Much effort in recent language modeling research is focused on scalability issues of existing word embedding methods. On this basis, inductive transfer learning strategies and pre-trained embedding models have gained important application in the literature, especially when the amount of labeled data to train a model is relatively small. With that, models obtained from the aforementioned contextual embeddings methods are rarely trained from scratch but are instead fine-tuned from models pre-trained on datasets with a huge amount of texts~\cite{howard2018universal,peters2018deep,gpt}. Pre-trained models reduce the use of computational resources and tend to increase the classification performance of several NLP tasks, sentiment analysis included.

Despite the successful achievements in developing efficient word representation methods in NLP literature, there is still a gap regarding a robust evaluation of existing language models applied to the sentiment analysis task on tweets. Most studies are mainly focused on evaluating those models for different NLP tasks using only a small number of datasets~\cite{elmo,lan2020albert,liu2019roberta,deepmit,xu-etal-2018-emo2vec}. Our main goal is to identify appropriate embeddings-based text representations for the sentiment analysis of English tweets in this study. For this purpose, we evaluate embeddings of different natures, including: i) static embeddings learned from generic texts~\cite{agrawal-etal-2018-learning,mikolov2017advances,mikolov2013distributed,pennington-etal-2014-glove}; ii) static embeddings learned from datasets of Twitter sentiment analysis~\cite{10.1016/j.eswa.2017.02.002,7817108,felbo-etal-2017-using,pennington-etal-2014-glove,tang-etal-2014-learning,xu-etal-2018-emo2vec}; iii) contextualized embeddings learned from transformer-based autoencoders with generic texts with no adjustments~\cite{devlin2018bert,liu2019roberta}; iv) contextualized embeddings learned from Transformer-based autoencoders with a dataset of tweets with no adjustments~\cite{nguyen2020bertweet}; v) contextualized embeddings fine-tuned to the tweets language; and vi) contextualized embeddings fine-tuned to the tweets \emph{sentiment} language. In all assessments, we use a representative set of twenty-two sentiment datasets~\cite{JonnathanAIR} as input to five classifiers to evaluate the predictive performance of the embeddings. To the best of our knowledge, there is no previous study that has conducted such a robust evaluation regarding language models of several flavors and a large number of datasets. In order to identify the most appropriate text embeddings, we conduct this study to answer the following four research questions.

\textit{RQ1. Which static embeddings are the most effective in the sentiment classification of tweets?} Our motivation to evaluate those models is that many state-of-the-art deep learning models can require a lot of computational power, such as memory usage and storage. Thus, running those models locally on some devices may be difficult for mass-market applications that depend on low-cost hardware. To overcome this limitation, embeddings generated by language models can be gathered by simply looking up at the embedding table to achieve a static representation of textual content. We intend to assess how these static representations work and which are the most appropriate in this context. We answer this research question by evaluating a rich set of text representations from the literature~\cite{agrawal-etal-2018-learning,10.1016/j.eswa.2017.02.002,7817108,devlin2018bert,felbo-etal-2017-using,mikolov2017advances,mikolov2013distributed,nguyen2020bertweet,pennington-etal-2014-glove,tang-etal-2014-learning,xu-etal-2018-emo2vec,zhu2015aligning}. To achieve a good overview of the static representations, we conduct an experimental evaluation in the sentiment analysis task with five different classifiers and 22 datasets.

\textit{RQ2. Considering state-of-the-art Transformer-based autoencoder models, which are the most effective in the sentiment classification of tweets?} Regarding recent advances in language modeling, Transformer-based architectures have achieved state-of-the-art performances in many NLP tasks. Specifically, BERT~\cite{devlin2018bert} is the first method that successfully uses the encoders components of the Transformer architecture~\cite{vaswani2017attention} to learn contextualized embeddings from texts. Shortly after that, RoBERTa~\cite{liu2019roberta} was introduced by Facebook as an extension of BERT that uses an optimized training methodology. Next, BERTweet~\cite{nguyen2020bertweet} was proposed as an alternative to RoBERTa for NLP tasks focusing on tweets. While RoBERTa was trained on traditional English texts, such as Wikipedia, BERTweet was trained from scratch using a massive corpus of 850M English tweets. In this context, to answer this research question, we conduct an experimental evaluation of BERT, RoBERTa, and BERTweet models in the sentiment analysis task with five different classifiers and 22 datasets to obtain a comprehensive analysis of their predictive performances. By evaluating these models we may obtain a robust overview of the Transformer-based autoencoder representations that better fit tweet's style.

\textit{RQ3. Can the fine-tuning of Transformer-based autoencoder models using a large set of English tweets improve the sentiment classification performance?} One of the benefits of pre-trained language models, such as the Transformer-based models exploited in this study, is the possibility to adjust the language model to a specific domain by applying a fine-tuning procedure. We aim at assessing whether the sentiment analysis of tweets can benefit from fine-tuning BERT, RoBERTa and BERTweet language models with a vast, generic, and unlabeled set of around 6.7M English tweets from distinct domains. To that, we fine-tuned the pre-trained language model using the intermediate masked-language model task. Besides, considering that the fine-tuning procedure can be a very data-intensive task that may demand a lot of computational power, in addition to the large corpus of 6.7M tweets, we use in the fine-tuning process nine other samples with different sizes, varying from 500 to 1.5M tweets. We conduct an experimental evaluation with all models in the sentiment analysis task with five different classifiers and 22 datasets as in the previous questions.

\textit{RQ4. Can Transformer-based autoencoder models benefit from a fine-tuning procedure with tweets from sentiment analysis datasets?} Although using unlabeled generic tweets to adjust a language model seems to be promising regarding the availability of data, we believe that the fine-tuning procedure may benefit from the sentiment information that tweets from labeled datasets contain. In this context, we aim at identifying whether fine-tuning models with positive and negative tweets can boost the sentiment classification of tweets. We perform this evaluation by assessing three distinct strategies in order to simulate three real-world situations, as follows. In the first strategy, we use a specific sentiment dataset itself as the target domain dataset to fine-tune a language model. The second strategy simulates the case where a collection of general sentiment dataset is available to fine-tune a language model. In the third and last strategy, we combine the two previous situations. In short, we put together tweets from a target dataset and from a collection of sentiment datasets in the fine-tuning procedure. Finally, we present a comparison between the predictive performances achieved by these three evaluations and the fine-tuned models evaluated in RQ3. As in the previous questions, we conduct the experiments with five different classifiers and 22 datasets.

In summation, given the large number of language models exploited in this study, our main contributions are: (i) a comparative study of a rich collection of publicly available static representations generated from distinct deep learning methods, and with different dimensions, vocabulary size, and from various kinds of corpora; (ii) an assessment of state-of-the-art contextualized language models from the literature, that is, Transformer-based autoencoder models, including BERT, RoBERTa, and BERTweet; (iii) an evaluation of distinct strategies for fine-tuning Transformer-based autoencoder language models; and (iv) a general comparison over static, Transformer-based autoencoder, and fine-tuned language models, aiming at determining the most suitable ones for detecting the sentiment expressed in tweets{\footnote{The code and detailed results from our investigation are publicly available at \url{https://github.com/MeLL-UFF/tuning_sentiment}}}.

In order to present our contributions, we organized this article as follows. Section~\ref{sec:relatedwork} presents a literature review related to the language models examined in this study. In Section~\ref{sec:Methodology}, we describe the experimental methodology we followed in the computational experiments, which are reported in Sections~\ref{sec:static-exp},~\ref{sec:context},~\ref{sec:finetuning-exp}, and~\ref{sec:finetuning-benchmark}, responding the four research question, respectively. Finally, in Section~\ref{sec:conclusions_future}, we present the conclusions and directions for future research.

\section{Literature Review}
\label{sec:relatedwork}

Sentiment analysis is an automated process used to predict people's opinions, sentiments, evaluations, appraisals, attitudes, and emotions towards entities such as products, services, organizations, individuals, issues, events, topics, and their attributes~\cite{liu2020sentiment}. Recently, sentiment analysis has been recognized as a suitcase research problem~\cite{cambria2017suitcase}, which involves solving different NLP classification sub-tasks, including sarcasm, subjectivity, and polarity detection, which is the focus of this study.

Pioneer works in the sentiment classification of tweets mainly focused on the polarity detection task, which aims at categorizing a piece of text as carrying a positive or negative connotation. For example, Go et al.~\cite{Go_Bhayani_Huang_2009} define sentiment as a personal positive or negative feeling.  There, they used unigrams as features to train different machine learning classifiers, using tweets with emoticons as training data. The unigram model, or bag-of-words (BoW), is the most basic representation in text classification problems.

Over the years, different techniques have been developed in NLP literature in an effort to make natural language easily triable by  computers. Vector Space Models (VSMs)~\cite{Salton} are one of the earliest strategies used to represent the knowledge extracted from a given corpus. Earlier approaches to build VSMs are grounded on count-based methods, such as BoW~\cite{BoW} with TF-IDF (Term Frequency-Inverse Document Frequency)~\cite{tfidf} representation, which measures how important a word is to a document, relying on its frequency of occurrence in a corpus. 

The BoW model, which assumes word order is not important, is based on the hypothesis that the frequencies of words in a document tend to indicate the relevance of the document to a query~\cite{Salton}. This hypothesis expresses the belief that a column vector in a term-document matrix captures an aspect of the meaning of the corresponding document or phrase. Precisely, Let X be a term-document matrix. Suppose the document collection contains $n$ documents and $m$ unique terms. The matrix X will then have $m$ rows (one row for each unique term in the vocabulary) and $n$ columns (one column for each document). Let w$_i$ be the $i$-th term in the vocabulary and let d$_j$ be the $j$-th document in the collection. The $i$-th row in X is the row vector x$_{i:}$ and the $j$-th column in X is the column vector x$_{:j}$. The row vector x$_{i:}$ contains $n$ elements, one element for each document, and the column vector x$_{:j}$ contains $m$ elements, one element for each term. Suppose X is a simple matrix of frequencies, then the element x$_{ij}$ in X is the frequency of the $i$-th term w$_i$ in the $j$-th document d$_j$~\cite{BoW}.

 Such a simple way of creating numeric representations from texts have motivated early studies in detecting the sentiment expressed in tweets~\cite{barbosa2010robust,Go_Bhayani_Huang_2009,pak2010twitter}. However, though widely adopted, this kind of feature representation leads to the curse of dimensionality due to the large number of uncommon words tweets contain~\cite{saif2015thesis}.

Thus, with the revival and success of neural-based learning techniques, several methods that learn dense real-valued low dimensional vectors to represent words have been proposed, such as Word2Vec~\cite{w2v}, FastText~\cite{fastext}, and GloVe~\cite{pennington-etal-2014-glove}. Word2Vec~\cite{w2v} is one of the pioneer models to become popular taking advantage from the development of neural networks over the years. Wor2Vec is actually a software package composed of two distinct implementations of language-models, both based on a feed-forward neural architecture, namely Continuous Bag-Of-Words (CBOW) and Skip-gram. The CBOW model aims at predicting a word given its surrounding context words. Conversely, the Skip-gram model predicts the words in the surrounding context given a target word. Both architectures consist of input, a hidden layer and an output layer. The input layer has the size of the vocabulary and encodes the context by combining the one-hot vector representations of surrounding words of a given target word. The output layer has the same size as the input layer and contains a one-hot vector of the target word obtained during the training. However, one of the main disadvantages of those models is that they usually struggle to deal with out-of-vocabulary (OOV) words, i.e., words that have not been seen in the training data before. To address this weakness, more complex approaches have been proposed, such as FastText~\cite{fastext}.

FastText~\cite{fastext} is based on the Skip-gram model~\cite{w2v}, still it considers each word as a bag of character n-grams, which are contiguous sequences of $n$ characters from a word, including the word itself. A dense vector is learned to each character n-gram and the dense vector associated to a word is taken from the sum of those representations. Thus, FastText can deal with different morphological structure of words that covers the words not seen in the training phase, i.e., OOV words. For that reason, FastText is also able to deal with tweets, considering the huge number of uncommon and unique words in this kind of text.

Going to another direction, the GloVe model~\cite{pennington-etal-2014-glove} attempts at making efficient use of statistics of word occurrences in a corpus to learn better word representations. In~\cite{pennington-etal-2014-glove}, Pennington et al. present a model that rely on the insight that ratios of co-occurrences, rather than raw counts, encode semantic information about pair of words. This relationship is used to derive a suitable loss function for a log-linear model, which is then trained to maximize the similarity of every word pair, as measured by the ratios of co-occurrences. Given a probe word, the ratio can be small, large or equal to one depending on their correlations. This ratio gives hints on the relations between three different words. For example, given a probe word and two others w$_i$ and w$_j$,  if the ratio is large, the probe word is related to w$_i$ but not w$_j$. 

In general, methods for learning word embeddings deal well with the the syntactic context of words but ignore the potential sentiment they carry. In the context of sentiment analysis, words with similar syntactic structure but opposite sentiment polarity, such as \textit{good} and \textit{bad}, are usually mismapped to neighbouring word vectors. To address this issue, Tang et al.~\cite{tang-etal-2014-learning} proposed the Sentiment-Specific Word Embedding model (SSWE), which encodes the sentiment information in the embeddings. Specifically, they developed neural networks that incorporate the supervision from sentiment polarity of texts in their loss function. To that, they slide the window of n-gram across a sentence, and then predict the sentiment polarity based on each n-gram with a shared neural network. In addition to SSWE, other methods have been proposed in order to improve the quality of word representations in sentiment analysis, by leveraging the sentiment information in the training phase, such as DeepMoji~\cite{felbo-etal-2017-using}, Emo2Vec~\cite{xu-etal-2018-emo2vec}, and EWE~\cite{agrawal-etal-2018-learning}.

The aforementioned word embedding models have been used as standard components in most sentiment analysis methods. However, they pre-compute the representation for each word independently from the context they are going to appear. This static nature of these models results in two problems: (i) they ignore the diversity of meaning each word may have, and (ii) they suffer from learning long-term dependencies of meaning. Different from those static word embedding techniques, \emph{contextualized embeddings} are not fixed, adapting the word representation to the context it appears. Precisely, at training time, for each word in a given input text, the learning model analyzes the context, usually using sequence-based models, such as recurrent neural networks (RNNs), and adjusts the representation of the target word by looking at the context. These context-awareness embeddings are actually the internal states of a deep neural network trained in an self-supervised setting. Thus, the training phase is carried out independently from the primary task on an extensive unlabeled data. Depending on the sequence-based model adopted, these contextualized models can be divided into two main groups, namely RNN-based~\cite{elmo} and Transformers-based~\cite{le2020flaubert,liu2019roberta,nguyen2020bertweet,lan2020albert,vaswani2017attention}.

Transfer learning strategies have also been emerging to improve the quality of word representation, such as ULMFit (Universal Language Model Fine-tuning)~\cite{howard2018universal}. ULMFit is an effective transfer learning method that can be applied to any NLP task, and introduces key techniques for fine-tuning a language model, consisting of three stages, described as follows. First, the language model is trained on a general-domain corpus to capture generic features of the language in different layers. Next, the full language model is fine-tuned on the target task data using discriminative fine-tuning and slanted triangular learning rates (STLR) to learn task-specific features. Lastly, the model is fine-tuned on the target task using gradual unfreezing and STLR to preserve low-level representations and to adapt high-level ones.

Fine-tuning techniques made possible the development and availability of pre-trained contextualized language models using massive amounts of data. For example, Peters et al.~\cite{peters2018deep} introduced ELMo (Embeddings from Language Models), a deep contextualized model for word representation. ELMo comprises a Bi-directional Long-Short-Term-Memory Recurrent Neural Network (BiLSTM) to combine a forward model, looking at the sequence in the traditional order, and a backward model, looking at the sequence in the reverse order. ELMo is composed of two layers of BiLSTM sequence encoder responsible for capturing the semantics of the context. Besides, some weights are shared between the two directions of the language modeling unit and there is also a residual connection between the LSTM layers to accommodate the deep connections without the gradient vanishing issue. ELMo also makes use of the character-based technique for computing embeddings. Therefore, it benefits from the characteristics of character-based representations to avoid OOV words. 

Although ELMo is more effective as compared to static pre-trained models, its performance may be degraded when dealing with long texts, exposing a trade-off between efficient learning by gradient descent and latching on information for long periods~\cite{Bengio_1994}. Transformers-based language models, on the other hand, have been proposed to solve the gradient propagation problems described in~\cite{Bengio_1994}. Compared to RNNs, which process the input sequentially, Transformers work in parallel, which brings benefits when dealing with large corpora. Moreover, while RNNs by default process the input in one direction, Transformers-based models can attend to the context of a word from distant parts of a sentence and pay attention to the part of the text that really matters, using self-attention~\cite{vaswani2017attention}.

The OpenAI Generative Pre-Training Transformer model (GPT)~\cite{gpt} is one of the first attempts to learn representations using Transformers. It encompasses only the decoder component of the Transformer architecture with some adjustments, discarding the encoder part. Therefore, instead of having a source and a target sentence for the sequence transduction model, a single sentence is given to the decoder. GPT' objective function targets at predicting the next word given a sequence of words, as a standard language modeling goal. To comply with the standard language model task, while reading a token, GPT can only attend to previously seen tokens in the self-attention layers. This setting can be limiting for encoding sentences, since understanding a word might require processing the ones coming after it in the sentence. 

Devlin et al.~\cite{devlin2018bert} addressed the unidirectional nature of GPTs by presenting an strategy called BERT (Bidirectional Encoder Representations from Transformers) that, as the name says, encodes sentences by looking them at both directions. BERT is also based on the Transformer architecture but, contrary to the GPT, it is based on the encoder component of that architecture. The essential improvement over GPT is that BERT provides a solution for making Transformers bidirectional by applying masked language models, which randomly masks some percentage of the input tokens, and the objective is to predict those masked tokens based on their context. Also, in~\cite{devlin2018bert}, they use a next sentence prediction task for predicting whether two text segments follow each other. All those improvements have made BERT to achieve state-of-the-art results in various NLP tasks when it was published.

Later, Liu et al.~\cite{liu2019roberta} proposed RoBERTa (Robustly optimized BERT approach), achieving even better results than BERT. RoBERTa is an extension of BERT with some modifications, such as: (i) training the model for a longer period of time, with bigger batches, over more data, (ii) removing the next sentence prediction objective, (iii) training on longer sequences, and (iv) dynamically changing the masking pattern applied to the training data.

Recently, Nguyen et al.~\cite{nguyen2020bertweet} introduced BERTweet, an extension of RoBERTa trained from scratch with tweets. BERTweet has also the same architecture as BERT, but it is trained using the same Roberta pre-training procedure instead. BERTweet consumes a corpus of 850M English tweets, which is a concatenation of two corpora. The first corpus contains 845M English tweets from the Twitter Stream dataset and the second one contains 5M English tweets related to the COVID-19 pandemic. In~\cite{nguyen2020bertweet}, the proposed BERTweet model outperformed RoBERTa baselines in some tasks on tweets, including sentiment analysis.

As far as we know, most studies in language modeling focus on designing new effective models in order to improve the predictive performance of distinct NLP tasks. For example, Devlin et al.~\cite{devlin2018bert} and Liu et al.~\cite{liu2019roberta} have respectively introduced BERT and RoBERTa, which achieved state-of-the-art results in many NLP tasks. Nevertheless, they did not evaluate the performance of such methods on the sentiment classification of tweets. Nguyen et al.~\cite{nguyen2020bertweet}, on the other hand, used only a unique generic collection of tweets when evaluating their BERTweet strategy. In this context, we fulfill a robust evaluation of existing language models from distinct natures, including static representations, Transformer-based autoencoder models, and fine-tuned models, by using a significant set of 22 datasets of tweets from different domains and sizes. In the following sections, we present the assessment of such models.

\section{Experimental Methodology}

\label{sec:Methodology}

This section presents the experimental methodology we followed in this article. We begin by describing, in Section \ref{sec:mth-dataset}, the twenty-two benchmark datasets used to evaluate the different language models we investigate in this study. In Section~\ref{sec:classf}, we present the experimental protocol we followed. Then, in Section~\ref{sec:comp-exp}, we describe the computational experiments reported in Sections~\ref{sec:static-exp},~\ref{sec:context},~\ref{sec:finetuning-exp}, and~\ref{sec:finetuning-benchmark}.

\subsection{Datasets}
\label{sec:mth-dataset}

We used a large set of twenty-two datasets \cite{jonnathan} to assess the effectiveness of the distinct language models described in Section \ref{sec:relatedwork}{\footnote{The datasets are publicly available at \url{https://github.com/joncarv/air-datasets}}}. Table~\ref{tab:datasets-info} summarizes the main characteristics of these datasets, namely the abbreviation we use when reporting the experimental results to save space (\textit{Abbrev.} column), the domain they belong (\textit{Domain} column), number of positive tweets (\#\textit{pos.} column), proportion of positive tweets (\%\textit{pos.} column), number of negative tweets (\#\textit{neg.} column), proportion of negative tweets (\%\textit{neg.} column), and the total number of tweets (\textit{Total} column).

Those datasets have been extensively used in the literature of Twitter sentiment analysis and we believe they provide a diverse scenario in evaluating embeddings of tweets in the sentiment classification task, regarding a variety of domains, sizes, and class balance. For example, while datasets SemEval13, SemEval16, SemEval17, and SemEval18 contain generic tweets, other datasets, such as iphone6, movie, and archeage, contain tweets of a particular domain. Also, the datasets vary a lot in size, with some of them containing only dozens of tweets, such as irony and sarcasm. We believe that this diverse and large collection of datasets may help drawing more concise and robust conclusions on the effectiveness of distinct language models in the sentiment analysis task.

\begin{table}%[H]
\caption{Characteristics of the Twitter sentiment datasets ordered by size (Total column)}
\label{tab:datasets-info}
\scalebox{0.76}{
\begin{tabular}{llllllll}
\hline\noalign{\smallskip}
\textbf{Dataset} & \textbf{Abbrev.}& \textbf{Domain}& \textbf{\#pos.} & \textbf{\%pos.} & \textbf{\#neg.} & \textbf{\%neg.}& \textbf{Total} \\
\noalign{\smallskip}\hline\noalign{\smallskip}
irony~\cite{brasnam} & iro&Irony& 22 & 34\% & 43&66\%&65  \\
sarcasm~\cite{brasnam} & sar & Sarcasm&33 & 46\% & 38& 54\%&71\\
aisopos\tablefootnote{\url{http://www.grid.ece.ntua.gr}} &ntu& Generic&159 & 57\% & 119&43\%&278\\
SemEval-Fig\tablefootnote{\url{http://www.alt.qcri.org/semeval2015/task11}}& S15 &Irony$/$Metaphors& 47 & 15\% & 274&85\%&321 \\
sentiment140~\cite{Go_Bhayani_Huang_2009} & stm&Generic& 182 & 51\% & 177&49\%&359\\
person~\cite{Chen2012ExtractingDS}& per&Towards a Person & 312 & 71\% & 127&29\%&439\\
hobbit~\cite{Lochter2016ShortTO} & hob&Movies& 354 & 68\% & 168 & 32\%&522\\
iphone6~\cite{Lochter2016ShortTO} & iph&Products&371 & 70\% & 161& 30\%&532 \\
movie~\cite{Chen2012ExtractingDS} & mov& Movies&460 & 82\% & 101&18\%&561 \\
sanders\tablefootnote{\url{https://www.github.com/karanluthra/twitter-sentiment-training}} & san&Business & 570 & 47\% & 654&53\%&1,224\\
Narr~\cite{Narr2012LanguageIndependentTS} & Nar&Generic&1,739 & 60\% & 488&40\%&1,227\\
archeage~\cite{Lochter2016ShortTO} & arc& Games& 724 & 42\% & 994&58\%&1,718\\
SemEval18~\cite{SemEval2018Task1} & S18&Equity Evaluation Corpus&865 & 47\% & 994&53\%&1,859\\
OMD~\cite{10.1145/1753326.1753504} & OMD &Presidential Debate& 710 & 37\% & 1,196 & 63\% & 1,906 \\
HCR~\cite{10.5555/2140458.2140465} & HCR & Health Care Reform&539 & 28\% & 1,369 & 72\% & 1,908\\
STS-gold~\cite{saif2013evaluation} & STS& Generic &632 & 31\% & 1,402 & 69\% & 2,034\\
SentiStrength~\cite{journals/jasis/ThelwallBP12} & SSt&Generic&1,340 & 59\% & 949 & 41\% & 2,289\\
Target-dependent~\cite{dong2014adaptive} & Tar& Celebrities& 1,734 & 50\% & 1,733 & 50\% & 3,467\\
Vader~\cite{gilbert2014vader} & vad & Generic&2,897 & 69\% & 1,299 & 31\% & 4,196\\
SemEval13\tablefootnote{\url{https://www.cs.york.ac.uk/semeval-2013/task2.html}} & S13 &Generic& 3,183 & 73\% & 1,195 & 23\% & 4,378\\
SemEval17~\cite{rosenthal-etal-2017-semeval} &S17& Generic&2,375 & 37\% & 3,972 & 63\% & 6,347\\
SemEval16~\cite{nakov-etal-2016-semeval} &S16&Generic& 8,893 & 73\% & 3,323 & 27\% & 12,216\\
\noalign{\smallskip}\hline
\end{tabular}
}
\end{table}

\subsection{Experimental protocol}
\label{sec:classf}

To assess the effect of different kinds of language models in the polarity classification task, we follow the protocol of first extracting the features from the several vector-based language representation mechanisms (BOW, static embeddings, contextualized embeddings, fine-tuned embeddings). Next, those features compose the input attribute space for five distinct classifiers, namely Support Vector Machine (SVM), Logistic Regression (LR), Random Forest (RF), XGBoost (XBG), and Multi-layer Perceptron (MLP). We adopted scikit-learn's\footnote{\url{https://scikit-learn.org}} implementations of those machine learning algorithms. Although we have used the default parameters in most of the cases, it is important to mention that we set the class balance parameter for SVM, LR, and RF (\textit{class}\_\textit{weight} = \textit{balanced}). Also, for LR, we set the maximum number of iterations to 500 (\textit{max}\_\textit{iter} = $500$) and the solver parameter to \textit{liblinear}. Moreover, for MLP, we set the number of hidden layers to 100. Table \ref{tab:class-set} shows a summary of the classification algorithms used in this study, remarking their characteristics. We aim at determining which language models are the most effective ones in Twitter sentiment analysis by leveraging classifiers from distinct natures, thus examining how they deal with the peculiarities of each evaluated model. Furthermore, it is important to note that we do not aim at establishing the best classifier for the sentiment analysis task, which may require a specific study and additional computational experiments.

\begin{table}%[H]
\centering
\caption{Summary of the advantages and disadvantages of the classification algorithms adopted in this study}
\label{tab:class-set}
\scalebox{0.84}{
\begin{tabular}{lll} 
\hline\noalign{\smallskip}
\textbf{Classifier} & \textbf{Advantages}& \textbf{Disadvantages}\\ 
\hline\noalign{\smallskip}
\multirow{4}{*}{SVM} &-- Works well when there is a clear&-- Underperforms when the number of\\ 
&margin of separation between classes& features for each data point exceeds\\
&-- More effective in high dimensional spaces& the number  of training data samples\\
&-- Memory efficient&\\

\rule{0pt}{0.5ex}\\

\multirow{2}{*}{LR} & -- Easy to interpretate & -- Non-linear problems cannot be solved\\
&-- Provides association directions&-- Needs that independent variables are i.i.d\\
&&-- Linearly related to the log odds \\

\rule{0pt}{0.5ex}\\

\multirow{2}{*}{RF} &-- No feature scaling required &-- Long training period\\
&-- Handles non-linear parameters efficiently&-- Difficult to interpret\\

\rule{0pt}{0.5ex}\\

\multirow{2}{*}{XGB} &-- Highly scalable/parallelizable& -- More likely to overfit\\ 
&-- Quick to execute&-- Many parameters\\
%&Quick to execute&\\
\rule{0pt}{0.5ex}\\

\multirow{3}{*}{MLP} &-- Learns non-linear and complex& -- Difficult to interpret parameters\\
&relationships&-- Convergence of the weights can\\
&&be very slow\\

%\rule{0pt}{0.5ex}\\
\noalign{\smallskip}\hline
\end{tabular}
}
\end{table}

Preprocessing is the first step in many text classification problems and the use of appropriate techniques can reduce noise hence improving classification effectiveness~\cite{Fayyad}. As this manuscript's main goal is to evaluate the performance of different models of tweet representation, the preprocessing step is simple so that the focus is on the language models and classifiers. Thus, for each tweet in a given dataset, we only replace URLs by the token \textit{someurl}, user mentions by the token \textit{someuser}, and all tokens were lowercased.

In the experimental evaluation, the predictive performance of the sentiment classification is measured in terms of accuracy and $F_1$-macro. Precisely, for each evaluated dataset, the accuracy of the classification was computed as the ratio between the number of correctly classified tweets and the total number of tweets, following a stratified ten-fold cross-validation.~$F_1$-macro was computed as the unweighted average of the $F_1$-score for the positive and negative classes.

Moreover, all experiments were performed by using Tesla P100-SXM2 GPU within Ubuntu operational system, running in a machine with Intel(R) Xeon(R) CPU E5-2698 v4 processor.

\subsection{Computational experiments details}
\label{sec:comp-exp}

In the next sections, we evaluate a significant collection of vector-based textual representations attempting to answer the research questions introduced in Section~\ref{sec:intro}. Specifically, we conduct a comparative study of vector-based language representation models from distinct natures, including Bag of Words, as a classic baseline, static representations and representations induced from Transformer-based autoencoder models, by fine-tuning or not the intermediate masked language task, in order to acknowledge their effectiveness in the polarity classification of English tweets. These language representation models are incrementally evaluated throughout Sections~\ref{sec:static-exp},~\ref{sec:context},~\ref{sec:finetuning-exp}, and~\ref{sec:finetuning-benchmark}.

In Section~\ref{sec:static-exp}, we begin by analyzing the predictive performance of the static representations, which include 13 pretrained embeddings from the literature, as shown in Table~\ref{tab:staticemb}, as well as the classical BOW with TF-IDF representation schema. Regarding the static embeddings described in Table~\ref{tab:staticemb}, we have selected representations trained on distinct kinds of texts (Corpus column) and built from different architectures (Architecture column), from feedforward neural networks to Transformer-based ones. The $|D|$ and $|V|$ columns refer to the dimension and vocabulary size of each pretrained embedding, respectively. Although the most usual way of employing embeddings trained from Transformer-based architectures is running the text trough the model to obtain contextualized representations, here we first investigate how these models behave when the experimental protocol is the same as earlier embeddings models: pretrained embeddings are collected from the embeddings layer and are the input of the classifiers.  

\begin{table}%[!htbp]
\caption{Characteristics of the static pretrained embeddings ordered by the number of dimensions}
\label{tab:staticemb}
\scalebox{0.78}{
\begin{tabular}{lrrlll}
\hline\noalign{\smallskip}
\textbf{Embedding} & \textbf{$|$\textit{D}$|$} & \textbf{$|$\textit{V}$|$} & \textbf{Architecture}&\textbf{Corpus}  \\
\noalign{\smallskip}\hline\noalign{\smallskip}
SSWE~\cite{tang-etal-2014-learning} & 50 & 137K &Feed-forward network &Twitter (10M tweets)  \\
Emo2Vec~\cite{xu-etal-2018-emo2vec} & 100 & 1.2M &Convolutional network& Twitter (1.9M tweets)  \\ %1,193,516
GloVe-TWT~\cite{pennington-etal-2014-glove} & 200 & 1.2M &log-bilinear model& Twitter (27B tokens) \\ %1,193,514
DeepMoji~\cite{felbo-etal-2017-using} & 256 & 50K &Recurrent network& Twitter (1B tweets)  \\
EWE~\cite{agrawal-etal-2018-learning} & 300 & 183K &Recurrent network& Amazon reviews (200K reviews)  \\

GloVe-WP~\cite{pennington-etal-2014-glove} & 300 & 400K &log-bilinear model& Wikipedia/Gigaword (6B tokens) \\
fastText~\cite{mikolov2017advances} & 300 & 1M &Feed-forward network& Wikipedia/web pages/news (16B tokens) \\
w2v-GN~\cite{mikolov2013distributed} & 300 & 3M &Feed-forward network& Google news (100B tokens)  \\
w2v-Edin~\cite{7817108} & 400 & 259K & Feed-forward network&Twitter (10M tweets)  \\
w2v-Araque~\cite{10.1016/j.eswa.2017.02.002} & 500 & 57K & Feed-forward network&Twitter (1.28M tweets)  \\
BERT~\cite{devlin2018bert} & 768 & 30K&Transformers&BooksCorpus~\cite{zhu2015aligning}/Eng. Wiki (3.3B words)\\
RoBERTa (RoB)~\cite{zhu2015aligning} & 768 & 50K &Transformers& 5 datasets~\cite{liu2019roberta} (161GB)\\
BERTweet (BTWT)~\cite{nguyen2020bertweet} & 768 & 64K &Transformers& Twitter (850M tweets)\\
\noalign{\smallskip}\hline
\end{tabular}
}
\end{table}

Next, in Section~\ref{sec:context}, we present an evaluation of state-of-the-art Transformer-based autoencoder models, including BERT~\cite{devlin2018bert}, RoBERTa~\cite{liu2019roberta}, and BERTweet \cite{nguyen2020bertweet}. In this evaluation, for each assessed dataset, we represent their tweets as the average of the concatenation of the last four layers for each token representation of the models. For the sake of simplicity, the Transformer-based autoencoder models assessed in this study are referred to hereafter as Transformer-based models.

Lastly, in Sections~\ref{sec:finetuning-exp} and~\ref{sec:finetuning-benchmark}, we evaluate the effectiveness of fine-tuning the aforementioned Transformer-based models regarding the intermediate masked-language task in two different ways: (i) by using a huge collection of unlabeled, or non-sentiment, tweets, and (ii) by using tweets from sentiment datasets.

In Section~\ref{sec:finetuning-exp}, regarding the non-sentiment fine-tuning approach, we adopted the general purpose collection of unlabeled tweets from the Edinburgh corpus~\cite{petrovic-etal-2010-edinburgh}, which contains 97M tweets in multiple languages. Tweets written in languages other than English were discarded, resulting in a final corpus of 6.7M English tweets, which was then used to fine-tune BERT, RoBERTa, and BERTweet. In addition to the entire corpus of 6.7M tweets, we used nine other samples with different sizes, varying from 500 to 1.5M tweets. Specifically, we generated samples containing 500 (0.5K), 1K, 5K, 10K, 25K, 50K, 250K, 500K, and 1.5M non-sentiment tweets.

Conversely, in Section~\ref{sec:finetuning-benchmark}, we evaluated the sentiment fine-tuning procedure using positive and negative tweets from the twenty-two benchmark datasets described in Table~\ref{tab:datasets-info}. For this purpose, we used each dataset once as the target dataset, while the others were used as the source datasets. More clearly, for each assessed dataset, referred to as the target dataset, we explored three distinct strategies to fine-tune the masked-language model: (i) by using only the tweets from the target sentiment dataset itself, (ii) by using the tweets from the remaining 21 datasets, and (iii) by using the entire collection of tweets from the 22 datasets, including the tweets from the target dataset.

\section{Evaluation of static text representations}
\label{sec:static-exp}

The computational experiments conducted in this section aim at answering the research question RQ1, as follows:

\textit{RQ1. Which static embeddings are the most effective in the sentiment classification of tweets?}

We answer this question by assessing the predictive power of the 13 pretrained embeddings described in Table~\ref{tab:staticemb}. These embeddings were generated from distinct neural networks architectures, with different dimensions and vocabulary size, and trained on various kinds of corpora. Recall that by static embeddings we mean that the features are gathered from the embeddings layer working as a look-up table of tokens. In addition to the pretrained embeddings, we evaluate the BoW model with the TF-IDF representation, which is the most basic text representation used in Twitter sentiment analysis and text classification tasks in general. For all tweet representation, we take the average of all tokens representation of the tweet.

We begin by evaluating the predictive performance of the static representations for each classification algorithm presented in Table~\ref{tab:class-set}. We report the computational results in detail for SVM as an example of this evaluation (refer to Online Resource 1 for the detailed evaluation for each classifier). Tables~\ref{tab:static-svm-results-acc} and~\ref{tab:static-svm-results-f1} show the results achieved by using each static representation to train an SVM classifier, in terms of classification accuracy and unweighted $F_1$-macro, respectively. The boldfaced values indicate the best results, and the last three lines show the total number of wins for each static representation (\#wins row), as well as a ranking of the results (rank sums and position rows). Precisely, for each dataset, we assign scores, from 1.0 to 14.0, to each assessed representation (each column), in ascending order of accuracy ($F_1$-macro), where the score 1.0 is assigned to the representation with the highest accuracy ($F_1$-macro). Thus, low score values indicate better results. When two assessed representation has the same performance, we take an average of their scores. If two assessed representations achieve the best performance, they will receive a score of 1.5 ((1+2)/2). Finally, we sum up the assigned scores obtained in each dataset for each assessed representation to calculate rank sums. With the rank sum of each assessed representation, we rank the rank-sum result from the best (1) to the worst (14), calculating the rank position.

%%%%%%%%%%%%%%%%%%%%%%%%%%%%%%%%%%%%%%%%%%%%%%%%%%%%%%%%%%%%%%%%%%%%%%%%%%%%%%%%%%%%
\begin{table}%[!htbp]
{\caption{Accuracies (\%) achieved using the SVM classifier}
\label{tab:static-svm-results-acc}
\scalebox{0.59}{
\begin{tabular}{llllllllllllllll}
\hline\noalign{\smallskip}
\multirow{3}{*}{\textbf{Dataset}} & \textbf{w2v} & \textbf{GloVe} & \textbf{Fast} & \textbf{EWE} & \textbf{GloVe} & \textbf{BERT} & \textbf{TF-} & \textbf{w2v} & \textbf{w2v} & \textbf{SSWE} & \textbf{Emo2} & \textbf{Deep} & \textbf{RoB}& \textbf{BTWT}\\
& \textbf{GN} & \textbf{WP} & \textbf{Text} & \textbf{} & \textbf{TWT} & \textbf{static} & \textbf{IDF} & \textbf{Araque} & \textbf{Edin} & \textbf{} & \textbf{Vec} & \textbf{Moji} & \textbf{static}& \textbf{static}\\
\noalign{\smallskip}\hline\noalign{\smallskip}
iro&70.95&69.52&67.86&52.14&51.67&60.24&64.52&63.10&66.43&70.48&\textbf{78.33}&61.90&70.48&39.29\\
sar&68.75&68.75&61.61&68.75&67.32&70.54&54.64&71.79&68.93&\textbf{87.32}&73.04&67.50&56.07&57.68\\
ntu&79.15&88.47&73.04&70.54&73.70&83.47&74.40&77.71&93.15&\textbf{93.53}&81.28&91.73&89.91&81.98\\
S15&87.88&83.20&81.95&88.80&83.50&89.72&87.23&87.54&89.41&77.56&82.88&88.16&\textbf{90.34}&87.52\\
stm&85.22&83.02&83.56&83.01&83.56&81.06&81.59&82.71&\textbf{87.74}&80.75&85.24&81.34&84.94&71.59\\
per&\textbf{83.37}&76.75&76.77&74.69&73.56&73.32&80.18&74.93&79.03&72.44&76.09&76.98&80.40&69.23\\
hob&89.66&85.24&83.51&90.22&79.88&91.75&\textbf{93.29}&91.37&90.04&76.99&87.18&91.57&92.90&90.03\\
iph&78.00&75.57&76.13&72.17&73.88&79.33&80.07&80.46&78.01&76.49&\textbf{81.57}&78.56&80.08&76.51\\
mov&82.90&76.66&76.65&75.95&71.67&83.96&84.67&83.78&84.14&78.09&83.07&81.29&\textbf{85.38}&79.35\\
san&82.59&78.50&80.55&80.54&80.38&81.37&83.33&81.46&83.25&80.14&81.12&80.80&\textbf{83.57}&78.84\\
Nar&84.51&83.94&84.03&83.78&85.33&83.46&80.36&84.18&88.67&\textbf{88.84}&88.34&88.83&87.21&80.61\\
arc&85.45&83.59&85.16&84.05&85.27&86.61&\textbf{87.54}&85.16&86.67&79.63&83.35&86.09&87.43&84.87\\
S18&80.37&78.48&80.69&80.26&80.20&84.56&82.09&75.79&82.36&81.28&80.96&78.97&\textbf{86.50}&79.51\\
OMD&83.79&82.21&81.42&77.55&77.38&84.15&79.22&77.70&82.84&76.50&75.76&77.02&\textbf{85.10}&82.37\\
HCR&74.78&72.58&72.64&71.59&73.21&76.36&\textbf{80.24}&73.95&75.94&67.29&70.54&73.21&78.30&73.27\\
STS&85.94&83.19&85.74&84.71&84.90&86.97&83.97&86.92&87.02&\textbf{88.99}&86.19&88.69&87.86&83.09\\
SSt&77.98&75.10&76.98&77.76&77.24&78.51&73.48&76.28&79.56&79.77&\textbf{84.93}&79.64&80.21&73.00\\
Tar&83.67&81.92&83.82&83.18&82.81&83.44&82.35&81.40&83.39&79.18&82.98&82.90&\textbf{84.42}&80.36\\
Vad&88.25&84.51&87.01&86.44&85.96&87.51&83.27&85.30&87.73&85.70&85.94&86.77&\textbf{89.32}&81.82\\
S13&81.52&78.53&79.49&78.39&78.69&81.59&80.52&79.19&80.31&81.04&\textbf{87.80}&81.89&83.60&77.64\\
S17&88.61&86.34&88.29&87.22&87.71&88.47&87.95&84.53&87.96&81.63&85.60&85.47&\textbf{89.03}&86.01\\
S16&84.50&82.51&84.50&83.15&83.87&85.27&85.85&81.37&84.39&78.51&82.29&82.79&\textbf{86.00}&81.69\\
\noalign{\smallskip}\hline
\#wins &1& 0& 0& 0& 0&0&3&0& 1& 4& 4& 0&\textbf{9}&0\\
rank sums &116.5&221.0&183.5&212.0&216.0&117.0&155.0&185.5&95.0&200.5&154.0&152.5&\textbf{56.5}&245.0\\
position &3.0&13.0&8.0&11.0&12.0&4.0&7.0&9.0&2.0&10.0&6.0&5.0&\textbf{1.0}&14.0\\
\noalign{\smallskip}\hline\noalign{\smallskip}
\end{tabular}
}}
\end{table}

%%%%%%%%%%%%%%%%%%%%%%%%%%%%%%%%%%%%%%%%%%%%%%%%%%%%%%%%%%%%%%%%%%%%%%%%%%%%%%%%%%%%

%%%%%%%%%%%%%%%%%%%%%%%%%%%%% F1 macro %%%%%%%%%%%%%%%%%%%%%%%%%%%%%%%%%%%%%
\begin{table}%[!htbp]
{\caption{$F_1$-macro scores (\%) achieved by evaluating static representation using the SVM classifier}
\label{tab:static-svm-results-f1}
\scalebox{0.59}{
\begin{tabular}{llllllllllllllll}
\hline\noalign{\smallskip}
\multirow{3}{*}{\textbf{Dataset}} & \textbf{w2v} & \textbf{GloVe} & \textbf{Fast} & \textbf{EWE} & \textbf{GloVe} & \textbf{BERT} & \textbf{TF-} & \textbf{w2v} & \textbf{w2v} & \textbf{SSWE} & \textbf{Emo2} & \textbf{Deep} & \textbf{RoB}& \textbf{BTWT}\\
& \textbf{GN} & \textbf{WP} & \textbf{Text} & \textbf{} & \textbf{TWT} & \textbf{static} & \textbf{IDF} & \textbf{Araque} & \textbf{Edin} & \textbf{} & \textbf{Vec} & \textbf{Moji} & \textbf{static}& \textbf{static}\\
\noalign{\smallskip}\hline\noalign{\smallskip}
iro&60.03&64.46&60.78&47.51&49.40&54.84&39.11&48.35&58.24&67.04&\textbf{72.17}&53.46&61.16&35.48\\
sar&66.00&67.75&60.22&67.55&63.32&69.29&51.68&69.97&66.70&\textbf{86.93}&70.72&64.72&50.28&52.69\\
ntu&78.79&87.81&72.46&70.19&73.02&83.15&71.55&76.72&92.99&\textbf{93.29}&80.99&91.53&89.42&81.22\\
S15&69.91&65.74&68.13&75.02&70.93&77.04&58.51&66.33&77.62&67.16&70.24&73.65&\textbf{78.48}&75.38\\
stm&85.18&82.94&83.46&82.96&83.51&80.96&81.46&82.60&\textbf{87.67}&80.71&85.18&81.28&84.87&71.50\\
per&\textbf{80.58}&74.31&74.86&73.03&71.80&70.85&72.27&71.59&76.97&70.46&74.16&74.46&77.30&67.53\\
hob&88.56&83.89&82.25&89.26&78.71&90.92&91.73&90.52&88.95&75.49&86.03&90.73&\textbf{92.08}&89.05\\
iph&75.96&74.31&74.96&71.00&72.63&77.34&71.55&77.92&76.18&74.93&\textbf{79.97}&76.62&78.00&74.70\\
mov&73.42&68.28&68.39&67.23&64.17&73.72&58.15&73.55&\textbf{76.15}&71.60&76.06&72.97&73.86&68.37\\
san&82.38&78.24&80.28&80.32&80.03&81.20&82.99&81.33&83.10&79.99&80.98&80.60&\textbf{83.41}&78.43\\
Nar&84.08&83.50&83.76&83.42&85.02&82.93&79.32&83.72&88.38&\textbf{88.48}&88.01&88.45&86.82&80.06\\
arc&85.14&83.27&84.84&83.80&84.87&86.37&87.11&84.90&86.34&78.88&82.84&85.61&\textbf{87.19}&84.56\\
S18&80.21&78.31&80.51&80.03&79.91&84.44&81.55&75.44&82.15&81.13&80.86&78.74&\textbf{86.40}&79.24\\
OMD&82.50&80.91&80.03&76.00&75.81&82.70&76.57&76.01&81.36&74.86&74.28&75.71&\textbf{83.85}&80.63\\
HCR&70.97&68.96&69.78&68.26&70.44&72.94&72.16&69.17&72.38&62.99&66.96&68.79&\textbf{74.55}&69.74\\
STS&84.12&81.29&83.99&82.93&83.26&84.99&79.56&84.95&85.20&\textbf{87.62}&84.49&87.08&85.95&80.67\\
SSt&77.69&74.90&76.69&77.59&77.07&78.20&72.49&75.89&79.27&79.49&\textbf{84.68}&79.36&79.83&72.74\\
Tar&83.67&81.91&83.81&83.18&82.80&83.43&82.33&81.39&83.38&79.13&82.96&82.87&\textbf{84.42}&80.32\\
Vad&86.68&82.62&85.33&84.82&84.23&85.76&78.48&83.38&86.14&84.08&84.28&85.14&\textbf{87.80}&79.66\\
S13&78.43&75.83&76.82&75.52&76.10&78.67&72.24&75.95&77.64&78.40&\textbf{85.59}&79.16&80.40&74.63\\
S17&88.02&85.67&87.69&86.61&87.05&87.75&86.9&83.52&87.30&80.51&84.89&84.69&\textbf{88.37}&85.13\\
S16&81.58&79.52&81.62&80.15&81.10&82.48&81.68&78.17&81.63&75.82&79.54&80.00&\textbf{83.18}&78.62\\
\noalign{\smallskip}\hline
\#wins &1& 0& 0& 0& 0&0&0&0& 2& 4& 4& 0&\textbf{11}&0\\
rank sums &123.5&215.0&168.0&206.0&202.0&111.0&212.0&198.0&89.0&193.0&143.5&153.0&\textbf{56.0}&240.0\\
position &4.0&13.0&7.0&11.0&10.0&3.0&12.0&9.0&2.0&8.0&5.0&6.0&\textbf{1.0}&14.0\\
\noalign{\smallskip}\hline\noalign{\smallskip}
\end{tabular}
}}
\end{table}

As we can see in Tables~\ref{tab:static-svm-results-acc} and~\ref{tab:static-svm-results-f1}, RoBERTa (RoBstatic column) achieved the best performance in nine out of the 22 datasets in terms of accuracy, in 11 out of the 22 datasets in terms of $F_1$-macro, and was ranked first in the overall evaluation (position row). Regarding the number of wins (\#wins row), we can note that Emo2Vec and SSWE achieved the second best results, reaching the best performance in four out of the 22 datasets for both accuracy and $F_1$-macro. However, regarding the overall evaluation (position row), w2v-Edin and w2v-GN were ranked among the top three best static representations along with RoBERTa, in terms of accuracy. Regarding $F_1$-macro, the top three best static representations were RoBERTa, w2v-Edin and BERT (BERT-static column).

Tables~\ref{tab:overview-static-emb-acc} and~\ref{tab:overview-static-emb-f1} show a summary of the results by evaluating each static representation on the 22 datasets, for each classification algorithm. Each cell indicates the number of wins, the rank sums, and the rank position achieved by the related static representation (each line) used to train the corresponding classifier (each column). The \textit{Total} column indicates the total number of wins, the total rank sums, and the total rank position, i.e., the sum of the rank positions presented in each cell for each assessed model. Moreover, in the total column, we underline the top three best overall results in terms of total rank position.

%%%%%%%%%%%%%%%%%%%%%%%%%%% Resumo Acc %%%%%%%%%%%%%%%%%%%%%%%%%%%%%%%%%%%%%

\begin{table}%[!htbp]
\caption{Overview of the results (number of wins, rank sum, and rank position, respectively) achieved by evaluating each static representation on the 22 datasets, for each classification algorithm, in terms of accuracy}
\label{tab:overview-static-emb-acc}
\scalebox{0.76}{
\begin{tabular}{lllllll}
\hline\noalign{\smallskip}
\textbf{Representation} & \textbf{LR} & \textbf{SVM} & \textbf{MLP} & \textbf{RF} & \textbf{XGB} &\textbf{Total} \\
\noalign{\smallskip}\hline\noalign{\smallskip}
\rule{0pt}{0ex}
w2v-GN&3/116.5/4.0&1/116.5/3.0&1/151.5/6.0&1/159.0/6.0&3/125.5/4.0&9/669.0/\underline{23.0}\\
\rule{0pt}{3.0ex}
GloveWP&0/149.0/7.0&0/221.0/13.0&0/212.5/11.0&0/207.0/11.0&0/195.5/10.0&0/985.0/52.0\\
\rule{0pt}{3.0ex}
FastText&0/192.5/10.0&0/183.5/8.0&2/148.5/5.0&0/190.0/9.0&0/151.5/6.0&2/866.0/38.0\\
\rule{0pt}{3.0ex}
EWE&1/101.5/3.0&0/212.0/11.0&0/142.0/4.0&0/170.5/7.0&0/153.5/8.0&1/779.5/33.0\\
\rule{0pt}{3.0ex}
GloveTW&1/97.0/2.0&0/216.0/12.0&1/152.0/7.0&1/105.0/3.0&2/123.0/3.0&5/693.0/27.0\\
\rule{0pt}{3.0ex}
SSWE&4/152.0/8.0&4/200.5/10.0&1/234.0/14.0&6/79.5/2.0&4/153.0/7.0&19/819.0/41.0\\
\rule{0pt}{3.0ex}
TF-IDF&\textbf{5}/146.5/6.0&3/155.0/7.0&1/225.0/13.0&3/116.0/5.0&2/215.5/13.0&14/858.0/44.0\\
\rule{0pt}{3.0ex}
DeepMoji&0/174.0/9.0&0/152.5/5.0&1/167.5/8.0&0/171.0/8.0&1/144.0/5.0&2/809.0/35.0\\
\rule{0pt}{3.0ex}
w2v-Araque&0/204.0/11.0&0/185.5/9.0&0/221.0/12.0&0/202.5/10.0&0/214.5/12.0&0/1027.5/54.0\\
\rule{0pt}{3.0ex}
w2v-Edin&0/220.5/12.0&1/95.0/2.0&2/100.5/2.0&1/105.5/4.0&2/106.5/2.0&6/628.0/\textbf{\underline{22.0}}\\
\rule{0pt}{3.0ex}
Emo2Vec&4/120.0/5.0&4/154.0/6.0&2/192.0/10.0&\textbf{10}/\textbf{46.0}/\textbf{1.0}&\textbf{7}/\textbf{98.5}/\textbf{1.0}&\textbf{27}/\textbf{610.5}/\underline{23.0}\\
\rule{0pt}{3.0ex}
BERT-static&0/264.0/13.0&0/117.0/4.0&3/117.5/3.0&0/235.5/12.0&0/207.5/11.0&3/941.5/43.0\\
\rule{0pt}{3.0ex}
RoBERTa-static&\textbf{5}/\textbf{82.5}/\textbf{1.0}&\textbf{9}/\textbf{56.5}/\textbf{1.0}&\textbf{8}/\textbf{74.0}/\textbf{1.0}&0/244.0/13.0&1/167.5/9.0&23/624.5/25.0\\
\rule{0pt}{3.0ex}
BERTweet-static&0/290.0/14.0&0/245.0/14.0&0/172.0/9.0&0/278.5/14.0&0/254.0/14.0&0/1239.5/65.0\\
\noalign{\smallskip}\hline
\end{tabular}
}
\end{table}

%%%%%%%%%%%%%%%%%%%%%%%%%%%%%% F1 macro %%%%%%%%%%%%%%%%%%%%%%%%%%%%%%%%%%%%%%%%%%%

\begin{table}%[!htbp]
\caption{Overview of the results (number of wins, rank sum, and rank position, respectively) achieved by evaluating each static representation on the 22 datasets, for each classification algorithm, in terms of $F_1$-macro}
\label{tab:overview-static-emb-f1}
\scalebox{0.75}{
\begin{tabular}{lllllll}
\hline\noalign{\smallskip}
\textbf{Representation} & \textbf{LR} & \textbf{SVM} & \textbf{MLP} & \textbf{RF} & \textbf{XGB} & \textbf{Total} \\
\noalign{\smallskip}\hline\noalign{\smallskip}
\rule{0pt}{0ex}
w2v-GN&3/113.0/4.0&1/123.5/4.0&1/146.5/5.0&1/161.5/6.0&3/129.0/4.0&9/673.5/\underline{23.0}\\
\rule{0pt}{3.0ex}
GloveWP&0/152.0/6.5&0/215.0/13.0&1/213.0/11.0&0/208.0/11.0&0/195.5/10.0&1/983.5/51.5\\
\rule{0pt}{3.0ex}
FastText&0/189.0/10.0&0/168.0/7.0&1/149.5/6.0&0/200.0/9.0&0/153.0/7.0&1/859.5/39.0\\
\rule{0pt}{3.0ex}
EWE&1/102.5/3.0&0/206.0/11.0&0/144.0/4.0&0/165.5/7.0&0/160.5/8.0&1/778.5/33.0\\
\rule{0pt}{3.0ex}
GloveTW&1/96.0/2.0&0/202.0/10.0&1/152.0/7.0&0/109.5/5.0&1/123.0/3.0&3/682.5/27.0\\
\rule{0pt}{3.0ex}
SSWE&\textbf{5}/152.0/6.5&4/193.0/8.0&1/228.0/13.0&6/70.0/2.0&4/135.0/5.0&20/778.0/34.5\\
\rule{0pt}{3.0ex}
TF-IDF&3/166.0/8.0&0/212.0/12.0&1/243.0/14.0&4/106.5/4.0&3/211.5/11.0&11/939.0/49.0\\
\rule{0pt}{3.0ex}
DeepMoji&0/169.0/9.0&0/153.0/6.0&1/164.0/8.0&0/173.5/8.0&1/143.0/6.0&2/802.5/37.0\\
\rule{0pt}{3.0ex}
w2v-Araque&0/200.0/11.0&0/198.0/9.0&0/222.0/12.0&0/204.5/10.0&0/216.5/12.0&0/1041.0/54.0\\
\rule{0pt}{3.0ex}
w2v-Edin&0/221.0/12.0&2/89.0/2.0&2/95.0/2.0&0/105.0/3.0&2/110.0/2.0&6/620.0/\textbf{\underline{21.0}}\\
\rule{0pt}{3.0ex}
Emo2Vec&4/117.5/5.0&4/143.5/5.0&2/191.0/10.0&\textbf{11}/\textbf{42.0}/\textbf{1.0}&\textbf{8}/\textbf{81.5}/\textbf{1.0}&\textbf{29}/\textbf{575.5}/\underline{22.0}\\
\rule{0pt}{3.0ex}
BERT-static&0/261.0/13.0&0/111.0/3.0&3/112.5/3.0&0/234.0/12.0&0/217.5/13.0&3/936.0/44.0\\
\rule{0pt}{3.0ex}
RoBERTa-static&\textbf{5}/\textbf{82.0}/\textbf{1.0}&\textbf{11}/\textbf{56.0}/\textbf{1.0}&\textbf{8}/\textbf{75.5}/\textbf{1.0}&0/245.5/13.0&0/177.0/9.0&24/636.0/25.0\\
\rule{0pt}{3.0ex}
BERTweet-static&0/289.0/14.0&0/240.0/14.0&0/174.0/9.0&0/284.5/14.0&0/257.0/14.0&0/1244.5/65.0\\
\noalign{\smallskip}\hline
\end{tabular}
}
\end{table}

Regarding the overall evaluation (Total column), from Tables~\ref{tab:overview-static-emb-acc} and~\ref{tab:overview-static-emb-f1}, we can see that although Emo2Vec achieved the highest total number of wins (i.e., 27 wins in terms of accuracy, and 29 wins in terms of $F_1$-macro), w2v-Edin was ranked as the best overall model, achieving the lowest total rank position for both accuracy (22.0) and $F_1$-macro (21.0). Nevertheless, considering each classifier (each column), we can note that RoBERTa achieved the best performance when used to train LR, SVM, and MLP, for both accuracy and $F_1$-macro.~Conversely, Emo2Vec achieved the best overall results when used to train RF and XGB classifiers. Analyzing the overall results in terms of the total rank position (Total column), we observe that Emo2Vec and w2v-GN, along with w2v-Edin, are ranked as the top three best static representations. These results suggest that w2v-Edin, Emo2Vec, and w2v-GN are well-suited static representations for Twitter sentiment analysis.

In the previous evaluations, we analyzed the predictive performance achieved by each representation for one classification algorithm at a time, focusing on the individual contribution of the text representations in the performance on the final task. Next, we investigate the classification performance of the final sentiment analysis process, that is, the combination of text representation and classifier. Considering that the final classification is a combination of both representation and classifier, an appropriate choice of the classification algorithm may affect the performance of a text representation. For this purpose, we present an overall evaluation of all possible combinations of text representations and classification algorithms, examining them as pairs \{\textit{text representation}, \textit{classifier}\}. More clearly, we evaluate the classification effectiveness of 70 possible combinations of text representations and classifiers (14 $\times$ 5) on the 22 datasets of tweets. Table~\ref{tab:top10_ranksum} presents the top ten results in terms of the average rank position and ~\ref{tab:tail10_ranksum} presents the ten worst average rank position. Specifically, for each dataset, we calculate a rank of the 70 combinations and then average the rank position of each combination over the 22 datasets. From Table~\ref{tab:top10_ranksum}, we can note that the best overall results were achieved by using RoBERTa to train an SVM classifier for both accuracy and $F_1$-macro. Also, w2v-Edin $+$ SVM and RoBERTA $+$ MLP appear in the top three results along with RoBERTa $+$ SVM. By Table~\ref{tab:tail10_ranksum}, we can notice the high-frequency of RF in the pair Model-Classifier.

Tables~\ref{tab:emb_ranksum} and~\ref{tab:class_ranksum} show a summary of the results for each text representation and classifier, respectively, from best to worst, in terms of the average rank position. As we can observe, Emo2Vec, RoBERTa, and w2v-Edin appear in the top three, being the representations that achieved the best overall performances. Among the classifiers, we can note that SVM and MLP seem to be good choices in Twitter sentiment Analysis regarding the usage of static text representations. Conversely, RF achieved the worst overall performance across all evaluations.

In addition to the individual assessment of text representations and classifiers presented in Tables~\ref{tab:emb_ranksum} and~\ref{tab:class_ranksum}, Table~\ref{tab:bestoverall-static-emb} shows the best results achieved for each dataset. We can see that RoBERTa achieved the highest accuracies in seven out of the 22 datasets, and highest $F_1$-macro scores in nine out of the 22 datasets. Furthermore, as highlighted in Table~\ref{tab:top10_ranksum}, RoBERTA $+$ SVM achieved the best performances in six out of the 22 datasets in terms of accuracy, and in eight out of the 22 datasets in terms of $F_1$-macro.

The top three static representations identified in the previous experiments, i.e., RoBERTa, w2v-Edin, and Emo2Vec, are very different from each other. While w2v-Edin and Emo2Vec were trained from scratch on tweets, RoBERTa was trained on traditional English texts. 

The better performance of Emo2Vec and w2v-Edin can be caused by the inclusion of the sentiment analysis task in its training process. We also have other models built in this same strategy and trained from scratch with tweets, such as Deepmoji and SSWE, respectively, the seventh and eighth position by Table~\ref{tab:emb_ranksum}. The Emo2Vec better performance may be a result of its multi-task learning approach. Considering another model with the same architecture of w2v-Edin and also trained from scratch with tweets, the differential performance between w2v-Edin and w2v-Araque (the fourteenth position by Table~\ref{tab:emb_ranksum}) may lie in its volume of training data (w2v-Araque: 1.28M, and W2V-Edin: 10M) and the vocabulary size (w2v-Araque: 57K and w2v-Edin: 259K).

However, among these, RoBERTa is the only Transformer-based model, which holds state-of-the-art performance in capturing context and semantics of terms from texts. Furthermore, regarding w2v-Edin, although it was trained with a more straightforward architecture (feedforward neural network) as compared to others, its training parameters were optimized for the emotion detection task on tweets~\cite{7817108}, which may have helped determining the sentiment expressed in tweets.

Surprisingly, as shown in Table~\ref{tab:emb_ranksum}, BERTweet achieved the worst overall performance among all assessed text representations, despite having been trained using the same state-of-the-art Transformer-based architecture as RoBERTa \emph{while yet using tweets}. One possible explanation for this behavior is that BERTweet training procedure limits the representation of its training tweets to 60 tokens only, while RoBERTa uses a limit of 512 tokens. For that reason, we believe that RoBERTa model is able to capture more semantic information to the tokens from its training vocabulary as compared to BERTweet when one collects the token representation from the embeddings layer.

Finally, regarding research question RQ1, we can highlight and suggest that: (i) disregarding the classification algorithms, Emo2Vec, w2v-Edin, and RoBERTa seem to be well-suited representations for determining the sentiment expressed in tweets, and (ii) considering the combination of text representations and classifiers, RoBERTa $+$ SVM achieved the best overall performance, which may represent a good choice for Twitter sentiment analysis in hardware-restricted environments, since the cost here is most due to the classifier induction.

\begin{table}%[!htbp]
\caption{Top 10 average rank results achieved for each combination Model-Classifier}
\label{tab:top10_ranksum}
\scalebox{0.83}{
\begin{tabular}{lll|lll}
\hline\noalign{\smallskip}
\multirow{2}{*}{\textbf{Representation}} & \multirow{2}{*}{\textbf{Classifier}} & \textbf{Accuracy} & \multirow{2}{*}{\textbf{Representation}} & \multirow{2}{*}{\textbf{Classifier}} & \textbf{$\bm{F_1}$-macro} \\
& & \textbf{avg. rank pos.} & & & \textbf{avg. rank pos.}\\
\noalign{\smallskip}\hline\noalign{\smallskip}
\rule{0pt}{3ex}
RoBERTa-static&SVM&9.32&RoBERTa-static&SVM&8.59\\
\rule{0pt}{3.0ex}
RoBERTa-static&MLP&11.57&W2V-Edin&SVM&9.39\\
\rule{0pt}{3.0ex}
W2V-Edin&SVM&14.50&RoBERTa-static&MLP&12.52\\
\rule{0pt}{3.0ex}
W2V-Edin&MLP&15.36&BERT-static&SVM&14.70\\
\rule{0pt}{3.0ex}
BERT-static&MLP&16.68&W2V-GN&SVM&14.95\\
\rule{0pt}{3.0ex}
W2V-GN&SVM&17.80&W2V-Edin&MLP&15.55\\
\rule{0pt}{3.0ex}
BERT-static&SVM&19.02&RoBERTa-static&LR&16.02\\
\rule{0pt}{3.0ex}
Emo2Vec&Xgb&21.23&BERT-static&MLP&17.48\\
\rule{0pt}{3.0ex}
W2V-GN&MLP&22.50&GloVe-TWT&LR&17.91\\
\rule{0pt}{3.0ex}
fastText&MLP&23.20&Emo2Vec&SVM&18.05\\

\noalign{\smallskip}
\hline
\end{tabular}
}
\end{table}

\begin{table}%[!htbp]
\caption{Tail 10 average rank results achieved for each combination Model-Classifier}
\label{tab:tail10_ranksum}
\scalebox{0.83}{
\begin{tabular}{lll|lll}
\hline\noalign{\smallskip}
\multirow{2}{*}{\textbf{Representation}} & \multirow{2}{*}{\textbf{Classifier}} & \textbf{Accuracy} & \multirow{2}{*}{\textbf{Representation}} & \multirow{2}{*}{\textbf{Classifier}} & \textbf{$\bm{F_1}$-macro} \\
& & \textbf{avg. rank pos.} & & & \textbf{avg. rank pos.}\\
\noalign{\smallskip}\hline\noalign{\smallskip}
\rule{0pt}{3ex}
DeepMoji&RF&51.41&EWE&RF&56.86\\
\rule{0pt}{3.0ex}
BERTweet-static&Xgb&52.14&DeepMoji&RF&57.43\\
\rule{0pt}{3.0ex}
fastText&RF&52.75&BERTweet-static&LR&57.45\\
\rule{0pt}{3.0ex}
GloVe-WP&RF&54.11&W2V-GN&RF&58.02\\
\rule{0pt}{3.0ex}
W2V-Araque&RF&54.68&fastText&RF&60.36\\
\rule{0pt}{3.0ex}
BERT-static&RF&57.18&W2V-Araque&RF&61.00\\
\rule{0pt}{3.0ex}
RoBERTa-static&RF&57.95&GloVe-WP&RF&61.16\\
\rule{0pt}{3.0ex}
BERT-static&LR&60.86&BERT-static&RF&63.91\\
\rule{0pt}{3.0ex}
BERTweet-static&RF&61.02&RoBERTa-static&RF&64.11\\
\rule{0pt}{3.0ex}
BERTweet-static&LR&66.09&BERTweet-static&RF&67.32\\

\noalign{\smallskip}
\hline
\end{tabular}
}
\end{table}

%%%%%%%%%%%%%%%%%%%%%%%%%%%%%%%%%%%%%%%%%%%%%%%%%%%%%%%%%%%%%%%%%%%%%%%%%
\begin{table}%[!htbp]
\caption{Average Rank results achieved for each Embedding, evaluating static representations}
\label{tab:emb_ranksum}
%\scalebox{0.95}{
\begin{tabular}{ll|ll}
\hline\noalign{\smallskip}
\multirow{2}{*}{\textbf{Representation}} & \textbf{Accuracy} & \multirow{2}{*}{\textbf{Representation}} & \textbf{$\bm{F_1}$-macro} \\
& \textbf{avg. rank pos.} & & \textbf{avg. rank pos.}\\
\noalign{\smallskip}\hline\noalign{\smallskip}
\rule{0pt}{3ex}
Emo2Vec&26.35&Emo2Vec&25.34\\
\rule{0pt}{3.0ex}
RoBERTa-static&27.93&RoBERTa-static&29.06\\
\rule{0pt}{3.0ex}
W2V-Edin&29.55&W2V-Edin&30.05\\
\rule{0pt}{3.0ex}
W2V-GN&30.05&W2V-GN&31.04\\
\rule{0pt}{3.0ex}
GloVe-TWT&32.00&GloVe-TWT&31.55\\
\rule{0pt}{3.0ex}
EWE&34.50&SSWE&33.40\\
\rule{0pt}{3.0ex}
SSWE&34.78&EWE&34.15\\
\rule{0pt}{3.0ex}
DeepMoji&35.93&DeepMoji&34.98\\
\rule{0pt}{3.0ex}
TF-IDF&36.21&fastText&36.34\\
\rule{0pt}{3.0ex}
fastText&37.00&BERT-static&39.38\\
\rule{0pt}{3.0ex}
BERT-static&39.43&GloVe-WP&39.43\\
\rule{0pt}{3.0ex}
GloVe-WP&40.25&TF-IDF&40.25\\
\rule{0pt}{3.0ex}
W2V-Araque&43.15&W2V-Araque&42.85\\
\rule{0pt}{3.0ex}
BERTweet-static&49.88&BERTweet-static&49.18\\
\noalign{\smallskip}
\hline
\end{tabular}
%}
\end{table}

%%%%%%%%%%%%%%%%%%%%%%%%%%%%%%%%%%%%%%%%%%%%%%%%%%%%%%%%%%%%%%%%%%%%%%%%%
\begin{table}%[!htbp]
\caption{Average Rank results achieved for each Classifier, evaluating static representations}
\label{tab:class_ranksum}
%\scalebox{0.8}{
\begin{tabular}{ll|ll}
\hline\noalign{\smallskip}
\multirow{2}{*}{\textbf{Classifier}} & \textbf{Accuracy} & \multirow{2}{*}{\textbf{Classifier}} & \textbf{$\bm{F_1}$-macro} \\
& \textbf{avg. rank pos.} & & \textbf{avg. rank pos.}\\
\noalign{\smallskip}\hline\noalign{\smallskip}
\rule{0pt}{3ex}
MLP&26.28&SVM&23.07\\
\rule{0pt}{3.0ex}
SVM&28.30&MLP&26.69\\
\rule{0pt}{3.0ex}
XGB&35.83&LR&31.27\\
\rule{0pt}{3.0ex}
LR&39.17&XGB&41.33\\
\rule{0pt}{3.0ex}
RF&47.92&RF&55.15\\

\noalign{\smallskip}
\hline
\end{tabular}
%}
\end{table}
%%%%%%%%%%%%%%%%%%%%%%%%%%%%%%%%%%%%%%%%%%%%%%%%%%%%%%%%%%%%%%%%%%%%%%%%%
\begin{table}%[!htbp]
\caption{Best results achieved for each dataset}
\label{tab:bestoverall-static-emb}
\scalebox{0.83}{
\begin{tabular}{llll|lll}
\hline\noalign{\smallskip}
\textbf{Dataset}&\textbf{Accuracy}& \textbf{Classifier} &\textbf{Representation} & \bm{$F_1$}\textbf{-macro}& \textbf{Classifier} &\textbf{Representation} \\
\noalign{\smallskip}\hline\noalign{\smallskip}
\rule{0pt}{3ex}
iro&78.81&LR&Emo2Vec&75.87&LR&Emo2Vec\\
\rule{0pt}{3.0ex}
sar&87.50&LR&SSWE&87.19&LR&SSWE\\
\rule{0pt}{3.0ex}
ntu&95.30&MLP&w2v-Edin&95.19&MLP&w2v-Edin\\
\rule{0pt}{3.0ex}
S15&90.35&LR&TF-IDF&78.48&SVM&RoBERTa-static\\
\rule{0pt}{3.0ex}
stm&87.74&SVM&w2v-Edin&87.67&SVM&w2v-Edin\\
\rule{0pt}{3.0ex}
per&83.83&MLP&w2v-GN&80.58&SVM&w2v-GN\\
\rule{0pt}{3.0ex}
hob&94.82&MLP&BERT-static&94.05&MLP&BERT-static\\
\rule{0pt}{3.0ex}
iph&84.39&MLP&GloVe-TWT&81.15&MLP&GloVe-TWT\\
\rule{0pt}{3.0ex}
mov&88.78&XGB&Emo2Vec&77.86&MLP&fastText\\
\rule{0pt}{3.0ex}
san&84.71&MLP&TF-IDF&84.56&MLP&TF-IDF\\
\rule{0pt}{3.0ex}
Nar&89.00&LR&SSWE&88.58&LR&SSWE\\
\rule{0pt}{3.0ex}
arc&87.60&MLP&RoBERTa-static&87.29&MLP&RoBERTa-static\\
\rule{0pt}{3.0ex}
S18&86.50&SVM&RoBERTa-static&86.40&SVM&RoBERTa-static\\
\rule{0pt}{3.0ex}
OMD&85.10&SVM&RoBERTa-static&83.85&SVM&RoBERTa-static\\
\rule{0pt}{3.0ex}
HCR&80.24&SVM&TF-IDF&74.55&SVM&RoBERTa-static\\
\rule{0pt}{3.0ex}
STS&89.08&LR&SSWE&87.70&LR&SSWE\\
\rule{0pt}{3.0ex}
SST&85.06&LR&Emo2Vec&84.77&LR&Emo2Vec\\
\rule{0pt}{3.0ex}
Tar&84.42&SVM&RoBERTa-static&84.42&SVM&RoBERTa-static\\
\rule{0pt}{3.0ex}
Vad&89.32&SVM&RoBERTa-static&87.80&SVM&RoBERTa-static\\
\rule{0pt}{3.0ex}
S13&88.24&XGB&Emo2Vec&85.59&LR&Emo2Vec\\
\rule{0pt}{3.0ex}
S17&89.03&SVM&RoBERTa-static&88.37&SVM&RoBERTa-static\\
\rule{0pt}{3.0ex}
S16&86.00&SVM&RoBERTa-static&83.18&SVM&RoBERTa-static\\

\noalign{\smallskip}
\hline
\end{tabular}
}
\end{table}

\section{Evaluation of the Transformer-based text representations}
\label{sec:context}

In this section, we address the research question RQ2, as follows:

\textit{RQ2.Considering state-of-the-art Transformer-based autoencoder models, which are the most effective in the sentiment classification of tweets?}

To answer that question, we conduct a thorough evaluation of the widely used BERT and RoBERTa models and the BERT-based transformer trained from scratch with tweets, namely, BERTweet. These models represent a set of the most recent Transformer-based autoencoder language modeling techniques that have achieved state-of-the-art performance in many NLP tasks. While BERT is the first Transformer-based autoenconder model to appear in the literature, RoBERTa is an evolution of BERT with improved training methodology, due to the elimination of the Next Sentence Prediction task, which may fit NLP tasks on tweets considering they are limited in size and self-contained in context. Moreover, by evaluating BERTweet we analyze the performance of a Transformer-based model trained from scratch on tweets. 

In this set of experiments, we give an example tweet as input to the transformer model and concatenate its last four layers to be the token representation and the tweet representation is the average of the tokens representation. Next, those representations collected from the whole dataset are given as input to the learning classifier method together with the labels of the tweets. Finally, the learned classifier is employed to perform the evaluation. In this way, we once again follow the feature extraction plus classification strategy but now using the contextualized embedding from each tweet. 

Table~\ref{tab:context-svm-results-acc-f1} presents the classification results when using the SVM classifier in terms of accuracy and $F_1$-macro, and Table~\ref{tab:overview-context-emb} shows a summary of the complete evaluation regarding all classifiers. As in previous section, to limit the number of tables in the manuscript, we only report the computational results in detail for the LR classifier as an example of this evaluation (refer to Online Resource 1 for the detailed evaluation). From Table~\ref{tab:context-svm-results-acc-f1}, we can note that BERTweet achieved the best results in 18 out of the 22 datasets for both accuracy and $F_1$-macro. Similarly, regarding all classifiers, Table~\ref{tab:overview-context-emb} shows that BERTweet outperformed BERT and RoBERTa by a significant difference in terms of the total number of wins for both accuracy~and~$F_1$-macro.

\begin{table}%[!htbp]
{\caption{Accuracies and $F_1$-macro scores (\%) achieved by evaluating Transformer-Autoencoder language models using the SVM classifier}
\label{tab:context-svm-results-acc-f1}
\scalebox{0.94}{
\begin{tabular}{llll|lll}
\hline\noalign{\smallskip}
 \multirow{2}{*}{\textbf{Dataset}}& \multicolumn{3}{c}{\textbf{Accuracy}} & \multicolumn{3}{c}{\bm{$F_1$}\textbf{-macro}} \\
\noalign{\smallskip}\noalign{\smallskip}
 & \textbf{RoBERTa} & \textbf{BERT} & \textbf{BERTweet} & \textbf{RoBERTa} & \textbf{BERT} & \textbf{BERTweet}\\
\noalign{\smallskip}\hline\noalign{\smallskip}
iro&46.67&\textbf{71.43}&69.52&31.0&\textbf{66.51}&61.22\\
sar&57.86&\textbf{76.07}&61.96&46.23&\textbf{75.63}&59.24\\
ntu&78.45&87.02&\textbf{91.03}&78.18&86.73&\textbf{90.86}\\
S15&86.31&90.03&\textbf{91.59}&73.33&77.58&\textbf{82.06}\\
stm&84.12&89.14&\textbf{90.25}&84.04&89.11&\textbf{90.23}\\
per&72.66&83.13&\textbf{83.14}&71.18&80.73&\textbf{81.34}\\
hob&69.15&\textbf{83.52}&83.13&68.62&\textbf{81.76}&81.53\\
iph&75.96&81.58&\textbf{83.65}&74.65&79.63&\textbf{81.97}\\
mov&74.35&84.14&\textbf{86.47}&68.61&77.58&\textbf{80.55}\\
san&83.66&85.54&\textbf{89.87}&83.43&85.47&\textbf{89.81}\\
Nar&89.73&91.6&\textbf{95.35}&89.48&91.34&\textbf{95.22}\\
arc&88.18&87.25&\textbf{90.16}&88.0&87.02&\textbf{89.99}\\
S18&86.28&87.25&\textbf{88.97}&86.07&87.16&\textbf{88.87}\\
OMD&82.16&85.62&\textbf{87.36}&81.2&84.71&\textbf{86.4}\\
HCR&76.67&78.61&\textbf{79.82}&72.89&74.75&\textbf{76.22}\\
STS&89.48&90.46&\textbf{93.56}&88.11&89.16&\textbf{92.65}\\
SSt&84.01&85.19&\textbf{86.76}&83.83&84.87&\textbf{86.53}\\
Tar&84.83&85.64&\textbf{86.93}&84.81&85.62&\textbf{86.92}\\
Vad&87.73&89.63&\textbf{90.56}&86.24&88.18&\textbf{89.28}\\
S13&84.26&86.62&\textbf{88.15}&82.0&84.21&\textbf{86.13}\\
S17&90.61&91.54&\textbf{92.56}&90.08&91.03&\textbf{92.08}\\
S16&87.62&88.77&\textbf{90.72}&85.5&86.59&\textbf{88.86}\\
\noalign{\smallskip}\hline
\#wins &0& 3& \textbf{19}& 0& 3&\textbf{19}\\
rank sums &65.0&42.0&\textbf{25.0}&65.0&42.0&\textbf{25.0}\\
position &3.0&2.0&\textbf{1.0}&3.0&2.0&\textbf{1.0}\\
\noalign{\smallskip}\hline
\end{tabular}
}}
\end{table}

%%%%%%%%%%%%%%%%%%%% Resumo Acc %%%%%%%%%%%%%%%%%%%%%%%

\begin{table}%[!htbp]
\caption{Overview of the results (number of wins, rank sum, and rank position, respectively) achieved by evaluating each Transformer-Autoencoder model on the 22 datasets, for each classification algorithm}
\label{tab:overview-context-emb}
\scalebox{0.78}{
\begin{tabular}{lllllll}
\hline\noalign{\smallskip}
\textbf{Embedding} & \textbf{LR} & \textbf{SVM} & \textbf{MLP} & \textbf{RF} & \textbf{XGB} &\textbf{Total}\\
\noalign{\smallskip}\hline\noalign{\smallskip}
\multicolumn{7}{c}{\textbf{ACCURACY}}\\
\noalign{\smallskip}\hline\noalign{\smallskip}
\rule{0pt}{0.1ex}
BERT&2/45.0/2.0&0/65.0/3.0&3/47.0/2.0&4/46.5/2.0&3/43.5/2.0&12/247.0/11.0\\
\rule{0pt}{3.0ex}
RoBERTa&2/60.0/3.0&3/42.0/2.0&2/57.0/3.0&5/52.5/3.0&0/63.0/3.0&12/274.5/14.0\\
\rule{0pt}{3.0ex}
BERTweet&\textbf{18}/\textbf{27.0}/\textbf{1.0}&\textbf{19}/\textbf{25.0}/\textbf{1.0}&\textbf{17}/\textbf{28.0}/\textbf{1.0}&\textbf{15}/\textbf{33.0}/\textbf{1.0}&\textbf{20}/\textbf{25.5}/\textbf{1.0}&\textbf{89}/\textbf{138.5}/\textbf{5.0}\\
\noalign{\smallskip}\hline\noalign{\smallskip}
\multicolumn{7}{c}{\bm{$F_1$}\textbf{-MACRO}}\\
\noalign{\smallskip}\hline\noalign{\smallskip}
\rule{0pt}{1ex}
BERT&2/45.0/2.0&0/65.0/3.0&3/47.0/2.0&3/48.0/2.0&3/44.0/2.0&11/249.0/11.0\\
\rule{0pt}{3.0ex}
RoBERTa&2/60.0/3.0&3/42.0/2.0&2/57.0/3.0&5/52.5/3.0&0/61.0/3.0&12/272.5/14.0\\
\rule{0pt}{3.0ex}
BERTweet&\textbf{18}/\textbf{27.0}/\textbf{1.0}&\textbf{19}/\textbf{25.0}/\textbf{1.0}&\textbf{17}/\textbf{28.0}/\textbf{1.0}&\textbf{15}/\textbf{31.5}/\textbf{1.0}&\textbf{19}/\textbf{27.0}/\textbf{1.0}&\textbf{88}/\textbf{138.5}/\textbf{5.0}\\
\noalign{\smallskip}\hline
\end{tabular}
}
\end{table}

%%%%%%%%%%%%%%%%%%%%%%%%%%%%%%%%%%%%%%%%%%%%%%%%%%%%%%%%%%%
Next, we present an overall analysis of using BERT, RoBERTa, and BERTweet models to train each one of the five classification algorithms, examining them as pairs \{\textit{language model}, \textit{classifier}\}. Table~\ref{tab:top10_ranksum_context} presents the average rank position across all 15 possible combinations (3 language models $\times$ 5 classification algorithms), from best to worst, as explained in Section~\ref{sec:static-exp}. We can observe that BERTweet combined with LR, MLP, and SVM classifiers achieved the best overall performances for both accuracy and $F_1$-macro. Conversely, using RF to train the Transformer-based embeddings seems to harm the performance of the models.

\begin{table}%[!htbp]
\caption{Average rank position results achieved for each combination Model-Classifier}
\label{tab:top10_ranksum_context}
\scalebox{0.94}{
\begin{tabular}{lll|lll}
\hline\noalign{\smallskip}
\multirow{2}{*}{\textbf{Model}} & \multirow{2}{*}{\textbf{Classifier}} & \textbf{Accuracy} & \multirow{2}{*}{\textbf{Model}} & \multirow{2}{*}{\textbf{Classifier}} & \textbf{$\bm{F_1}$-macro} \\
& & \textbf{avg. rank pos.} & & & \textbf{avg. rank pos.}\\
\noalign{\smallskip}\hline\noalign{\smallskip}
\rule{0pt}{3ex}
BERTweet&LR&2.55&BERTweet&LR&2.32\\
\rule{0pt}{3.0ex}
BERTweet&MLP&2.68&BERTweet&MLP&3.00\\
\rule{0pt}{3.0ex}
BERTweet&SVM&4.16&BERTweet&SVM&3.55\\
\rule{0pt}{3.0ex}
RoBERTa&MLP&5.05&RoBERTa&LR&4.98\\
\rule{0pt}{3.0ex}
RoBERTa&LR&5.68&RoBERTa&MLP&5.39\\
\rule{0pt}{3.0ex}
BERT&MLP&6.23&BERT&SVM&5.75\\
\rule{0pt}{3.0ex}
BERT&SVM&6.86&BERT&MLP&6.61\\
\rule{0pt}{3.0ex}
BERTweet&Xgb&7.34&BERT&LR&7.30\\
\rule{0pt}{3.0ex}
BERT&LR&7.73&BERTweet&Xgb&8.36\\
\rule{0pt}{3.0ex}
RoBERTa&Xgb&9.57&RoBERTa&Xgb&10.18\\
\rule{0pt}{3.0ex}
BERT&Xgb&11.61&RoBERTa&SVM&10.80\\
\rule{0pt}{3.0ex}
BERTweet&RF&11.95&BERT&Xgb&11.77\\
\rule{0pt}{3.0ex}
RoBERTa&SVM&12.05&BERTweet&RF&12.48\\
\rule{0pt}{3.0ex}
RoBERTa&RF&12.98&RoBERTa&RF&13.55\\
\rule{0pt}{3.0ex}
BERT&RF&13.57&BERT&RF&13.98\\

\noalign{\smallskip}
\hline
\end{tabular}
}
\end{table}

Tables~\ref{tab:emb_ranksum-context} and~\ref{tab:class_ranksum-contex} show a summary of the results for each model and classifier, respectively, from best to worst, in terms of the average rank position. From Table~\ref{tab:emb_ranksum-context}, we can see that BERTweet achieved the best overall classification effectiveness and was ranked first. Also, RoBERTa and BERT achieved comparable overall performances for both accuracy and $F_1$-macro. Regarding the classifiers, as shown in Table~\ref{tab:class_ranksum-contex}, MLP and LR achieved rather comparable performances and were ranked as the top two best classifiers regarding the Transformer-based models, followed by SVM, XGB, and RF.

\begin{table}%[!htbp]
\caption{Average rank position results achieved for each Embedding}
\label{tab:emb_ranksum-context}
%\scalebox{0.8}{
\begin{tabular}{ll|ll}
\hline\noalign{\smallskip}
\multirow{2}{*}{\textbf{Model}} & \textbf{Accuracy} & \multirow{2}{*}{\textbf{Model}} & \textbf{$\bm{F_1}$-macro} \\
& \textbf{avg. rank pos.} & & \textbf{avg. rank pos.}\\
\noalign{\smallskip}\hline\noalign{\smallskip}
\rule{0pt}{3ex}
BERTweet&5.74&BERTweet&5.94\\
\rule{0pt}{3.0ex}
RoBERTa&9.06&RoBERTa&8.98\\
\rule{0pt}{3.0ex}
BERT&9.20&BERT&9.08\\

\noalign{\smallskip}
\hline
\end{tabular}
%}
\end{table}

%%%%%%%%%%%%%%%%%%%%%%%%%%%%%%%%%%%%%%%%%%%%%%%%%%%%%%%%%%%%%%%%%%%%%%%%%
\begin{table}%[!htbp]
\caption{Average rank position results achieved for each Classifier}
\label{tab:class_ranksum-contex}
%\scalebox{0.8}{
\begin{tabular}{ll|ll}
\hline\noalign{\smallskip}
\multirow{2}{*}{\textbf{Classifier}} & \textbf{Accuracy} & \multirow{2}{*}{\textbf{Classifier}} & \textbf{$\bm{F_1}$-macro} \\
& \textbf{avg. rank pos.} & & \textbf{avg. rank pos.}\\
\noalign{\smallskip}\hline\noalign{\smallskip}
\rule{0pt}{3ex}
MLP&4.65&LR&4.86\\
\rule{0pt}{3.0ex}
LR&5.32&MLP&5.00\\
\rule{0pt}{3.0ex}
SVM&7.69&SVM&6.70\\
\rule{0pt}{3.0ex}
XGB&9.51&XGB&10.11\\
\rule{0pt}{3.0ex}
RF&12.83&RF&13.33\\
\noalign{\smallskip}
\hline
\end{tabular}
%}
\end{table}

Regarding the results achieved for each dataset, Table~\ref{tab:summarization-context-emb} presents the best results in terms of accuracy and $F_1$-macro. As we can notice, BERTweet outperformed BERT and RoBERTa in 17 out of the 22 datasets in terms of accuracy and in 18 out of the 22 datasets in terms of $F_1$-macro. These results may confirm that Twitter sentiment classification benefits most from contextualized language models trained from scratch on Twitter data. Unlike BERT and RoBERTa, which were trained on traditional English texts, BERTweet was trained on a huge amount of 850M tweets. This fact may have helped BERTweet on learning the specificities of tweets, such as their morphological and semantic characteristics.

\begin{table}%[!htbp]
\caption{Best results achieved for each dataset by evaluating the combination of language model and classifier}
\label{tab:summarization-context-emb}
\scalebox{0.9}{
\begin{tabular}{llll|lll}
\hline\noalign{\smallskip}
\textbf{Dataset}&\textbf{Accuracy}& \textbf{Classifier} &\textbf{Model} & $\bm{F_1}$\textbf{-macro}& \textbf{Classifier} &\textbf{Model} \\
\noalign{\smallskip}\hline\noalign{\smallskip}
\rule{0pt}{0ex}
iro&80.48&LR&BERT&73.08&LR&BERT\\
\rule{0pt}{3.0ex}
sar&76.07&SVM&BERT&75.63&SVM&BERT\\
\rule{0pt}{3.0ex}
ntu&91.03&LR&BERTweet&90.86&SVM&BERTweet\\
\rule{0pt}{3.0ex}
S15&92.23&LR&BERTweet&83.33&LR&BERTweet\\
\rule{0pt}{3.0ex}
stm&90.79&LR&RoBERTa&90.75&LR&RoBERTa\\
\rule{0pt}{3.0ex}
per&87.69&LR&BERTweet&85.29&LR&BERTweet\\
\rule{0pt}{3.0ex}
hob&87.93&LR&RoBERTa&86.29&LR&RoBERTa\\
\rule{0pt}{3.0ex}
iph&87.59&MLP&BERTweet&84.72&MLP&BERTweet\\
\rule{0pt}{3.0ex}
mov&89.47&MLP&RoBERTa&82.12&LR&BERTweet\\
\rule{0pt}{3.0ex}
sand&91.17&MLP&BERTweet&91.11&MLP&BERTweet\\
\rule{0pt}{3.0ex}
Nar&95.60&MLP&BERTweet&95.43&MLP&BERTweet\\
\rule{0pt}{3.0ex}
arc&90.74&MLP&BERTweet&90.51&MLP&BERTweet\\
\rule{0pt}{3.0ex}
S18&88.97&SVM&BERTweet&88.87&SVM&BERTweet\\
\rule{0pt}{3.0ex}
OMD&87.36&SVM&BERTweet&86.40&SVM&BERTweet\\
\rule{0pt}{3.0ex}
HCR&81.55&XGB&BERTweet&76.77&LR&BERTweet\\
\rule{0pt}{3.0ex}
STS&93.90&LR&BERTweet&92.98&LR&BERTweet\\
\rule{0pt}{3.0ex}
SSt&86.76&SVM&BERTweet&86.53&SVM&BERTweet\\
\rule{0pt}{3.0ex}
Tar&86.93&SVM&BERTweet&86.92&SVM&BERTweet\\
\rule{0pt}{3.0ex}
Vad&90.80&LR&BERTweet&89.38&LR&BERTweet\\
\rule{0pt}{3.0ex}
S13&89.61&LR&BERTweet&87.37&LR&BERTweet\\
\rule{0pt}{3.0ex}
S17&92.56&SVM&BERTweet&92.08&SVM&BERTweet\\
\rule{0pt}{3.0ex}
S16&91.03&LR&BERTweet&89.05&LR&BERTweet\\
\noalign{\smallskip}\hline
\end{tabular}
}
\end{table}

For a better understanding of the results, we present an analysis of the difference between the vocabulary embedded in the assessed models. For this purpose, Table~\ref{tab:vocab-similarity} highlights the number of tokens shared between BERT, RoBERTa, and BERTweet. In other words, we show the amount of tokens (in \%) embedded in the models presented in each row that are also included in the models presented in each column, i.e., the intersection between their vocabularies. For example, regarding BERT (first row), we can see that 61\% of its tokens can be found on RoBERTa (second column). The information below each model name in the columns refers to their vocabulary size (number of embedded tokens). It is possible to note that only 32\% of the 64K tokens from BERTweet vocabulary (i.e., about 20K tokens) can be found in BERT. It means that, when compared to BERT, BERTweet contains about 44K ($64-20$) specific tokens extracted from tweets. Similarly, 55\% of the tokens embedded in BERTweet (i.e., about 35K tokens) can be found in RoBERTa, meaning that BERTweet holds about 29K ($64-35$) specific tokens from tweets that are not included in RoBERTa. As a matter of fact, analyzing the tokens embedded in BERTweet, we find some specific tokens, such as ``KKK'', ``Awww'', ``hahaha'', ``broo'', and other internet expressions and slang that social media users often use to express themselves. While creating representations for these tokens is straightforward in BERTweet, BERT and RoBERTa need to do some extras steps. Specifically, when BERT and RoBERTa do not find a token in their vocabularies, they split the token into subtokens until all of them are found. For example, the token ``KKK'' would be split into ``K'', ``K'', and ``K'' to represent the original token. This analysis points out that this particular vocabulary, combined with a language model that was trained focused on learning the intrinsic structure of tweets, is the responsible for the BERTweet language model's best performance on tweet sentiment classification.

%To understand vocabulary differences, we conducted a paired comparison of language models vocabulary. We measure the proportion of the intersection set of tokens between each language model vocabulary. By Table~\ref{tab:vocab-similarity}, comparing BERTweet and BERT, we can conclude that BERTweet's vocabulary has almost 43K (68\% of total tokens) new tokens constructed directed from tweet environment. Comparing with RoBERTa, BERTweet has nearly 30K (55\% of the total) news and specific tweet tokens. Analyzing BERTweet's distinct vocabulary, we can find specific informal tokens that there is not in the other Transformer-based model, such as "KKK", "Awww", "hahaha", "broo", and other informal tokens that are common in social media text style. To build embeddings for this expression, BERT and RoBERTa, for example, have to split them and combine different embeddings. This process may cause semantic information loss and damage the language model's embedding performance, as observed in the above experiment results. So, by this analysis, we believe that this particular vocabulary, combined with a language model focused on learning the intrinsic structure of tweets, is responsible for the BERTweet language model's best performance on tweet sentiment classification.

\begin{table}
\caption{Percentage of vocabulary's tokens of language model in the line that is also in the vocabulary's tokens of language model in the column.
}
\label{tab:vocab-similarity}
%\scalebox{0.85}{
\begin{tabular}{l@{\hskip 0.3in}l@{\hskip 0.3in}l@{\hskip 0.3in}l}
\noalign{\smallskip}\hline\noalign{\smallskip}
\multirow{2}{*}{} & \textbf{BERT}&\textbf{RoBERTa}& \textbf{BERTweet}\\
\noalign{\smallskip}
&$\bm{|V|}$\textbf{ = 30K} & $\bm{|V|}$\textbf{ = 50K} & $\bm{|V|}$\textbf{ = 64K}\\
\noalign{\smallskip}\hline\noalign{\smallskip}
\rule{0pt}{1.8ex}
\textbf{BERT} & $-$ &61 & 62\\
\rule{0pt}{3.8ex}
\textbf{RoBERTa}    & 41 &$-$ & 71\\
\rule{0pt}{3.8ex}
\textbf{BERTweet}& 32 &55 & $-$\\
\noalign{\smallskip}\hline
\end{tabular}
%}
\end{table}

In this context, regarding RQ2, we believe BERTweet is an effective language modeling technique in distinguishing the sentiment expressed in tweets. Also, regarding the classifiers, in general, MLP and LR seem to be good choices when using Transformer-based models. %We were also able to realize this language model's good performance when combined with MLP and LR. Considering the entire classification system (Embeddings + classifier), we suggest analyze more deep the classification hyperparameters optimization because this is not part of the study objective.

%%%%%%%%%%%%%%%%%%%%%%%%%%%%%%%%%%%%%%%%%%%%%%%%%%%%%%%%%%%%%%%%%%%%%%%

%\subsection{Comparison between static and Transformer-based representations}

Different from static representation, when we used only the embedding layer of the 13 language models, in this section, we use the whole language model: the tweet goes from the embedding layer up to the last layer to be transformed in a vector representation. Attempting to understand the benefits from using the whole language model (embedding layer and language model), we compare the predictive performance of Transformer-based models evaluated in this section against all the static representations assessed in Section~\ref{sec:static-exp}. Table~\ref{tab:top10_ranksum_context_static} presents the top ten results across all 85 possible combinations of models and classifiers (17 models $\times$ 5 classification algorithms), and Table~\ref{tab:emb_ranksum_context_static} shows an overall evaluation of the models, from best to worst,in terms of the average rank position. In addition, Table~\ref{tab:bestoverall-static-context} shows the best results achieved for each dataset. %We can notice that Transformer-Autoencoder BERTweet outperformed static embedding in all analysis. In table~\ref{tab:emb_ranksum_context_static}, for example, we notice that Transformer-based autoencoders build the top 3, having a considerable difference for the first static embedding in the top 10.

From Tables~\ref{tab:top10_ranksum_context_static} and~\ref{tab:emb_ranksum_context_static} we can notice that the Transformer-based BERTweet model outperformed all other models and was ranked first in both evaluations. Also, Table~\ref{tab:emb_ranksum_context_static} shows that the Transformed-based models achieved the best overall results against all static models and were ranked as the top three best representations. Furthermore, from Table~\ref{tab:bestoverall-static-context}, the Transformer-based BERTweet model achieved the best overall classification effectiveness in 16 out of the 22 datasets in terms of accuracy and in 17 out of the 22 datasets in terms of $F_1$-macro.

These results point out that learning language model parameters is essential in distinguishing the sentiment expressed in tweets. Static representations may lose a lot of relevant information considering they ignore the diversity of meaning that words may have depending on the context they appear. In contrast, Transformer-based models benefit from learning how to encode the context information of a token in an embedding.
%%%%%%%%%%%%%%%%%%%%%%%%%%%%%%%%%%%%%%%%%%%%%%%%%%%%%%%%%%%%%%%%%%%%%%%%%
\begin{table}%[!htbp]
\caption{Top 10 average rank position results achieved for each combination Model-Classifier by evaluating Transformer-Autoencoder model and static embeddings}
\label{tab:top10_ranksum_context_static}
\scalebox{0.85}{
\begin{tabular}{lll|lll}
\hline\noalign{\smallskip}
\multirow{2}{*}{\textbf{Model}} & \multirow{2}{*}{\textbf{Classifier}} & \textbf{Accuracy} & \multirow{2}{*}{\textbf{Model}} & \multirow{2}{*}{\textbf{Classifier}} & \textbf{$\bm{F_1}$-macro} \\
& & \textbf{avg. rank pos.} & & & \textbf{avg. rank pos.}\\
\noalign{\smallskip}\hline\noalign{\smallskip}
\rule{0pt}{3ex}
BERTweet&LR&6.20&BERTweet&LR&5.91\\
\rule{0pt}{3.0ex}
BERTweet&MLP&6.32&BERTweet&MLP&6.55\\
\rule{0pt}{3.0ex}
RoBERTa&MLP&10.27&BERTweet&SVM&9.14\\
\rule{0pt}{3.0ex}
BERT&MLP&10.43&BERT&SVM&10.00\\
\rule{0pt}{3.0ex}
BERTweet&SVM&10.45&RoBERTa&MLP&10.89\\
\rule{0pt}{3.0ex}
RoBERTa&LR&12.23&RoBERTa&LR&10.91\\
\rule{0pt}{3.0ex}
BERT&LR&12.91&BERT&MLP&10.91\\
\rule{0pt}{3.0ex}
BERT&SVM&13.11&BERT&LR&12.07\\
\rule{0pt}{3.0ex}
BERTweet&XGB&17.02&RoBERTa-static&SVM&17.32\\
\rule{0pt}{3.0ex}
RoBERTa-static&SVM&18.66&W2V-Edin&SVM&18.75\\

\noalign{\smallskip}
\hline
\end{tabular}
}
\end{table}

%%%%%%%%%%%%%%%%%%%%%%%%%%%%%%%%%%%%%%%%%%%%%%%%%%%%%%%%%%%%%%%%%%%%%%%%%
\begin{table}%[!htbp]
\caption{Average rank position results achieved for each Embedding evaluating Transformer-Autoencoder model and static embedding}
\label{tab:emb_ranksum_context_static}
%\scalebox{0.8}{
\begin{tabular}{ll|ll}
\hline\noalign{\smallskip}
\multirow{2}{*}{\textbf{Model}}& \textbf{Accuracy} & \multirow{2}{*}{\textbf{Model}}& \textbf{$\bm{F_1}$-macro} \\
& \textbf{avg. rank pos.} & & \textbf{avg. rank pos.}\\
\noalign{\smallskip}\hline\noalign{\smallskip}
\rule{0pt}{3ex}
BERTweet&14.84&BERTweet&17.38\\
\rule{0pt}{3.0ex}
BERT&20.86&BERT&23.42\\
\rule{0pt}{3.0ex}
RoBERTa&23.00&RoBERTa&24.94\\
\rule{0pt}{3.0ex}
Emo2Vec&37.49&Emo2Vec&36.09\\
\rule{0pt}{3.0ex}
RoBERTa-static&39.65&RoBERTa-static&40.37\\
\rule{0pt}{3.0ex}
W2V-Edin&41.50&W2V-Edin&41.54\\
\rule{0pt}{3.0ex}
W2V-GN&42.37&2V-GN&43.02\\
\rule{0pt}{3.0ex}
GloVe-TWT&44.64&GloVe-TWT&43.55\\
\rule{0pt}{3.0ex}
SSWE&46.75&SSWE&44.78\\
\rule{0pt}{3.0ex}
EWE&47.21&EWE&46.17\\
\rule{0pt}{3.0ex}
DeepMoji&48.35&DeepMoji&46.79\\
\rule{0pt}{3.0ex}
TF-IDF&49.09&fastText&48.95\\
\rule{0pt}{3.0ex}
fastText&50.18&GloVe-WP&51.55\\
\rule{0pt}{3.0ex}
BERT-static&52.30&BERT-static&51.67\\
\rule{0pt}{3.0ex}
GloVe-WP&53.01&TF-IDF&53.00\\
\rule{0pt}{3.0ex}
W2V-Araque&56.25&W2V-Araque&55.45\\
\rule{0pt}{3.0ex}
BERTweet-static&63.52&BERTweet-static&62.34\\
\noalign{\smallskip}
\hline
\end{tabular}
%}
\end{table}

%%%%%%%%%%%%%%%%%%%%%%%%%%%%%%%%%%%%%%%%%%%%%%%%%%%%%%%%%%%%%%%%%%%%%%%%%

\begin{table}%[!htbp]
\caption{Best results achieved for each dataset by evaluating Transformer-Autoencoder and static models}
\label{tab:bestoverall-static-context}
\scalebox{0.9}{
\begin{tabular}{llll|lll}
\hline\noalign{\smallskip}
\textbf{Dataset}&\textbf{Accuracy}& \textbf{Classifier} &\textbf{Model} & \bm{$F_1$}\textbf{-macro}& \textbf{Classifier} &\textbf{Model} \\
\noalign{\smallskip}\hline\noalign{\smallskip}
\rule{0pt}{0ex}
iro&80.48&LR&BERT&75.87&LR&Emo2Vec\\
\rule{0pt}{3.0ex}
sar&87.50&LR&SSWE&87.19&LR&SSWE\\
\rule{0pt}{3.0ex}
ntu&95.30&MLP&w2v-Edin&95.19&MLP&w2v-Edin\\
\rule{0pt}{3.0ex}
S15&92.23&LR&BERTweet&83.33&LR&BERTweet\\
\rule{0pt}{3.0ex}
stm&90.79&LR&RoBERTa&90.75&LR&RoBERTa\\
\rule{0pt}{3.0ex}
per&87.69&LR&BERTweet&85.29&LR&BERTweet\\
\rule{0pt}{3.0ex}
hob&94.82&MLP&BERT-static&94.05&MLP&BERT-static\\
\rule{0pt}{3.0ex}
iph&87.59&MLP&BERTweet&84.72&MLP&BERTweet\\
\rule{0pt}{3.0ex}
mov&89.47&MLP&RoBERTa&82.12&LR&BERTweet\\
\rule{0pt}{3.0ex}
san&91.17&MLP&BERTweet&91.11&MLP&BERTweet\\
\rule{0pt}{3.0ex}
Nar&95.60&MLP&BERTweet&95.43&MLP&BERTweet\\
\rule{0pt}{3.0ex}
arc&90.74&MLP&BERTweet&90.51&MLP&BERTweet\\
\rule{0pt}{3.0ex}
S18&88.97&SVM&BERTweet&88.87&SVM&BERTweet\\
\rule{0pt}{3.0ex}
OMD&87.36&SVM&BERTweet&86.40&SVM&BERTweet\\
\rule{0pt}{3.0ex}
HCR&81.55&XGB&BERTweet&76.77&LR&BERTweet\\
\rule{0pt}{3.0ex}
STS&93.90&LR&BERTweet&92.98&LR&BERTweet\\
\rule{0pt}{3.0ex}
SST&86.76&SVM&BERTweet&86.53&SVM&BERTweet\\
\rule{0pt}{3.0ex}
Tar&86.93&SVM&BERTweet&86.92&SVM&BERTweet\\
\rule{0pt}{3.0ex}
Vad&90.80&LR&BERTweet&89.38&LR&BERTweet\\
\rule{0pt}{3.0ex}
S13&89.61&LR&BERTweet&87.37&LR&BERTweet\\
\rule{0pt}{3.0ex}
S17&92.56&SVM&BERTweet&92.08&SVM&BERTweet\\
\rule{0pt}{3.0ex}
S16&91.03&LR&BERTweet&89.05&LR&BERTweet\\

\noalign{\smallskip}
\hline
\end{tabular}
}
\end{table}

\section{Fine-tuning Transformer-based models using a large collection of English tweets}
\label{sec:finetuning-exp}

In this section, we aim at performing computational experiments in order to answer the research question RQ3, stated as follows:

\textit{RQ3. Can the fine-tuning of Transformer-based autoencoder models using a large set of English tweets improve the sentiment classification performance?}

To answer this research question, we evaluate the classification effectiveness of BERT, RoBERTa, and BERTweet language models fine-tuned with tweets from a corpus of 6.7M unlabeled, or generic unlabeled, tweets, as described in Section~\ref{sec:comp-exp}. Precisely, we use this set of tweets to fine tuning the model weights using the intermediate masked language model task as the training objective with the probability of 15\% to (randomly) mask tokens in the input.  We also compare the fine-tuning results of such models against those achieved by using the original weights of the Transformer-based models, as presented in Section~\ref{sec:context}, in order to analyze whether the adjustment of the models via fine-tuning improves the predictive performance of the sentiment classification.

In general, the performance of the fine-tuned models is very sensitive to different random seeds~\cite{DBLP:journals/corr/abs-2002-06305}. For that reason, all the results presented in this section are the average of three executions using different seeds (12,34,56) to account for the sensitivity of the fine-tuning process regarding different seeds~\cite{dodge2020fine}.

The first part of the experiments reported in this section consists in determining whether the predictive performance of the Transformer-based models are affected by the fine-tuning procedure using tweets from corpora of different sizes. For this purpose, in addition to the entire Edinburgh corpus of 6,657,700 tweets (around 6.7M tweets), we used nine other smaller samples of tweets with different sizes, varying from 500 to 1.5M tweets. Specifically, we generated samples containing 0.5K, 1K, 5K, 10K, 25K, 50K, 250K, 500K, and 1.5M generic unlabeled tweets. In the fine-tuning processes, we performed three training epochs, except for the tuned models with 6.7M tweets, when we used one epoch, as there was a degradation of some models, such as BERTweet. In all fine-tuning process, all layers are unfrozen. Regarding the batch size, we use the available hardware capacity of eight instances per device. We used a learning rate of 5e-5 with a linear scheduler and Adam optmizer with beta1 equal to 0.9, beta2, 0.999 ,and epsilon, 1e-8. We also use a max gradient of 1 and with no weight decay.

Tables~\ref{tab:ft-svm-results-acc} and~\ref{tab:ft-svm-results-f1} present the average classification accuracies and $F_1$-macro scores, respectively, when fine-tuning the Transformer-based models with the different samples of tweets generated from the Edinburgh corpus. These results were achieved by using the SVM classifier (refer to Online Resource 1 for the detailed evaluation of each classifier). Regarding the variance in performance across the different seeds, the mean and maximum standard deviations are 0.05\% and 0.5\% for both accuracy and $F_1$-macro, respectively.

\begin{table}%[!htbp]
{\caption{Average classification accuracies (\%) achieved by fine-tuning Transformer-based models with different samples of generic unlabeled tweets, using the SVM classifier}
\label{tab:ft-svm-results-acc}
\scalebox{0.85}{
\begin{tabular}{lllllllllll}
\hline\noalign{\smallskip}
\multicolumn{11}{c}{\textbf{BERT}}\\
\hline\noalign{\smallskip}
\textbf{Dataset}& \textbf{0.5K}& \textbf{1K}& \textbf{5K}& \textbf{10K}&\textbf{25K}& \textbf{50K} & \textbf{250K} & \textbf{500K} & \textbf{1.5M} & \textbf{6.7M}\\ 
\noalign{\smallskip}\hline\noalign{\smallskip}
iro&73.81&74.05&72.38&66.19&76.43&73.57&\textbf{76.67}&61.19&62.07&56.74\\
sar&68.93&70.18&74.46&67.32&73.04&\textbf{77.32}&74.46&70.36&71.31&63.69\\
ntu&88.84&86.65&84.18&85.97&83.08&85.62&84.89&92.09&93.88&\textbf{94.38}\\
S15&90.35&89.4&90.02&\textbf{90.64}&88.78&90.03&87.22&87.23&85.96&87.13\\
stm&89.14&89.14&88.29&\textbf{89.41}&89.13&88.86&89.13&88.3&88.57&87.01\\
per&84.72&82.91&\textbf{85.42}&83.14&82.91&82.01&81.1&81.32&79.5&76.16\\
hob&83.9&85.24&86.39&\textbf{86.78}&86.18&84.85&84.29&83.72&84.15&80.89\\
iph&81.02&81.39&81.58&82.15&81.02&81.2&82.15&81.21&80.34&78.96\\
mov&82.72&83.43&84.49&85.2&84.5&85.03&\textbf{85.91}&83.26&82.41&82.71\\
san&86.19&86.19&85.87&87.01&86.27&87.42&\textbf{87.66}&87.0&86.98&87.6\\
Nar&91.93&91.44&92.42&91.52&91.6&91.28&92.5&92.83&\textbf{93.67}&93.5\\
arc&86.85&88.07&88.94&88.19&87.49&88.01&87.95&\textbf{89.58}&88.13&88.5\\
S18&\textbf{87.25}&86.82&86.45&86.45&86.5&86.61&86.5&86.82&87.09&86.0\\
OMD&85.68&\textbf{85.99}&85.26&85.15&84.73&84.31&85.79&84.84&85.43&84.58\\
HCR&78.56&78.82&78.4&78.35&78.35&\textbf{79.5}&78.82&77.98&77.32&76.56\\
STS&90.36&90.51&90.41&90.81&90.22&91.05&91.79&\textbf{92.23}&91.89&91.29\\
SSt&84.58&85.06&\textbf{85.37}&84.58&84.01&83.79&84.97&84.8&84.84&84.43\\
Tar&85.67&85.67&85.93&\textbf{86.01}&85.9&85.84&85.9&85.69&85.69&85.4\\
Vad&89.75&89.66&89.63&89.44&89.56&90.3&90.28&89.51&\textbf{90.5}&90.0\\
S13&86.52&86.66&\textbf{87.64}&87.05&87.39&86.96&87.23&86.98&87.07&86.32\\
S17&91.08&91.11&91.05&\textbf{91.15}&90.61&90.92&90.63&90.18&89.85&89.56\\
S16&88.45&88.34&\textbf{89.24}&88.83&88.83&88.82&88.9&88.29&88.01&87.93\\
\noalign{\smallskip}\hline
\#wins &1& 1& 4& \textbf{5}& 0& 2& 3& 2& 2& 1\\
rank sums &125.5&112.0&101.0&101.5&132.5&117.0&\textbf{88.0}&132.0&129.5&171.0\\
position &6.0&4.0&2.0&3.0&9.0&5.0&\textbf{1.0}&8.0&7.0&10.0\\
\noalign{\smallskip}
\end{tabular}
}}

\scalebox{0.85}{
\begin{tabular}{lllllllllll}
\hline\noalign{\smallskip}
\multicolumn{11}{c}{\textbf{RoBERTa}}\\
\hline\noalign{\smallskip}
\textbf{Dataset}& \textbf{0.5K}& \textbf{1K}& \textbf{5K}& \textbf{10K}&\textbf{25K}& \textbf{50K} & \textbf{250K} & \textbf{500K} & \textbf{1.5M} & \textbf{6.7M}\\ 
\noalign{\smallskip}\hline\noalign{\smallskip}
iro&46.67&46.67&46.67&\textbf{48.1}&46.67&46.67&46.67&46.67&45.24&46.74\\
sar&60.71&59.29&65.0&57.86&62.14&60.71&65.0&62.14&63.57&64.7\\
ntu&81.69&81.31&79.15&81.34&82.06&85.63&87.08&90.67&\textbf{91.75}&90.41\\
S15&83.18&83.5&84.75&86.93&84.75&85.38&86.93&\textbf{88.79}&87.24&87.03\\
stm&85.23&86.35&89.7&85.24&87.75&88.04&88.03&87.48&\textbf{90.81}&86.35\\
per&69.92&69.68&70.38&68.33&71.74&71.28&72.66&74.26&76.32&\textbf{77.0}\\
hob&73.76&71.44&73.56&74.32&72.41&77.4&77.38&77.2&\textbf{77.97}&77.82\\
iph&78.41&78.41&77.46&78.96&78.78&79.71&79.89&\textbf{80.09}&79.15&79.02\\
mov&71.14&76.84&\textbf{81.64}&80.76&78.44&78.09&79.52&80.4&81.12&81.05\\
san&84.39&84.23&85.21&85.78&85.13&86.27&85.86&87.17&\textbf{88.23}&86.16\\
Nar&91.04&92.18&92.58&92.25&91.2&92.58&92.74&92.83&93.07&\textbf{93.88}\\
arc&88.88&88.53&89.0&87.72&88.48&89.23&88.59&\textbf{89.64}&89.47&88.57\\
S18&\textbf{87.9}&87.15&87.15&87.09&86.61&87.52&87.14&87.47&86.61&86.62\\
OMD&82.79&83.58&83.63&\textbf{84.0}&83.53&83.47&83.79&82.27&82.74&82.55\\
HCR&76.52&77.25&77.2&76.73&76.1&\textbf{77.78}&76.41&77.77&75.94&77.74\\
STS&89.62&90.71&91.39&90.95&91.84&92.08&92.08&92.52&\textbf{92.92}&92.0\\
SSt&84.67&\textbf{86.28}&85.93&85.98&85.67&86.06&85.8&86.02&86.02&85.22\\
Tar&85.43&85.67&\textbf{86.36}&85.78&85.95&85.41&85.52&85.23&85.98&84.33\\
Vad&88.25&89.56&89.82&89.73&89.3&89.63&89.85&90.4&\textbf{91.11}&89.87\\
S13&85.59&85.54&86.18&86.57&85.68&85.93&86.23&85.54&\textbf{87.03}&85.89\\
S17&90.77&91.0&\textbf{91.37}&91.29&91.07&91.15&91.02&90.86&90.63&89.68\\
S16&88.37&88.4&88.85&89.01&88.96&88.26&88.74&88.96&89.01&87.95\\
\noalign{\smallskip}\hline
\#wins &1& 1& 3& 2& 0& 1& 0& 3& \textbf{7}& 2\\
rank sums &174.0&161.0&111.0&124.0&148.0&105.5&101.5&92.0&\textbf{78.5}&114.5\\
position &10.0&9.0&5.0&7.0&8.0&4.0&3.0&2.0&\textbf{1.0}&6.0\\
\noalign{\smallskip}
\end{tabular}
}

\scalebox{0.85}{
\begin{tabular}{lllllllllll}
\hline\noalign{\smallskip}
\multicolumn{11}{c}{\textbf{BERTweet}}\\
\hline\noalign{\smallskip}
\textbf{Dataset}& \textbf{0.5K}& \textbf{1K}& \textbf{5K}& \textbf{10K}&\textbf{25K}& \textbf{50K} & \textbf{250K} & \textbf{500K} & \textbf{1.5M} & \textbf{6.7M}\\ 
\noalign{\smallskip}\hline\noalign{\smallskip}
iro&68.81&68.1&71.19&73.81&74.76&\textbf{79.76}&71.67&70.48&67.78&66.43\\
sar&64.82&69.11&67.68&73.21&73.21&68.93&70.36&70.36&72.26&71.79\\
ntu&89.95&91.4&91.03&92.82&93.17&92.09&93.16&93.16&\textbf{94.38}&92.45\\
S15&90.34&92.22&90.04&\textbf{92.53}&92.23&91.6&90.03&90.04&89.4&89.09\\
stm&90.52&91.92&\textbf{93.04}&91.36&91.07&91.9&92.19&93.02&91.45&90.52\\
per&84.28&84.04&85.64&86.77&\textbf{86.79}&85.41&86.1&85.87&83.14&81.32\\
hob&85.06&85.25&86.78&86.78&86.78&\textbf{87.92}&86.6&86.77&87.03&86.21\\
iph&84.04&83.84&83.47&85.35&84.78&84.22&83.47&\textbf{85.55}&82.22&82.34\\
mov&85.93&87.71&88.6&88.95&88.42&\textbf{89.49}&88.78&88.42&87.3&87.89\\
san&90.03&90.69&91.42&90.93&91.18&\textbf{91.67}&91.01&91.01&89.46&90.03\\
Nar&95.93&96.33&96.58&96.41&96.58&96.41&96.58&95.92&95.71&94.87\\
arc&90.05&90.39&90.63&90.34&91.09&90.45&\textbf{91.21}&90.98&91.12&90.74\\
S18&89.99&89.78&90.26&\textbf{90.48}&89.89&89.83&89.94&89.03&88.26&87.9\\
OMD&88.09&88.93&\textbf{88.99}&87.67&88.25&88.51&88.88&88.09&87.84&86.88\\
HCR&80.24&80.5&81.23&81.18&\textbf{81.24}&80.6&80.61&78.98&78.25&78.09\\
STS&94.35&\textbf{95.23}&95.08&94.99&94.84&95.13&94.49&94.15&93.97&94.3\\
SSt&87.64&88.16&88.73&\textbf{89.6}&88.99&88.03&87.51&87.59&86.98&87.68\\
Tar&87.02&\textbf{87.83}&87.68&87.54&87.68&87.6&87.28&86.79&86.38&85.84\\
Vad&90.8&92.42&92.56&92.66&92.23&92.59&\textbf{92.71}&92.18&92.11&92.09\\
S13&89.4&90.0&\textbf{90.25}&89.84&90.0&88.88&89.38&88.85&88.76&88.44\\
S17&92.75&93.37&93.35&\textbf{93.49}&93.4&92.97&92.75&92.01&91.31&91.16\\
S16&90.7&91.38&\textbf{91.98}&91.35&91.54&91.36&91.18&90.6&90.38&89.55\\
\noalign{\smallskip}\hline
\#wins &0& 2& \textbf{4}& \textbf{4}& 2& \textbf{4}& 2& 1& 1& 0\\
rank sums &165.0&116.5&84.5&82.0&\textbf{70.0}&94.5&103.5&135.0&169.0&190.0\\
position &8.0&6.0&3.0&2.0&\textbf{1.0}&4.0&5.0&7.0&9.0&10.0\\
\noalign{\smallskip}\hline
\end{tabular}
}
\end{table}

\begin{table}
{\caption{Average $F_1$-macro scores (\%) achieved by fine-tuning Transformer-Autoencoder models with different samples of unlabeled tweets, using the SVM classifier}
\label{tab:ft-svm-results-f1}
\scalebox{0.85}{
\begin{tabular}{lllllllllll}
\hline\noalign{\smallskip}
\multicolumn{11}{c}{\textbf{BERT}}\\
\hline\noalign{\smallskip}
\textbf{Dataset}& \textbf{0.5K}& \textbf{1K}& \textbf{5K}& \textbf{10K}&\textbf{25K}& \textbf{50K} & \textbf{250K} & \textbf{500K} & \textbf{1.5M} & \textbf{6.7M}\\ 
\noalign{\smallskip}\hline\noalign{\smallskip}
iro&\textbf{71.8}&59.35&57.57&58.71&55.4&58.55&62.92&49.72&55.0&49.95\\
sar&56.06&49.58&59.97&58.0&64.49&\textbf{68.15}&57.05&56.54&66.52&57.29\\
ntu&82.27&83.73&80.3&80.78&80.5&80.67&83.24&\textbf{90.02}&86.99&89.62\\
S15&65.69&62.12&60.62&65.57&63.82&64.1&64.6&61.86&62.76&\textbf{66.05}\\
stm&84.92&83.48&84.66&85.18&81.3&83.72&84.61&83.17&82.8&\textbf{86.29}\\
per&71.38&\textbf{77.86}&76.33&75.95&76.83&75.75&76.14&72.98&71.0&71.03\\
hob&79.14&80.33&80.63&78.85&\textbf{83.11}&80.32&81.6&80.79&82.09&81.83\\
iph&78.71&80.53&79.09&78.76&\textbf{81.91}&79.35&76.93&79.86&77.63&74.78\\
mov&60.89&60.98&64.29&60.29&57.75&58.49&63.56&66.66&60.49&\textbf{67.71}\\
san&85.75&85.4&84.66&85.03&83.38&85.66&\textbf{86.45}&85.5&85.44&85.46\\
Nar&88.81&88.71&88.52&89.31&88.58&88.79&89.79&90.29&90.81&\textbf{91.36}\\
arc&88.63&87.78&87.81&87.74&88.51&87.85&88.48&\textbf{89.52}&88.76&88.76\\
S18&83.83&\textbf{84.66}&83.8&83.52&83.42&83.31&83.04&83.57&83.31&84.03\\
OMD&\textbf{82.77}&82.03&80.68&79.83&81.66&79.66&82.59&81.9&81.81&80.56\\
HCR&72.62&\textbf{72.99}&71.85&72.68&72.31&71.46&72.48&71.03&69.89&68.45\\
STS&84.29&85.24&84.42&85.58&84.63&85.35&87.2&87.12&86.67&\textbf{87.81}\\
SSt&82.87&82.27&82.04&81.74&80.77&82.1&\textbf{82.94}&81.86&82.62&81.13\\
Tar&83.18&83.96&\textbf{84.22}&84.19&83.0&83.9&83.64&83.32&83.59&83.71\\
Vad&85.11&84.66&85.11&85.59&84.7&85.31&85.3&85.47&85.83&\textbf{85.85}\\
S13&82.45&82.15&82.67&80.76&81.98&82.26&\textbf{82.84}&82.43&82.35&81.75\\
S17&\textbf{89.12}&88.76&88.98&88.47&88.66&88.69&88.71&87.77&87.73&87.68\\
S16&84.65&84.86&84.59&\textbf{85.0}&84.64&84.63&84.64&84.08&83.75&83.64\\
\noalign{\smallskip}\hline
\#wins &\textbf{3}& \textbf{3}& 1& 1& 2& 1& \textbf{3}& 2& 0& 6 \\
rank sums &105.5&112.0&129.5&126.0&146.5&131.5&\textbf{95.5}&120.0&128.0&115.5\\
position &2.0&3.0&8.0&6.0&10.0&9.0&\textbf{1.0}&5.0&7.0&4.0\\
\noalign{\smallskip}
\end{tabular}
}}

\scalebox{0.85}{
\begin{tabular}{lllllllllll}
\hline\noalign{\smallskip}
\multicolumn{11}{c}{\textbf{RoBERTa}}\\
\hline\noalign{\smallskip}
\textbf{Dataset}& \textbf{0.5K}& \textbf{1K}& \textbf{5K}& \textbf{10K}&\textbf{25K}& \textbf{50K} & \textbf{250K} & \textbf{500K} & \textbf{1.5M} & \textbf{6.7M}\\ 
\noalign{\smallskip}\hline\noalign{\smallskip}
iro&54.77&61.69&47.99&58.14&57.59&\textbf{67.22}&42.02&54.39&53.42&54.93\\
sar&53.42&56.62&59.94&\textbf{73.89}&64.55&54.6&68.13&49.36&50.89&67.67\\
ntu&83.63&83.39&75.86&81.64&79.9&84.13&88.92&88.63&\textbf{91.87}&87.65\\
S15&67.46&63.75&\textbf{74.11}&66.24&62.03&73.59&67.08&69.26&61.61&66.92\\
stm&86.32&86.0&85.21&85.77&\textbf{87.43}&86.61&86.58&84.32&83.77&83.63\\
per&72.78&73.45&75.6&75.1&76.31&74.48&\textbf{77.13}&74.92&74.8&76.17\\
hob&76.79&79.35&77.42&76.73&80.1&79.58&80.82&80.22&78.26&\textbf{82.06}\\
iph&77.2&79.94&80.96&\textbf{82.27}&81.08&79.67&75.66&79.93&78.21&78.86\\
mov&64.06&70.45&\textbf{73.38}&66.84&68.1&66.22&69.41&72.07&68.51&68.91\\
san&86.56&87.22&86.8&\textbf{88.14}&86.99&87.8&86.24&87.12&87.73&86.33\\
Nar&89.57&91.0&\textbf{92.06}&90.61&90.4&91.92&91.4&91.76&91.21&91.42\\
arc&89.53&88.65&89.29&89.06&89.84&89.56&\textbf{90.22}&89.88&88.94&88.84\\
S18&85.46&\textbf{87.06}&85.68&86.27&86.86&86.33&86.7&85.68&84.43&84.61\\
OMD&83.23&83.07&\textbf{84.4}&83.69&83.75&82.83&83.73&81.68&80.76&81.73\\
HCR&71.23&\textbf{74.87}&72.89&72.13&72.09&73.59&72.17&73.06&71.71&70.58\\
STS&87.5&88.8&88.7&88.46&88.84&89.61&\textbf{89.71}&89.14&88.95&88.6\\
SSt&83.38&83.23&84.71&83.78&84.34&84.21&84.43&84.5&\textbf{84.8}&82.36\\
Tar&83.84&84.13&\textbf{85.14}&84.47&85.0&84.51&83.78&83.49&84.79&83.27\\
Vad&85.85&85.72&86.89&86.75&86.44&87.19&88.04&\textbf{88.13}&87.98&86.75\\
S13&82.03&82.75&83.57&83.82&83.63&83.73&\textbf{84.38}&83.54&83.73&82.31\\
S17&89.72&89.98&90.07&90.17&89.69&90.2&\textbf{90.21}&90.01&89.31&87.6\\
S16&85.29&85.99&86.44&86.27&\textbf{86.59}&86.3&86.31&86.06&86.4&84.46\\
\noalign{\smallskip}\hline
\#wins &0& 2& \textbf{5}& 3& 2& 1& \textbf{5}& 1& 2& 1 \\
rank sums &171.0&132.0&100.5&118.5&105.0&92.5&\textbf{87.0}&110.5&136.5&156.5\\
position &10.0&7.0&3.0&6.0&4.0&2.0&\textbf{1.0}&5.0&8.0&9.0\\
\noalign{\smallskip}
\end{tabular}
}

\scalebox{0.85}{
\begin{tabular}{lllllllllll}
\hline\noalign{\smallskip}
\multicolumn{11}{c}{\textbf{BERTweet}}\\
\hline\noalign{\smallskip}
\textbf{Dataset}& \textbf{0.5K}& \textbf{1K}& \textbf{5K}& \textbf{10K}&\textbf{25K}& \textbf{50K} & \textbf{250K} & \textbf{500K} & \textbf{1.5M} & \textbf{6.7M}\\ 
\noalign{\smallskip}\hline\noalign{\smallskip}
iro&54.36&61.9&53.1&49.09&51.4&53.28&62.24&\textbf{66.75}&53.0&42.85\\
sar&49.87&53.96&59.51&53.78&62.33&\textbf{62.79}&54.39&55.69&56.94&55.57\\
ntu&83.16&\textbf{89.22}&85.29&87.38&85.11&87.31&85.77&88.22&87.64&86.64\\
S15&71.56&\textbf{77.95}&73.8&69.48&73.47&70.08&63.77&63.21&61.07&58.37\\
stm&85.2&85.99&86.81&85.72&85.73&\textbf{88.27}&85.71&87.71&87.42&86.28\\
per&\textbf{82.43}&81.37&77.92&81.86&78.86&77.46&77.25&78.55&77.76&74.77\\
hob&77.56&\textbf{84.41}&81.24&81.87&80.0&81.5&79.35&80.79&80.85&78.1\\
iph&81.95&82.07&\textbf{84.28}&80.85&80.71&80.71&81.08&82.04&78.76&81.09\\
mov&71.91&73.18&76.03&75.34&73.91&\textbf{78.82}&70.84&72.74&73.3&70.2\\
san&88.62&89.79&90.55&\textbf{90.87}&88.88&88.46&88.7&88.29&88.19&86.71\\
Nar&91.39&\textbf{94.35}&93.48&93.12&92.72&92.78&92.88&92.83&92.26&91.6\\
arc&89.52&90.66&\textbf{91.59}&90.76&90.81&90.04&90.35&90.4&89.33&90.62\\
S18&87.25&88.09&\textbf{88.26}&87.55&87.99&87.03&87.35&85.32&85.07&84.41\\
OMD&83.01&83.39&\textbf{85.56}&84.26&83.75&83.14&85.0&83.56&82.72&82.69\\
HCR&73.08&72.69&74.28&\textbf{74.31}&73.16&73.97&74.0&73.57&70.16&69.09\\
STS&89.3&90.18&\textbf{90.85}&90.69&90.47&89.82&89.35&89.98&88.91&88.98\\
SSt&84.74&85.88&86.41&\textbf{86.7}&84.54&84.48&83.79&85.15&83.59&84.18\\
Tar&85.13&85.54&\textbf{85.8}&85.31&85.34&84.3&84.21&84.64&83.67&82.8\\
Vad&85.68&87.98&\textbf{88.26}&87.9&87.32&87.14&87.89&87.45&87.55&86.75\\
S13&83.74&84.73&\textbf{85.43}&84.33&83.94&83.48&83.49&83.42&82.51&81.36\\
S17&90.76&91.27&\textbf{91.79}&91.77&91.71&90.88&90.74&90.09&88.76&88.9\\
S16&87.16&88.07&\textbf{88.43}&88.06&87.94&87.26&87.4&86.3&86.11&85.17\\
\noalign{\smallskip}\hline
\#wins &1& 4& \textbf{10}& 3& 0& 3& 0& 1& 0& 0\\
rank sums &153.0&74.0&\textbf{53.0}&83.0&108.5&122.5&136.0&121.0&169.0&190.0\\
position &8.0&2.0&\textbf{1.0}&3.0&4.0&6.0&7.0&5.0&9.0&10.0\\
\noalign{\smallskip}\hline\noalign{\smallskip}
\end{tabular}
}
\end{table}

Note that BERT was most benefited when fine-tuned with samples of 250K tweets (position row), for both accuracy and $F_1$-macro. RoBERTa achieved the best overall results when fine-tuned with samples of 1.5M and 250K tweets, in terms of accuracy and $F_1$-macro, respectively. On the other hand, BERTweet benefited from smaller samples, achieving higher overall predictive performances when fine-tuned with samples of 25K and 5K tweets in terms of accuracy and $F_1$-macro, respectively. This is an expected result as BERTweet is already trained from scratch from tweets. As we are fine-tuning the language model task, BERT and RoBERTa seems to require more samples to accommodate the Twitter-based vocabulary into the weights' model. 

Next, we analyze the overall performance of the fine-tuned Transformer-based models for each classification algorithm. Table~\ref{tab:overview-ft-emb-acc} summarizes the results. Regarding the variance across the different seeds, the mean and maximum standard deviations are 0.2\% and 0.7\% in terms of accuracy, and 0.26\% and 0.98\% in terms of $F_1$-macro.

Interestingly, from Table~\ref{tab:overview-ft-emb-acc}, we can note that when fine-tuning a language model to fit a specific type of text, such as tweets, applying large corpora does not guarantee better predictive performances. Specifically, the best overall results (Total column) were achieved when fine-tuning BERT, RoBERTa, and BERTweet models with samples of 250K, 50K, and 5K tweets, respectively, for both accuracy and $F_1$-macro.

\begin{table}%[!htbp]
\caption{Overview of the results (number of wins, rank sum, and rank position, respectively) achieved by each classifier when fine-tuning the Transformer-Autoencoder models with different samples of unlabeled tweets in terms of accuracy}
\label{tab:overview-ft-emb-acc}
\scalebox{0.80}{
\begin{tabular}{lllllll}
\hline\noalign{\smallskip}
\multicolumn{7}{c}{\textbf{ACCURACY}}\\
\hline\noalign{\smallskip}
\textbf{Sample} & \textbf{LR} & \textbf{SVM} & \textbf{MLP} & \textbf{RF} & \textbf{XGB} &\textbf{Total} \\
\hline\noalign{\smallskip}
\multicolumn{7}{c}{\textbf{BERT}}\\
\hline\noalign{\smallskip}
\rule{0pt}{0ex}
0.5k&3/127.5/5.5&1/125.5/6.0&1/129.0/7.0&1/153.0/10.0&\textbf{5}/108.0/2.0&11/643.0/30.5\\
\rule{0pt}{2.5ex}
1k&0/134.0/7.0&1/112.0/4.0&1/139.5/8.0&3/114.0/3.5&2/113.5/3.0&7/613.0/25.5\\
\rule{0pt}{2.5ex}
5k&3/115.0/4.0&4/101.0/2.0&3/114.5/5.0&1/120.5/6.0&1/128.0/8.0&12/579.0/25.0\\
\rule{0pt}{2.5ex}
10k&0/143.5/9.0&\textbf{5}/101.5/3.0&1/142.0/10.0&\textbf{4}/119.0/5.0&2/125.5/7.0&12/631.5/34.0\\
\rule{0pt}{2.5ex}
25k&0/136.0/8.0&0/132.5/9.0&0/141.0/9.0&1/134.0/8.0&2/146.0/10.0&3/689.5/44.0\\
\rule{0pt}{2.5ex}
50k&0/146.0/10.0&2/117.0/5.0&2/121.5/6.0&3/113.5/2.0&0/131.5/9.0&7/629.5/32.0\\
\rule{0pt}{2.5ex}
250k&0/127.5/5.5&3/\textbf{88.0}/\textbf{1.0}&3/\textbf{101.5}/\textbf{1.0}&2/\textbf{73.5}/\textbf{1.0}&2/\textbf{97.5}/\textbf{1.0}&10/\textbf{488.0}/\textbf{9.5}\\
\rule{0pt}{2.5ex}
500k&1/110.5/3.0&2/132.0/8.0&2/108.0/4.0&1/114.0/3.5&2/119.5/5.0&8/584.0/23.5\\
\rule{0pt}{2.5ex}
1.5M&4/96.0/2.0&2/129.5/7.0&3/107.0/3.0&1/131.5/7.0&1/122.0/6.0&11/586.0/25.0\\
\rule{0pt}{2.5ex}
6.7M&\textbf{10}/\textbf{74.0}/\textbf{1.0}&1/171.0/10.0&\textbf{6}/106.0/2.0&\textbf{4}/137.0/9.0&\textbf{5}/118.5/4.0&\textbf{26}/606.5/26.0\\

\hline\noalign{\smallskip}
\multicolumn{7}{c}{\textbf{RoBERTa}}\\
\hline\noalign{\smallskip}
\rule{0pt}{1ex}
0.5k&1/140.0/9.0&1/174.0/10.0&0/165.5/9.0&0/171.5/9.0&0/173.0/10.0&2/824.0/47.0\\
\rule{0pt}{2.5ex}
1k&2/137.0/8.0&1/161.0/9.0&2/143.0/8.0&0/165.0/8.0&2/130.5/7.0&7/736.5/40.0\\
\rule{0pt}{2.5ex}
5k&3/\textbf{92.0}/\textbf{1.0}&3/111.0/5.0&0/99.5/3.5&4/104.0/5.0&4/100.0/\textbf{4.0}&14/506.5/18.5\\
\rule{0pt}{2.5ex}
10k&1/125.0/7.0&2/124.0/7.0&\textbf{4}/111.5/6.0&1/120.0/7.0&3/118.5/6.0&11/599.0/33.0\\
\rule{0pt}{2.5ex}
25k&\textbf{4}/103.0/3.0&0/148.0/8.0&2/107.5/5.0&1/104.5/6.0&2/98.0/3.0&9/561.0/25.0\\
\rule{0pt}{2.5ex}
50k&\textbf{4}/100.5/2.0&1/105.5/4.0&3/\textbf{85.5}/\textbf{1.0}&4/85.0/2.0&1/97.0/2.0&13/\textbf{473.5}/\textbf{11.0}\\
\rule{0pt}{2.5ex}
250k&0/124.0/6.0&0/101.5/3.0&3/131.5/7.0&\textbf{8}/\textbf{77.5}/\textbf{1.0}&\textbf{4}/\textbf{88.5}/\textbf{1.0}&15/523.0/18.0\\
\rule{0pt}{2.5ex}
500k&3/113.5/5.0&3/92.0/2.0&2/99.5/3.5&3/100.5/4.0&2/109.5/5.0&13/515.0/19.5\\
\rule{0pt}{2.5ex}
1.5M&3/109.0/4.0&\textbf{7}/\textbf{78.5}/\textbf{1.0}&3/91.5/2.0&1/98.5/3.0&2/136.0/8.0&\textbf{16}/513.5/18.0\\
\rule{0pt}{2.5ex}
6.7M&0/166.0/10.0&2/114.5/6.0&2/175.0/10.0&0/183.5/10.0&1/159.0/9.0&5/798.0/45.0\\

\hline\noalign{\smallskip}
\multicolumn{7}{c}{\textbf{BERTweet}}\\
\hline\noalign{\smallskip}
\rule{0pt}{1ex}
0.5k&1/143.0/7.0&0/165.0/8.0&0/167.0/9.0&0/174.5/8.0&0/152.5/8.0&1/802.0/40.0\\
\rule{0pt}{2.5ex}
1k&\textbf{5}/78.5/2.0&2/116.5/6.0&3/95.0/4.0&0/99.0/4.0&4/76.5/2.0&14/465.5/18.0\\
\rule{0pt}{2.5ex}
5k&\textbf{5}/\textbf{69.5}/\textbf{1.0}&\textbf{4}/84.5/3.0&3/\textbf{75.5}/\textbf{1.0}&\textbf{10}/\textbf{42.5}/\textbf{1.0}&\textbf{10}/\textbf{56.0}/\textbf{1.0}&\textbf{32}/\textbf{328.0}/\textbf{7.0}\\
\rule{0pt}{2.5ex}
10k&2/92.0/3.0&\textbf{4}/82.0/2.0&4/95.5/5.0&2/72.5/3.0&4/81.0/3.0&16/423.0/16.0\\
\rule{0pt}{2.5ex}
25k&1/95.0/4.0&2/\textbf{70.0}/\textbf{1.0}&3/78.5/2.0&5/71.0/2.0&0/112.0/4.0&11/426.5/13.0\\
\rule{0pt}{2.5ex}
50k&2/110.0/5.0&4/94.5/4.0&\textbf{6}/91.0/3.0&2/117.0/5.5&3/119.5/6.0&17/532.0/23.5\\
\rule{0pt}{2.5ex}
250k&2/114.5/6.0&2/103.5/5.0&0/128.0/6.0&0/117.0/5.5&0/138.0/7.0&4/601.0/29.5\\
\rule{0pt}{2.5ex}
500k&0/162.0/8.0&1/135.0/7.0&0/150.0/7.0&2/138.0/7.0&1/117.0/5.0&4/702.0/34.0\\
\rule{0pt}{2.5ex}
1.5M&0/172.5/9.0&1/169.0/9.0&0/174.0/10.0&0/176.0/9.0&0/171.0/9.0&1/862.5/46.0\\
\rule{0pt}{2.5ex}
6.7M&1/173.0/10.0&0/190.0/10.0&3/155.5/8.0&0/202.5/10.0&0/186.5/10.0&4/907.5/48.0\\
\noalign{\smallskip}\hline\noalign{\smallskip}
\multicolumn{7}{c}{\bm{$F_1$}\textbf{-MACRO}}\\
\hline\noalign{\smallskip}
\textbf{Sample} & \textbf{LR} & \textbf{SVM} & \textbf{MLP} & \textbf{RF} & \textbf{XGB} &\textbf{Total} \\
\hline\noalign{\smallskip}
\multicolumn{7}{c}{\textbf{BERT}}\\
\hline\noalign{\smallskip}
\rule{0pt}{0ex}
0.5k&3/128.0/6.0&1/127.0/6.0&1/132.0/7.0&0/155.0/10.0&3/105.5/2.0&8/647.5/31.0\\
\rule{0pt}{2.5ex}
1k&1/140.5/8.0&1/113.0/4.0&1/141.0/9.0&3/113.5/3.0&3/112.0/3.0&9/620.0/27.0\\
\rule{0pt}{2.5ex}
5k&2/120.0/4.0&4/99.5/2.0&3/115.5/5.0&1/118.5/5.0&1/129.5/8.0&11/583.0/24.0\\
\rule{0pt}{2.5ex}
10k&0/144.5/9.0&\textbf{5}/104.5/3.0&1/146.5/10.0&1/124.5/6.0&1/126.0/6.0&8/646.0/34.0\\
\rule{0pt}{2.5ex}
25k&0/140.0/7.0&0/131.0/8.0&0/139.0/8.0&1/136.5/8.0&2/146.5/10.0&3/693.0/41.0\\
\rule{0pt}{2.5ex}
50k&0/148.5/10.0&2/115.0/5.0&2/119.0/6.0&4/106.0/2.0&1/131.5/9.0&9/620.0/32.0\\
\rule{0pt}{2.5ex}
250k&0/122.0/5.0&4/\textbf{92.0}/\textbf{1.0}&3/\textbf{96.5}/\textbf{1.0}&\textbf{6}/\textbf{68.0}/\textbf{1.0}&3/\textbf{95.5}/\textbf{1.0}&16/\textbf{474.0}/\textbf{9.0}\\
\rule{0pt}{2.5ex}
500k&1/108.0/3.0&2/132.0/9.0&2/113.5/4.0&1/116.0/4.0&2/120.0/5.0&8/589.5/25.0\\
\rule{0pt}{2.5ex}
1.5M&4/87.5/2.0&2/129.5/7.0&3/104.0/3.0&1/135.0/7.0&0/128.0/7.0&10/584.0/26.0\\
\rule{0pt}{2.5ex}
6.7M&\textbf{11}/\textbf{71.0}/\textbf{1.0}&1/166.5/10.0&\textbf{6}/103.0/2.0&3/137.0/9.0&\textbf{6}/115.5/4.0&\textbf{27}/593.0/26.0\\

\hline\noalign{\smallskip}
\multicolumn{7}{c}{\textbf{RoBERTa}}\\
\hline\noalign{\smallskip}
\rule{0pt}{0ex}

0.5k&1/140.0/9.0&1/175.0/10.0&0/163.0/9.0&0/170.5/9.0&0/171.0/10.0&2/819.5/47.0\\
\rule{0pt}{2.5ex}
1k&3/132.5/8.0&1/160.0/9.0&2/142.0/8.0&0/165.0/8.0&2/132.0/7.0&8/731.5/40.0\\
\rule{0pt}{2.5ex}
5k&3/\textbf{90.0}/\textbf{1.0}&3/114.0/5.0&0/102.5/3.0&4/105.5/5.0&\textbf{5}/100.5/3.0&15/512.5/17.0\\
\rule{0pt}{2.5ex}
10k&1/125.5/6.0&1/129.0/7.0&3/112.0/6.0&1/118.5/7.0&3/118.5/6.0&9/603.5/32.0\\
\rule{0pt}{2.5ex}
25k&3/103.5/3.0&0/151.0/8.0&2/109.0/5.0&1/108.5/6.0&2/105.0/4.0&8/577.0/26.0\\
\rule{0pt}{2.5ex}
50k&\textbf{4}/99.0/2.0&1/106.5/4.0&\textbf{4}/\textbf{85.5}/\textbf{1.0}&4/83.0/2.0&1/92.5/2.0&14/\textbf{466.5}/\textbf{11.0}\\
\rule{0pt}{2.5ex}
250k&0/128.0/7.0&0/98.5/3.0&3/126.0/7.0&\textbf{8}/\textbf{71.0}/\textbf{1.0}&\textbf{5}/\textbf{87.0}/\textbf{1.0}&16/510.5/19.0\\
\rule{0pt}{2.5ex}
500k&2/115.0/5.0&3/88.0/2.0&1/104.0/4.0&4/102.0/4.0&1/110.5/5.0&11/519.5/20.0\\
\rule{0pt}{2.5ex}
1.5M&\textbf{4}/108.5/4.0&\textbf{8}/\textbf{72.0}/\textbf{1.0}&\textbf{4}/87.0/2.0&0/101.0/3.0&2/136.5/8.0&\textbf{18}/505.0/18.0\\
\rule{0pt}{2.5ex}
6.7M&0/168.0/10.0&4/116.0/6.0&2/179.0/10.0&0/185.0/10.0&1/156.5/9.0&7/804.5/45.0\\

\hline\noalign{\smallskip}
\multicolumn{7}{c}{\textbf{BERTweet}}\\
\hline\noalign{\smallskip}
\rule{0pt}{0ex}

0.5k&1/142.0/7.0&0/166.5/8.0&0/169.0/9.0&0/174.0/8.0&1/153.0/8.0&2/804.5/40.0\\
\rule{0pt}{2.5ex}
1k&\textbf{7}/79.5/2.0&2/112.0/6.0&3/89.0/3.0&0/99.5/4.0&4/74.0/2.0&16/454.0/17.0\\
\rule{0pt}{2.5ex}
5k&4/\textbf{71.0}/\textbf{1.0}&\textbf{5}/80.0/3.0&3/\textbf{74.0}/\textbf{1.0}&\textbf{12}/\textbf{39.5}/\textbf{1.0}&\textbf{10}/\textbf{53.0}/\textbf{1.0}&\textbf{34}/\textbf{317.5}/\textbf{7.0}\\
\rule{0pt}{2.5ex}
10k&3/89.5/3.0&\textbf{5}/79.0/2.0&4/95.0/5.0&1/76.5/3.0&3/83.0/3.0&16/423.0/16.0\\
\rule{0pt}{2.5ex}
25k&1/94.0/4.0&1/\textbf{73.5}/\textbf{1.0}&3/77.0/2.0&5/71.0/2.0&0/108.5/4.0&10/424.0/13.0\\
\rule{0pt}{2.5ex}
50k&2/110.5/5.0&4/96.5/4.0&6/94.0/4.0&2/117.0/\textbf{6.0}&3/122.5/6.0&17/540.5/25.0\\
\rule{0pt}{2.5ex}
250k&2/116.5/6.0&1/106.0/5.0&0/126.5/6.0&0/116.0/5.0&0/136.0/7.0&3/601.0/29.0\\
\rule{0pt}{2.5ex}
500k&0/163.0/8.0&0/135.5/7.0&1/151.5/7.0&2/135.5/7.0&1/121.0/5.0&4/706.5/34.0\\
\rule{0pt}{2.5ex}
1.5M&0/173.0/10.0&1/170.0/9.0&0/175.0/10.0&0/178.0/9.0&0/169.0/9.0&1/865.0/47.0\\
\rule{0pt}{2.5ex}
6.7M&2/171.0/9.0&0/191.0/10.0&2/159.0/8.0&0/203.0/10.0&0/190.0/10.0&4/914.0/47.0\\

\noalign{\smallskip}\hline
\end{tabular}
}
\end{table}
%%%%%%%%%%%%%%%%%%%%%%%%%%%%%%%%%%%%%%%%%%%%%%%%%%%%%%%%%%%

Regarding the results achieved for each dataset, Table~\ref{tab:overviewall-ft-embs} shows the best predictive performances in terms of accuracy and $F_1$-macro. We can see that BERTweet achieved the best results for most datasets when fine-tuned with fewer number of tweets. More specifically, BERTweet outperformed the other models when fine-tuned with samples varying from 1K to 25K tweets in 14 out of the 22 datasets for both accuracy and $F_1$-macro. 

\begin{table}%[!htbp]
\caption{Best results achieved for each dataset by fine-tuning the Transformer-based models with different samples of generic tweets}
\label{tab:overviewall-ft-embs}
\scalebox{0.83}{
\begin{tabular}{llll|lll}
\hline\noalign{\smallskip}
\textbf{Dataset}&\textbf{Accuracy}& \textbf{Classifier} &\textbf{Model} & \bm{$F_1$}\textbf{-macro}& \textbf{Classifier} &\textbf{Model} \\
\noalign{\smallskip}\hline\noalign{\smallskip}
\rule{0pt}{0ex}
iro&82.30&MLP&BERTweet-50K&75.87&LR&BERT-500K\\
\rule{0pt}{3.0ex}
sar&77.32&SVM&BERT-50K&75.85&SVM&BERT-50K\\
\rule{0pt}{3.0ex}
ntu&94.38&SVM&BERTweet-1.5M&94.26&SVM&BERTweet-1.5M\\
\rule{0pt}{3.0ex}
S15&94.18&MLP&BERTweet-1K&86.22&MLP&BERTweet-1K\\
\rule{0pt}{3.0ex}
stm&93.04&SVM&BERTweet-5K&93.02&SVM&BERTweet-5K\\
\rule{0pt}{3.0ex}
per&89.51&LR&BERTweet-10K&87.53&LR&BERTweet-10K\\
\rule{0pt}{3.0ex}
hob&89.83&MLP&BERTweet-50K&88.30&MLP&BERTweet-50K\\
\rule{0pt}{3.0ex}
iph&88.16&MLP&RoBERTa-25K&85.85&MLP&RoBERTa-25K\\
\rule{0pt}{3.0ex}
mov&93.29&MLP&BERTweet-50K&88.27&LR&BERTweet-50K\\
\rule{0pt}{3.0ex}
san&91.83&LR&BERTweet-10K&91.77&LR&BERTweet-10K\\
\rule{0pt}{3.0ex}
Nar&97.04&MLP&BERTweet-1K&96.91&MLP&BERTweet-1K\\
\rule{0pt}{3.0ex}
arc&92.08&LR&BERTweet-25K&91.92&LR&BERTweet-25K\\
\rule{0pt}{3.0ex}
S18&90.48&SVM&BERTweet-10K&90.40&SVM&BERTweet-10K\\
\rule{0pt}{3.0ex}
OMD&88.99&SVM&BERTweet-5K&88.17&SVM&BERTweet-5K\\
\rule{0pt}{3.0ex}
HCR&82.18&XGB&RoBERTa-1K&78.18&LR&BERTweet-250K\\
\rule{0pt}{3.0ex}
STS&95.38&MLP&BERTweet-50K&94.59&MLP&BERTweet-50K\\
\rule{0pt}{3.0ex}
SSt&89.60&SVM&BERTweet-10K&89.36&SVM&BERTweet-10K\\
\rule{0pt}{3.0ex}
Tar&87.83&SVM&BERTweet-1K&87.82&SVM&BERTweet-1K\\
\rule{0pt}{3.0ex}
Vad&92.80&LR&BERTweet-1K&91.64&LR&BERTweet-1K\\
\rule{0pt}{3.0ex}
S13&90.70&LR&BERTweet-5K&88.59&LR&BERTweet-5K\\
\rule{0pt}{3.0ex}
S17&93.49&SVM&BERTweet-10K&93.07&SVM&BERTweet-10K\\
\rule{0pt}{3.0ex}
S16&91.98&SVM&BERTweet-5K&90.30&SVM&BERTweet-5K\\
\noalign{\smallskip}
\hline
\end{tabular}
}
\end{table}

As in previous sections, we also present an overall evaluation of combining all fine-tuned models and classifiers across the 22 datasets, in terms of the average rank position. Table~\ref{tab:top10_ranksum_ft} shows the top ten results among all 150 possible combinations (3 models $\times$ 10 samples of tweets $\times$ 5 classification algorithms). As we can see in Table~\ref{tab:top10_ranksum_ft}, fine-tuned BERTweet embeddings achieved the best overall performances when used to train LR, MLP, and SVM, mastering the top ten results. Also, note that by using LR, MLP, and SVM, BERTweet outperformed all other models when fine-tuned with samples containing 50K tweets or less.

Tables~\ref{tab:emb_ranksum-ft} and~\ref{tab:class_ranksum-ft} show the top ten results among all fine-tuned models and a summary of the results for each classifier, from best to worst, respectively, in terms of the average rank position. From Table~\ref{tab:emb_ranksum-ft}, we can notice that all BERTweet fine-tuned models  (0.5K, 1K, 5K, 10K, 25K, 50K, 250K, 500K, 1.5M, and 6.7M) were ranked in the top ten results. Furthermore, neither BERT nor RoBERTa appear in the top results, even when they are fine-tuned with the entire corpus of 6.7M tweets. RoBERTa appears only in the top 24 accuracy score with an average rank of 37.02 tuned with 50K tweets and combined MLP classifier and in top 28 F$_1$-macro score with an average rank of 37.27 tuned with 50K tweets and combined LR classifier. BERT appears only in the top 56 accuracy score with an average rank of 66.05 tuned with 1.5M tweets and combined MLP classifier and in top 51 F$_1$-macro score with an average rank of 60.77 tuned with 6.7M tweets and combined LR classifier. Among the classifiers, as we can see in Table~\ref{tab:class_ranksum-ft}, MLP and LR achieved the best predictive performances and were ranked as the top two best classifiers. Conversely, RF was ranked as the worst classifier.

%%%%%%%%%%%%%%%%%%%%%%%%%%%%%%%%%%%%%%%%%%%%%%%%%%%%%%%%%%%%%%%%%%%%%%%%%
\begin{table}%[!htbp]
\caption{Top 10 average rank position results achieved for each combination Model-Classifier by evaluating Transformer-Autoencoder models}
\label{tab:top10_ranksum_ft}
\scalebox{0.85}{
\begin{tabular}{lll|lll}
\hline\noalign{\smallskip}
\multirow{2}{*}{\textbf{Model}} & \multirow{2}{*}{\textbf{Classifier}} & \textbf{Accuracy} & \multirow{2}{*}{\textbf{Model}} & \multirow{2}{*}{\textbf{Classifier}} & \textbf{$\bm{F_1}$-macro} \\
& & \textbf{avg. rank pos.} & & & \textbf{avg. rank pos.}\\
\noalign{\smallskip}\hline\noalign{\smallskip}
\rule{0pt}{3ex}
BERTweet-5K&LR&11.95&BERTweet-5K&LR&11.18\\
\rule{0pt}{3.0ex}
BERTweet-5K&MLP&14.05&BERTweet-25K&SVM&13.43\\
\rule{0pt}{3.0ex}
BERTweet-25K&MLP&14.64&BERTweet-10K&SVM&13.95\\
\rule{0pt}{3.0ex}
BERTweet-25K&LR&16.14&BERTweet-10K&LR&14.20\\
\rule{0pt}{3.0ex}
BERTweet-50K&MLP&16.43&BERTweet-25K&LR&14.32\\
\rule{0pt}{3.0ex}
BERTweet-1K&MLP&16.77&BERTweet-1K&LR&15.11\\
\rule{0pt}{3.0ex}
BERTweet-10K&LR&16.82&BERTweet-5K&MLP&15.95\\
\rule{0pt}{3.0ex}
BERTweet-25K&SVM&17.02&BERTweet-25K&MLP&16.11\\
\rule{0pt}{3.0ex}
BERTweet-1K&LR&17.68&BERTweet-50K&LR&16.80\\
\rule{0pt}{3.0ex}
BERTweet-10K&SVM&17.93&BERTweet-50K&SVM&17.43\\

\noalign{\smallskip}
\hline
\end{tabular}
}
\end{table}

%%%%%%%%%%%%%%%%%%%%%%%%%%%%%%%%%%%%%%%%%%%%%%%%%%%%%%%%%%%%%%%%%%%%%%%%%
\begin{table}%[!htbp]
\caption{Top 10 average rank position results achieved for each Embedding evaluating Transformer-Autoencoder model}
\label{tab:emb_ranksum-ft}
\begin{tabular}{ll|ll}
\hline\noalign{\smallskip}
\multirow{2}{*}{\textbf{Model}}& \textbf{Accuracy} & \multirow{2}{*}{\textbf{Model}}& \textbf{$\bm{F_1}$-macro} \\
& \textbf{avg. rank pos.} & & \textbf{avg. rank pos.}\\
\noalign{\smallskip}\hline\noalign{\smallskip}
\rule{0pt}{3ex}
BERTweet-5K&37.76&BERTweet-5K&40.13\\
\rule{0pt}{3.0ex}
BERTweet-10K&40.81&BERTweet-10K&43.27\\
\rule{0pt}{3.0ex}
BERTweet-25K&41.24&BERTweet-25K&43.80\\
\rule{0pt}{3.0ex}
BERTweet-1K&43.09&BERTweet-1K&44.78\\
\rule{0pt}{3.0ex}
BERTweet-50K&45.22&BERTweet-50K&47.42\\
\rule{0pt}{3.0ex}
BERTweet-250K&48.65&BERTweet-250K&50.19\\
\rule{0pt}{3.0ex}
BERTweet-500K&53.96&BERTweet-500K&55.59\\
\rule{0pt}{3.0ex}
BERTweet-0.5K&58.00&BERTweet-0.5K&59.59\\
\rule{0pt}{3.0ex}
BERTweet-1.5M&66.63&BERTweet-1.5M&66.48\\
\rule{0pt}{3.0ex}
BERTweet-6.7M&71.09&BERTweet-6.7M&70.46\\

\noalign{\smallskip}
\hline
\end{tabular}
%}
\end{table}

%%%%%%%%%%%%%%%%%%%%%%%%%%%%%%%%%%%%%%%%%%%%%%%%%%%%%%%%%%%%%%%%%%%%%%%%%
\begin{table}%[!htbp]
\caption{Average rank position results achieved for each Classifier evaluating Transformer-Autoencoder model}
\label{tab:class_ranksum-ft}
\begin{tabular}{ll|ll}
\hline\noalign{\smallskip}
\multirow{2}{*}{\textbf{Classifier}}& \textbf{Accuracy} & \multirow{2}{*}{\textbf{Classifier}}& \textbf{$\bm{F_1}$-macro} \\
& \textbf{avg. rank pos.} & & \textbf{avg. rank pos.}\\
\noalign{\smallskip}\hline\noalign{\smallskip}
\rule{0pt}{3ex}
MLP&48.52&LR&48.71\\
\rule{0pt}{3.0ex}
LR&53.17&MLP&49.92\\
\rule{0pt}{3.0ex}
SVM&70.74&SVM&60.83\\
\rule{0pt}{3.0ex}
XGB&84.90&XGB&93.16\\
\rule{0pt}{3.0ex}
RF&120.18&RF&124.87\\
\noalign{\smallskip}
\hline
\end{tabular}
%}
\end{table}

%%%%%%%%%%%%%%%%%%%%%%%%%%%%%%%%%%%%%%%%%%%%%%%%

From all previous evaluations, we can note that as the size of the samples increases, the fine-tuning procedure seems to be less effective. It may be due to the adjustment of the weights of the models' layers during the back-propagation process. Considering that the fine-tuning procedure consists in unfreezing the entire model obtained previously and adjusting their weights with the new data, the original model and the semantic and syntactic knowledge learned in its layers are changed. In that case, we believe that after some training iterations, the adjustment of the weights starts to damage the original knowledge embedded in the models' layers. The aforementioned conclusion may further explain why BERTweet achieved improved classification performance by using smaller samples of tweets as compared to BERT and RoBERTa. 
Our hypothesis is that, considering that the weights in BERTweet's layers are specifically adjusted to fit tweets' language style, using more data to fine-tune the model means only continue the initial training. It may be that lots of data may harm the learned weights of the model. Thus, we suggest that when fine-tuning Transformer-based models, such as BERT, RoBERTa, and BERTweet, samples of different sizes may be exploited instead of adopting a dataset with a massive number of instances.

Additionally, we present a comparison among all fine-tuned Transformer-based models against their original versions. Tables~\ref{tab:emb_ranksum-ft-context-bert},~\ref{tab:emb_ranksum-ft-context-rob}, and~\ref{tab:emb_ranksum-ft-context-bt} report this comparison in terms of the average rank position for BERT, RoBERTa, and BERTweet, respectively. We can see that the fine-tuned versions achieved meaningful predictive performances as compared to their  original models, which indicates that fine-tuning strategies can boost classification performance in Twitter sentiment analysis. Moreover, from Tables~\ref{tab:emb_ranksum-ft-context-bert} and~\ref{tab:emb_ranksum-ft-context-rob}, we note that the fine-tuned versions of BERT and RoBERTa benefited most from samples containing a large amount of tweets. Conversely, as pointed out before, BERTweet achieved better overall performances by using smaller samples, as shown in~Table~\ref{tab:emb_ranksum-ft-context-bt}.

\begin{table}%[!htbp]
\caption{Average rank position results achieved for BERT model and its tuned models}
\label{tab:emb_ranksum-ft-context-bert}
\begin{tabular}{ll|ll}
\hline\noalign{\smallskip}
\multirow{2}{*}{\textbf{Model}}& \textbf{Accuracy} & \multirow{2}{*}{\textbf{Model}}& \textbf{$\bm{F_1}$-macro} \\
& \textbf{avg. rank pos.} & & \textbf{avg. rank pos.}\\
\noalign{\smallskip}\hline\noalign{\smallskip}
\rule{0pt}{3ex}
BERT-250K&25.62&BERT-250K&25.62\\
\rule{0pt}{3.0ex}
BERT-5K&26.95&BERT-1.5M&26.45\\
\rule{0pt}{3.0ex}
BERT-1.5M&26.96&BERT-6.7M&26.69\\
\rule{0pt}{3.0ex}
BERT-500K&27.09&BERT-500K&26.70\\
\rule{0pt}{3.0ex}
BERT-6.7M&27.67&BERT-5K&27.69\\
\rule{0pt}{3.0ex}
BERT-0.5K&28.16&BERT-50K&28.36\\
\rule{0pt}{3.0ex}
BERT-50K&28.38&BERT-0.5K&28.40\\
\rule{0pt}{3.0ex}
BERT-1K&28.46&BERT-1K&28.95\\
\rule{0pt}{3.0ex}
BERT-10K&29.52&BERT (original)&29.50\\
\rule{0pt}{3.0ex}
BERT (original)&29.52&BERT-10K&29.68\\
\rule{0pt}{3.0ex}
BERT-25K&29.66&BERT-25K&29.95\\

\noalign{\smallskip}
\hline
\end{tabular}
%}
\end{table}

\begin{table}%[!htbp]
\caption{Average rank position results achieved for RoBERTa model and its tuned models}
\label{tab:emb_ranksum-ft-context-rob}
%\scalebox{0.9}{
\begin{tabular}{ll|ll}
\hline\noalign{\smallskip}
\multirow{2}{*}{\textbf{Model}}& \textbf{Accuracy} & \multirow{2}{*}{\textbf{Model}}& \textbf{$\bm{F_1}$-macro} \\
& \textbf{avg. rank pos.} & & \textbf{avg. rank pos.}\\
\noalign{\smallskip}\hline\noalign{\smallskip}
\rule{0pt}{3ex}
RoBERTa-50K&24.34&RoBERTa-50K&24.24\\
\rule{0pt}{3.0ex}
RoBERTa-500K&24.69&RoBERTa-1.5M&24.61\\
\rule{0pt}{3.0ex}
RoBERTa-1.5M&24.82&RoBERTa-500K&24.95\\
\rule{0pt}{3.0ex}
RoBERTa-5K&25.44&RoBERTa-5K&25.54\\
\rule{0pt}{3.0ex}
RoBERTa-250K&25.53&RoBERTa-250K&25.66\\
\rule{0pt}{3.0ex}
RoBERTa-25K&26.49&RoBERTa-25K&27.05\\
\rule{0pt}{3.0ex}
RoBERTa-10K&27.28&RoBERTa-10K&27.50\\
\rule{0pt}{3.0ex}
RoBERTa-1K&29.84&RoBERTa-1K&29.65\\
\rule{0pt}{3.0ex}
RoBERTa-0.5K&32.01&RoBERTa-0.5K&31.78\\
\rule{0pt}{3.0ex}
RoBERTa-6.7M&32.75&RoBERTa-6.7M&31.96\\
\rule{0pt}{3.0ex}
RoBERTa (original)&34.81&RoBERTa (original)&35.06\\

\noalign{\smallskip}
\hline
\end{tabular}
%}
\end{table}

\begin{table}%[!htbp]
\caption{Average rank position results achieved for BERTweet model and its tuned models}
\label{tab:emb_ranksum-ft-context-bt}
%\scalebox{0.9}{
\begin{tabular}{ll|ll}
\hline\noalign{\smallskip}
\multirow{2}{*}{\textbf{Model}}& \textbf{Accuracy} & \multirow{2}{*}{\textbf{Model}}& \textbf{$\bm{F_1}$-macro} \\
& \textbf{avg. rank pos.} & & \textbf{avg. rank pos.}\\
\noalign{\smallskip}\hline\noalign{\smallskip}
\rule{0pt}{3ex}
BERTweet-5K&20.74&BERTweet-5K&21.38\\
\rule{0pt}{3.0ex}
BERTweet-25K&22.62&BERTweet-25K&22.94\\
\rule{0pt}{3.0ex}
BERTweet-10K&22.85&BERTweet-10K&23.10\\
\rule{0pt}{3.0ex}
BERTweet-1K&23.73&BERTweet-1K&23.93\\
\rule{0pt}{3.0ex}
BERTweet-50K&25.25&BERTweet-50K&25.55\\
\rule{0pt}{3.0ex}
BERTweet-250K&26.67&BERTweet-250K&26.66\\
\rule{0pt}{3.0ex}
BERTweet-500K&30.31&BERTweet-500K&30.48\\
\rule{0pt}{3.0ex}
BERTweet-0.5K&31.70&BERTweet-0.5K&31.72\\
\rule{0pt}{3.0ex}
BERTweet (original)&33.80&BERTweet (original)&33.35\\
\rule{0pt}{3.0ex}
BERTweet-1.5M&34.80&BERTweet-1.5M&34.05\\
\rule{0pt}{3.0ex}
BERTweet-6.7M&35.53&BERTweet-6.7M&34.85\\

\noalign{\smallskip}
\hline
\end{tabular}
%}
\end{table}

Addressing research question RQ3, we could see that fine-tuning Transformer-based models improves the classification effectiveness in Twitter sentiment analysis. Nevertheless, using large sets of tweets does not guarantee better predictive performances, particularly for those models trained from scratch on tweets, such as BERTweet. We could observe that BERTweet benefited most from samples of tweets containing 50K tweets or less. Furthermore, regarding the classifiers, in general, MLP and LR seem to be good choices of classifiers to be employed after extracting the features from fine-tuned Transformer-based models.

%%%%%%%%%%%%%%%%%%%%%%%% Seção 7 %%%%%%%%%%%%%%%%%%%
\section{Fine-tuning Transformer-based models using sentiment datasets}
\label{sec:finetuning-benchmark}

The experiments conducted in this section aim at answering the research question RQ4, stated as follows:

\textit{RQ4. Can Transformer-based autoencoder models benefit from a fine-tuning procedure with tweets from sentiment analysis datasets?}

We address this research question by evaluating whether the sentiment classification of tweets benefits from fine-tuned language models using tweets from sentiment analysis datasets. For this purpose, we use the same collection of 22 benchmark datasets presented in Section~\ref{sec:mth-dataset} (Table~\ref{tab:datasets-info}). We perform this evaluation by assessing three distinct strategies to simulate three real-world scenarios. In addition, as done in Section~\ref{sec:finetuning-exp}, all experiments were performed three times using different seeds (12,34,56), with all the same hyperparameter and we report the average of the results.

The first fine-tuning strategy we investigate, referred to as \emph{InData}, simulates the usage of a specific sentiment dataset itself as the new domain dataset to fine-tune a pre-trained language model. Precisely, each one of the 22 datasets is used once as the target dataset. For each of the 22 datasets, we use a 10-fold cross-validation procedure. In each of the ten executions, we use the tweets from nine folds as the source data (i.e., the training data) used to adjust a language model, which is then validated on the one remaining part of the data, referred to as the target dataset (i.e., the test data).

The second strategy, referred to as \emph{LOO} (Leave One dataset Out), aims at simulating the situation where a collection of general sentiment datasets is available to fine-tune the language model. We use each dataset once as the target dataset while the tweets from the remaining 21 datasets are combined to tune the language model. Although the target dataset contains sentiment labels for each tweet, these labels are not used in the fine-tuning process as we leverage the intermediate self-supervised masked language model task to fine-tune the network parameters.

The third and last strategy, referred to as \emph{AllData}, is a combination of the two others. Specifically, as for strategy InData, for each assessed dataset (target dataset), and for each of the nine folds in the 10-fold cross-validation procedure, we combine the tweets from the nine folds (i.e., the training data of the target dataset) with the tweets from the remaining 21 datasets to fine-tune a language model. This last strategy evaluates the benefits of combining the tweets from a specific sentiment target dataset with a representative general sentiment dataset corpus in the fine-tuning process.

Table~\ref{tab:finetuning-svm-results-acc-bert-f1} presents the predictive performances achieved by fine-tuning each language model with strategies InData, LOO, and AllData, one at a time, by using the SVM classifier. As in previous sections, for space constraints, we only report the detailed evaluation using the SVM classifier (refer to Online Resource 1 for the detailed assessment of each classifier).

From Table~\ref{tab:finetuning-svm-results-acc-bert-f1}, we can observe that BERT benefited most from strategy InData, which uses only the target dataset itself to adjust the language models. Conversely, fine-tuning RoBERTa and BERTweet models using strategies that combine tweets from distinct sentiment analysis corpora achieved the best results for most datasets. More clearly, AllData, which combines the tweets from the target dataset and tweets from a collection of sentiment datasets, achieved the best overall results with both RoBERTa and BERTweet. Also, regarding BERTweet, note that strategy LOO achieved comparable performances to AllData. It is also noteworthy that smaller datasets seem to have benefited most from fine-tuning RoBERTa and BERTweet by using strategy LOO. On the other hand, larger datasets achieved higher predictive performances when using strategy AllData to fine-tune RoBERTa and BERTweet. Table~\ref{tab:overview-acc-emb-finetuning-acc} shows a summary of the complete evaluation regarding all classifiers.

%%%%%%%%%%%%%%%%%%%%%%%%%%%%%%%%%%%%%%%%%%%%%%%%%%%%%%%%%%%%%%%%%%%%%%%%
\begin{table}%[!htbp]
\caption{Accuracies and $F_1$-macro scores (\%) achieved by evaluating InData, LOO, and AllData fine-tuning strategies using the SVM classifier}
\label{tab:finetuning-svm-results-acc-bert-f1}
\scalebox{0.64}{
\begin{tabular}{llll|lll}
\hline\noalign{\smallskip}
& \multicolumn{3}{c}{\textbf{Accuracy}} & \multicolumn{3}{c}{\bm{$F_1$}\textbf{-macro}} \\
\noalign{\smallskip}\hline\noalign{\smallskip}
\multirow{2}{*}{\textbf{Dataset}}&\multicolumn{6}{c}{\textbf{BERT}}\\
\noalign{\smallskip}\cline{2-7}\noalign{\smallskip}
& \textbf{AllData} & \textbf{LOO} & \textbf{InData} & \textbf{AllData} & \textbf{LOO} & \textbf{InData}\\
\noalign{\smallskip}\hline\noalign{\smallskip}
iro&74.40&\textbf{78.81}&67.90&65.40&\textbf{70.55}&59.60\\
sar&\textbf{71.0}&70.18&64.10&68.50&\textbf{68.58}&60.20\\
ntu&85.00&82.74&\textbf{88.10}&84.70&82.39&\textbf{87.80}\\
S15&89.70&88.14&\textbf{89.80}&77.50&77.11&\textbf{77.80}\\
stm&88.80&\textbf{90.25}&89.90&88.70&\textbf{90.24}&89.80\\
per&84.40&\textbf{85.66}&82.00&82.20&\textbf{83.49}&80.00\\
hob&\textbf{84.60}&84.46&82.30&82.90&\textbf{83.08}&80.70\\
iph&82.70&\textbf{83.09}&83.00&80.80&81.07&\textbf{81.40}\\
mov&85.80&\textbf{86.46}&84.60&79.40&\textbf{80.14}&78.10\\
san&87.60&87.49&\textbf{87.80}&87.50&87.43&\textbf{87.70}\\
Nar&92.20&92.50&\textbf{94.90}&92.00&92.25&\textbf{94.70}\\
arc&89.10&88.42&\textbf{89.90}&88.90&88.20&\textbf{89.80}\\
S18&87.70&87.36&\textbf{89.70}&87.60&87.26&\textbf{89.60}\\
OMD&85.90&85.73&\textbf{87.30}&85.00&84.74&\textbf{86.40}\\
HCR&79.30&79.03&\textbf{79.60}&75.70&75.25&\textbf{75.90}\\
STS&91.70&90.71&\textbf{93.50}&90.50&89.35&\textbf{92.60}\\
SSt&84.70&84.71&\textbf{87.50}&84.30&84.39&\textbf{87.20}\\
Tar&85.70&86.24&\textbf{86.90}&85.70&86.23&\textbf{86.90}\\
Vad&89.90&90.16&\textbf{91.50}&88.50&88.84&\textbf{90.30}\\
S13&87.50&87.60&\textbf{88.70}&85.20&85.31&\textbf{86.60}\\
S17&91.80&91.56&\textbf{92.90}&91.30&91.04&\textbf{92.40}\\
S16&89.50&89.07&\textbf{90.70}&87.40&86.93&\textbf{88.80}\\
\noalign{\smallskip}\hline
\#wins &2& 5& \textbf{15}& 0& 6&\textbf{16}\\
rank sums &49.0&49.0&\textbf{34.0}&51.0&48.0&\textbf{33.0}\\
position &2.5&2.5&\textbf{1.0}&3.0&2.0&\textbf{1.0}\\
\noalign{\smallskip}
\end{tabular}
}

\scalebox{0.64}{
\begin{tabular}{llll|lll}
\hline\noalign{\smallskip}
\multirow{2}{*}{\textbf{Dataset}}& \multicolumn{6}{c}{\textbf{RoBERTa}}\\
\noalign{\smallskip}\cline{2-7}\noalign{\smallskip}
& \textbf{AllData} & \textbf{LOO} & \textbf{InData} & \textbf{AllData} & \textbf{LOO} & \textbf{InData}\\
\noalign{\smallskip}\hline\noalign{\smallskip}
iro&46.70&46.67&46.70&31.00&31.00&31.00\\
sar&64.00&\textbf{65.00}&64.00&52.90&53.49&\textbf{54.20}\\
ntu&\textbf{84.30}&83.48&81.20&\textbf{83.90}&83.07&80.80\\
S15&\textbf{87.20}&86.31&86.20&\textbf{74.90}&74.28&72.40\\
stm&90.00&\textbf{90.25}&87.60&89.90&\textbf{90.21}&87.60\\
per&\textbf{71.10}&70.38&65.50&\textbf{70.20}&69.17&64.90\\
hob&72.80&\textbf{73.94}&71.30&71.90&\textbf{73.10}&70.50\\
iph&\textbf{79.90}&78.96&78.60&\textbf{78.60}&77.72&77.20\\
mov&\textbf{81.50}&79.87&72.10&\textbf{75.30}&74.01&66.50\\
san&\textbf{87.50}&87.17&85.50&\textbf{87.40}&87.04&85.30\\
Nar&\textbf{93.10}&92.50&92.10&\textbf{92.90}&92.35&91.90\\
arc&89.40&\textbf{89.47}&89.00&\textbf{89.30}&89.27&88.70\\
S18&88.40&\textbf{88.54}&88.00&88.30&\textbf{88.44}&87.80\\
OMD&85.60&84.58&\textbf{85.70}&84.50&83.60&\textbf{84.70}\\
HCR&76.90&76.04&\textbf{78.10}&73.20&72.59&\textbf{74.20}\\
STS&\textbf{92.60}&92.13&91.50&\textbf{91.60}&90.97&90.30\\
SSt&\textbf{86.40}&85.63&85.90&\textbf{86.10}&85.41&85.70\\
Tar&85.90&85.87&\textbf{86.30}&85.90&85.85&\textbf{86.30}\\
Vad&89.80&89.37&\textbf{89.90}&88.60&88.10&88.60\\
S13&\textbf{86.60}&86.41&86.10&\textbf{84.60}&84.41&83.90\\
S17&\textbf{92.00}&91.71&91.60&\textbf{91.50}&91.21&91.10\\
S16&89.50&\textbf{89.71}&89.30&87.60&\textbf{87.75}&87.40\\
\noalign{\smallskip}\hline
\#wins &\textbf{12}& 6& 5& \textbf{13}& 4&5\\
rank sums &\textbf{33.0}&44.0&55.0&\textbf{32.5}&45.0&54.5\\
position &\textbf{1.0}&2.0&3.0&\textbf{1.0}&2.0&3.0\\
\noalign{\smallskip}
\end{tabular}
}

\scalebox{0.64}{
\begin{tabular}{llll|lll}
\hline\noalign{\smallskip}
\multirow{2}{*}{\textbf{Dataset}}&\multicolumn{6}{c}{\textbf{BERTweet}}\\
\noalign{\smallskip}\cline{2-7}\noalign{\smallskip}
& \textbf{AllData} & \textbf{LOO} & \textbf{InData} & \textbf{AllData} & \textbf{LOO} & \textbf{InData}\\
\noalign{\smallskip}\hline\noalign{\smallskip}
iro&74.60&\textbf{83.10}&66.70&66.50&\textbf{77.24}&60.10\\
sar&\textbf{68.60}&67.50&61.80&\textbf{65.50}&64.32&56.40\\
ntu&92.10&\textbf{93.54}&90.10&91.80&\textbf{93.33}&89.80\\
S15&90.70&\textbf{92.84}&90.20&80.00&\textbf{84.76}&78.60\\
stm&92.70&\textbf{92.75}&90.50&92.60&\textbf{92.73}&90.50\\
per&86.10&\textbf{86.55}&82.50&84.10&\textbf{84.48}&80.70\\
hob&87.10&\textbf{87.17}&82.50&85.60&\textbf{85.62}&80.80\\
iph&\textbf{85.10}&83.48&83.30&\textbf{83.50}&81.79&81.80\\
mov&\textbf{89.90}&88.42&87.00&\textbf{84.50}&81.99&80.90\\
san&\textbf{91.40}&91.34&89.00&\textbf{91.40}&91.27&88.90\\
Nar&\textbf{97.00}&96.66&96.20&\textbf{96.80}&96.54&96.00\\
arc&\textbf{91.40}&90.57&90.70&\textbf{91.30}&90.40&90.50\\
S18&\textbf{90.90}&90.26&90.60&\textbf{90.80}&90.19&90.50\\
OMD&89.20&\textbf{89.77}&88.40&88.40&\textbf{88.99}&87.50\\
HCR&\textbf{81.50}&81.27&80.40&\textbf{77.90}&77.79&76.70\\
STS&\textbf{95.20}&94.99&94.70&\textbf{94.50}&94.21&93.90\\
SSt&\textbf{89.10}&88.51&88.80&\textbf{88.90}&88.25&88.50\\
Tar&\textbf{87.70}&87.63&87.30&\textbf{87.70}&87.62&87.30\\
Vad&92.50&\textbf{92.73}&92.30&91.40&\textbf{91.70}&91.20\\
S13&\textbf{90.00}&89.52&89.40&\textbf{88.00}&87.49&87.40\\
S17&\textbf{93.60}&93.59&93.20&93.10&\textbf{93.17}&92.80\\
S16&\textbf{91.80}&91.62&91.50&\textbf{90.10}&89.90&89.80\\
\noalign{\smallskip}\hline
\#wins &\textbf{14}& 8& 0& \textbf{13}& 9&0\\
rank sums &\textbf{30.0}&39.0&63.0&\textbf{31.0}&39.0&62.0\\
position &\textbf{1.0}&2.0&3.0&\textbf{1.0}&2.0&3.0\\
\noalign{\smallskip}\hline
\end{tabular}
}
\end{table}

\begin{table}%[!htbp]
\caption{Overview of the results (number of wins, rank sum, and rank position, respectively) achieved by each classifier when fine-tuning the Transformer-Autoencoder models using strategies InData, LOO, and AllData in terms of accuracy and F$_1$-macro}
\label{tab:overview-acc-emb-finetuning-acc}
\scalebox{0.81}{
\begin{tabular}{lllllll}
\hline\noalign{\smallskip}
\multicolumn{7}{c}{\textbf{ACCURACY}}\\
\hline\noalign{\smallskip}
\textbf{Strategy}& \textbf{LR} & \textbf{SVM} & \textbf{MLP} & \textbf{RF} & \textbf{XGB} &\textbf{Total} \\

\hline\noalign{\smallskip}
\multicolumn{7}{c}{\textbf{BERT}}\\
\noalign{\smallskip}\hline\noalign{\smallskip}
\rule{0pt}{0ex}
AllData&1/51.0/2.0&2/49.0/2.5&3/43.5/2.0&5/42.0/2.0&2/47.5/2.0&13/233.0/10.5\\
\rule{0pt}{3.0ex}
LOO&1/56.0/3.0&5/49.0/2.5&3/56.5/3.0&6/49.0/3.0&3/53.0/3.0&18/263.5/14.5\\
\rule{0pt}{3.0ex}
InData&\textbf{20}/\textbf{25.0}/\textbf{1.0}&\textbf{15}/\textbf{34.0}/\textbf{1.0}&\textbf{14}/\textbf{32.0}/\textbf{1.0}&\textbf{9}/\textbf{41.0}/\textbf{1.0}&\textbf{14}/\textbf{31.5}/\textbf{1.0}&\textbf{72}/\textbf{163.5}/\textbf{5.0}\\

\hline\noalign{\smallskip}
\multicolumn{7}{c}{\textbf{RoBERTa}}\\
\noalign{\smallskip}\hline\noalign{\smallskip}
\rule{0pt}{0ex}

AllData&\textbf{11}/\textbf{32.0}/\textbf{1.0}&\textbf{11}/\textbf{33.0}/\textbf{1.0}&\textbf{10}/\textbf{35.0}/\textbf{1.0}&\textbf{11}/\textbf{34.0}/\textbf{1.0}&\textbf{12}/\textbf{35.0}/\textbf{1.0}&\textbf{55}/\textbf{169.0}/\textbf{5.0}\\
\rule{0pt}{3.0ex}
LOO&6/48.0/2.0&6/44.0/2.0&9/42.0/2.0&10/37.0/2.0&8/41.0/2.0&39/212.0/10.0\\
\rule{0pt}{3.0ex}
InData&3/52.0/3.0&4/55.0/3.0&2/55.0/3.0&1/61.0/3.0&2/56.0/3.0&12/279.0/15.0\\

\hline\noalign{\smallskip}
\multicolumn{7}{c}{\textbf{BERTweet}}\\
\noalign{\smallskip}\hline\noalign{\smallskip}
\rule{0pt}{0ex}

AllData&\textbf{10}/\textbf{36.5}/\textbf{1.0}&\textbf{14}/\textbf{30.0}/\textbf{1.0}&9/\textbf{34.5}/\textbf{1.0}&\textbf{13}/\textbf{31.0}/\textbf{1.0}&\textbf{12}/\textbf{33.5}/\textbf{1.0}&\textbf{58}/\textbf{165.5}/\textbf{5.0}\\
\rule{0pt}{3.0ex}
LOO&9/39.5/2.0&8/39.0/2.0&\textbf{11}/37.0/2.0&8/39.5/2.0&8/39.0/2.0&44/194.0/10.0\\
\rule{0pt}{3.0ex}
InData&2/56.0/3.0&0/63.0/3.0&1/60.5/3.0&1/61.5/3.0&1/59.5/3.0&5/300.5/15.0\\

\noalign{\smallskip}\hline\noalign{\smallskip}
\multicolumn{7}{c}{\bm{$F_1$}\textbf{-MACRO}}\\
\hline\noalign{\smallskip}
\textbf{Strategy} & \textbf{LR} & \textbf{SVM} & \textbf{MLP} & \textbf{RF} & \textbf{XGB} &\textbf{Total} \\
\hline\noalign{\smallskip}
\multicolumn{7}{c}{\textbf{BERT}}\\
\noalign{\smallskip}\hline\noalign{\smallskip}
\rule{0pt}{0ex}
AllData&1/51.0/2.0&0/51.0/3.0&3/45.0/2.0&6/41.0/2.0&3/48.5/2.0&13/236.5/11.0\\
\rule{0pt}{3.0ex}
LOO&2/56.0/3.0&6/48.0/2.0&4/55.0/3.0&5/51.0/3.0&3/52.0/3.0&20/262.0/14.0\\
\rule{0pt}{3.0ex}
InData&\textbf{19}/\textbf{25.0}/\textbf{1.0}&\textbf{16}/\textbf{33.0}/\textbf{1.0}&\textbf{15}/\textbf{32.0}/\textbf{1.0}&\textbf{11}/\textbf{40.0}/\textbf{1.0}&\textbf{15}/\textbf{31.5}/\textbf{1.0}&\textbf{76}/\textbf{161.5}/\textbf{5.0}\\

\hline\noalign{\smallskip}
\multicolumn{7}{c}{\textbf{RoBERTa}}\\
\noalign{\smallskip}\hline\noalign{\smallskip}
\rule{0pt}{0ex}

AllData&\textbf{13}/\textbf{31.0}/\textbf{1.0}&\textbf{12}/\textbf{32.5}/\textbf{1.0}&9/\textbf{35.5}/\textbf{1.0}&\textbf{11}/\textbf{34.0}/\textbf{1.0}&\textbf{12}/\textbf{35.0}/\textbf{1.0}&\textbf{57}/\textbf{168.0}/\textbf{5.0}\\
\rule{0pt}{3.0ex}
LOO&5/49.0/2.0&4/45.0/2.0&\textbf{10}/42.0/2.0&10/36.0/2.0&8/40.0/2.0&37/212.0/10.0\\
\rule{0pt}{3.0ex}
InData&4/52.0/3.0&4/54.5/3.0&2/54.5/3.0&1/62.0/3.0&2/57.0/3.0&13/280.0/15.0\\

\hline\noalign{\smallskip}
\multicolumn{7}{c}{\textbf{BERTweet}}\\
\noalign{\smallskip}\hline\noalign{\smallskip}
\rule{0pt}{0ex}

AllData&\textbf{10}/\textbf{35.5}/\textbf{1.0}&\textbf{13}/\textbf{31.0}/\textbf{1.0}&\textbf{13}/\textbf{31.0}/\textbf{1.0}&\textbf{13}/\textbf{31.0}/\textbf{1.0}&\textbf{10}/\textbf{35.0}/\textbf{1.0}&\textbf{59}/\textbf{163.5}/\textbf{5.0}\\
\rule{0pt}{3.0ex}
LOO&\textbf{10}/\textbf{37.5}/\textbf{2.0}&9/39.0/2.0&9/39.0/2.0&8/39.5/2.0&\textbf{10}/37.5/2.0&46/192.5/10.0\\
\rule{0pt}{3.0ex}
InData&1/59.0/3.0&0/62.0/3.0&0/62.0/3.0&1/61.5/3.0&0/59.5/3.0&2/304.0/15.0\\
\noalign{\smallskip}\hline
\end{tabular}
}
\end{table}

Regarding the overall results achieved for each dataset, Table~\ref{tab:overviewall-ft-strategy} presents the best results. We can note that when fine-tuning the Transformer-based models with tweets from sentiment datasets, BERTweet outperformed BERT and RoBERTa for all datasets, except for datasets sarcasm (sar) and hobbit (hob). Interestingly, as mentioned before, while strategy LOO achieved the best results for smaller datasets, larger datasets seem to benefit from strategy AllData. Precisely, strategy AllData achieved the best overall performances in ten out of the 22 datasets in terms of accuracy and in 11 out of the 22 datasets in terms of $F_1$-macro. Strategy LOO achieved the best results in nine out of the 22 datasets for both accuracy and $F_1$-macro. 
The better performance of the AllData strategy for larger target datasets indicates that the significant amount of information present in the target dataset is indispensable for the fine-tuning process, while the information present in smaller datasets seems not to contribute to the fine-tuning process, making the LOO strategy adequate for datasets with a limited amount of tweets.

Conversely, strategy InData did not achieve meaningful results. The inferior performance of the InData strategy in almost all datasets shows that, regardless of the size of the dataset, the use of external and more extensive data brings more information to the fine-tuning process, improving the final performance.

%\ \textcolor{red}{With the noticed size sensibility, it is possible to conclude that choosing one of the fine-tuning strategies with sentiment analysis datasets is essential to consider the size of the target domain dataset. 
%\ So, the immediate decision is to use or not a combination of own dataset target with a set of benchmark sentiment analysis datasets.}

%%%%%%%%%%%%%%%%%%%%%%%%%%%%%%%%%%%%%%%%%%%%%%%%%
\begin{table}%[!htbp]
%\caption{Best results achieved, in terms of accuracy and F1-macro, for each dataset considering all Transformer-Autoencoder fine tuned models}
\caption{Best results achieved for each dataset by fine-tuning the Transformer-Autoencoder models using strategies InData, LOO, and AllData}
\label{tab:overviewall-ft-strategy}
\scalebox{0.8}{
\begin{tabular}{llll|lll}
\hline\noalign{\smallskip}
\textbf{Dataset}&\textbf{Accuracy}& \textbf{Classifier} &\textbf{Embedding} & \bm{$F_1$}\textbf{-macro}& \textbf{Classifier} &\textbf{Embedding} \\
\noalign{\smallskip}\hline\noalign{\smallskip}
\rule{0pt}{0ex}
iro&83.33&LR&BERTweet-LOO&77.24&SVM&BERTweet-LOO\\
\rule{0pt}{3.0ex}
sar&80.66&MLP&RoBERTa-LOO&79.35&MLP&RoBERTa-LOO\\
\rule{0pt}{3.0ex}
ntu&93.89&LR&BERTweet-LOO&93.66&LR&BERTweet-LOO\\
\rule{0pt}{3.0ex}
S15&94.39&MLP&BERTweet-LOO&87.31&MLP&BERTweet-LOO\\
\rule{0pt}{3.0ex}
stm&92.75&SVM&BERTweet-LOO&92.73&SVM&BERTweet-LOO\\
\rule{0pt}{3.0ex}
per&88.52&MLP&BERTweet-LOO&86.14&LR&BERTweet-LOO\\
\rule{0pt}{3.0ex}
hob&89.50&MLP&RoBERTa-InData&88.00&MLP&RoBERTa-InData\\
\rule{0pt}{3.0ex}
iph&87.80&LR&BERTweet-InData&85.80&LR&BERTweet-AllData\\
\rule{0pt}{3.0ex}
mov&91.10&LR&BERTweet-AllData&85.40&LR&BERTweet-AllData\\
\rule{0pt}{3.0ex}
san&91.60&LR&BERTweet-AllData&91.50&LR&BERTweet-AllData\\
\rule{0pt}{3.0ex}
Nar&97.00&SVM&BERTweet-AllData&96.80&MLP&BERTweet-AllData\\
\rule{0pt}{3.0ex}
arc&92.10&MLP&BERTweet-AllData&91.90&MLP&BERTweet-AllData\\
\rule{0pt}{3.0ex}
S18&90.90&SVM&BERTweet-AllData&90.80&SVM&BERTweet-AllData\\
\rule{0pt}{3.0ex}
OMD&89.77&SVM&BERTweet-LOO&88.99&SVM&BERTweet-LOO\\
\rule{0pt}{3.0ex}
HCR&82.21&XGB&BERTweet-LOO&77.90&SVM&BERTweet-AllData\\
\rule{0pt}{3.0ex}
STS&95.20&SVM&BERTweet-AllData&94.50&SVM&BERTweet-AllData\\
\rule{0pt}{3.0ex}
SSt&89.10&SVM&BERTweet-AllData&88.90&SVM&BERTweet-AllData\\
\rule{0pt}{3.0ex}
Tar&87.70&SVM&BERTweet-AllData&87.70&SVM&BERTweet-AllData\\
\rule{0pt}{3.0ex}
Vad&93.14&LR&BERTweet-LOO&92.05&LR&BERTweet-LOO\\
\rule{0pt}{3.0ex}
S13&90.40&LR&BERTweet-InData&88.30&LR&BERTweet-InData\\
\rule{0pt}{3.0ex}
S17&93.60&SVM&BERTweet-AllData&93.17&SVM&BERTweet-LOO\\
\rule{0pt}{3.0ex}
S16&91.80&SVM&BERTweet-AllData&90.10&SVM&BERTweet-AllData\\
\noalign{\smallskip}
\hline
\end{tabular}
}
\end{table}

Next, we present an overall evaluation of combining all fine-tuned models and classifiers across the 22 datasets, in terms of the average rank position. Table~\ref{tab:top10_ranksum_ft-sent} reports the top ten results among all 45 possible combinations (3 language models $\times$ 3 fine-tuning strategies $\times$ 5 classification algorithms). We can observe that the LR classifier trained with BERTweet embeddings fine-tuned via strategy AllData achieved the best overall predictive performances. Also, note that the fine-tuned BERTweet embeddings with strategies AllData and LOO, combined with LR, MLP, and SVM, appear at the top of the ranking (top six results). Another point worth highlighting is that BERTweet masters the top ten results, appearing in eight out of the ten positions in terms of accuracy and in nine out of the ten positions in terms of $F_1$-macro.

%%%%%%%%%%%%%%%%%%%%%%%%%%%%%%%%%%
Tables~\ref{tab:emb_ranksum-ft-sent} and~\ref{tab:class_ranksum-ft-sent} show the results among all fine-tuned models and a summary of the results for each classifier, from best to worst, respectively, in terms of the average rank position. Once again, from Table~\ref{tab:emb_ranksum-ft-sent}, we can notice that all BERTweet fine-tuned models  (InData, LOO, and AllData) were ranked in the top three results. Among the classifiers, as we can see in Table~\ref{tab:class_ranksum-ft-sent}, MLP and LR achieved the best predictive performances and were ranked as the top two best classifiers. Conversely, RF was ranked as the worst classifier.

%As in the preview sections, we also conducted an assessment to realize the more appropriated model for sentiment classification in English tweets using sentiment-dataset to fine-tuning language models. We conducted an overall analysis of the Average rank position, comparing all Tuned model-Classifier combinations. Table~\ref{tab:top10_ranksum_ft-sent} presents the TOP10 average rank position of the pair Embedding-Classifier of the total 45 combinations. As it is possible to realize, BERTweet dominates the top10 rank combined, mainly with LR and MLP.  All three strategies with BERTweet have better performance than any of BERT or RoBERTa. Among the three methods, AllData is the one that, combined with BERTweet, brings better performance.
%%%%%%%%%%%%%%%%%%%%%%%%%%%%%%%%%%%%%%%%%%%%%%%%%%%%%%%%%%%%%%%%%%%%%%%%%
\begin{table}%[!htbp]
\caption{Top 10 average rank position results achieved for each combination Model-Classifier by evaluating Transformer-Autoencoder model}
\label{tab:top10_ranksum_ft-sent}
\scalebox{0.8}{
\begin{tabular}{lll|lll}
\hline\noalign{\smallskip}
\multirow{2}{*}{\textbf{Model}} & \multirow{2}{*}{\textbf{Classifier}} & \textbf{Accuracy} & \multirow{2}{*}{\textbf{Model}} & \multirow{2}{*}{\textbf{Classifier}} & \textbf{$\bm{F_1}$-macro} \\
& & \textbf{avg. rank pos.} & & & \textbf{avg. rank pos.}\\
\noalign{\smallskip}\hline\noalign{\smallskip}
\rule{0pt}{3ex}
BERTweet-AllData&LR&4.86&BERTweet-AllData&LR&4.18\\
\rule{0pt}{3.0ex}
BERTweet-AllData&MLP&5.57&BERTweet-LOO&LR&5.34\\
\rule{0pt}{3.0ex}
BERTweet-LOO&MLP&6.07&BERTweet-AllData&MLP&5.45\\
\rule{0pt}{3.0ex}
BERTweet-LOO&LR&6.16&BERTweet-AllData&SVM&6.11\\
\rule{0pt}{3.0ex}
BERTweet-AllData&SVM&7.36&BERTweet-LOO&MLP&6.75\\
\rule{0pt}{3.0ex}
BERTweet-LOO&SVM&8.64&BERTweet-LOO&SVM&6.86\\
\rule{0pt}{3.0ex}
BERTweet-InData&LR&9.11&BERTweet-InData&LR&8.36\\
\rule{0pt}{3.0ex}
BERTweet-InData&MLP&11.27&RoBERTa-AllData&LR&11.89\\
\rule{0pt}{3.0ex}
RoBERTa-AllData&MLP&13.09&BERTweet-InData&MLP&11.95\\
\rule{0pt}{3.0ex}
RoBERTa-AllData&LR&13.48&BERTweet-InData&SVM&12.93\\

\noalign{\smallskip}
\hline
\end{tabular}
}
\end{table}

%%%%%%%%%%%%%%%%%%%%%%%%%%%%%%%%%%%%%%%%%%%%%%%%%%%%%%%%%%%%%%%%%%%%%%%%%

\begin{table}%[!htbp]
\caption{Average position results achieved for each Embedding}
\label{tab:emb_ranksum-ft-sent}
%\scalebox{0.8}{
\begin{tabular}{ll|ll}
\hline\noalign{\smallskip}
\multirow{2}{*}{\textbf{Model}}& \textbf{Accuracy} & \multirow{2}{*}{\textbf{Model}}& \textbf{$\bm{F_1}$-macro} \\
& \textbf{avg. rank pos.} & & \textbf{avg. rank pos.}\\
\noalign{\smallskip}\hline\noalign{\smallskip}
\rule{0pt}{3ex}
BERTweet-AllData&12.99&BERTweet-AllData&13.51\\
\rule{0pt}{3.0ex}
BERTweet-LOO&14.04&BERTweet-LOO&14.46\\
\rule{0pt}{3.0ex}
BERTweet-InData&19.59&BERTweet-InData&19.95\\
\rule{0pt}{3.0ex}
RoBERTa-AllData&22.76&RoBERTa-AllData&22.70\\
\rule{0pt}{3.0ex}
RoBERTa-LOO&24.46&RoBERTa-LOO&24.00\\
\rule{0pt}{3.0ex}
BERT-InData&24.53&BERT-InData&24.32\\
\rule{0pt}{3.0ex}
RoBERTa-InData&27.90&RoBERTa-InData&27.70\\
\rule{0pt}{3.0ex}
BERT-AllData&30.12&BERT-AllData&29.85\\
\rule{0pt}{3.0ex}
BERT-LOO&30.61&BERT-LOO&30.51\\
\noalign{\smallskip}
\hline
\end{tabular}
%}
\end{table}

%%%%%%%%%%%%%%%%%%%%%%%%%%%%%%%%%%%%%%%%%%%%%%%%%%%%%%%%%%%%%%%%%%%%%%%%%
\begin{table}%[!htbp]
\caption{Average rank position results achieved for each Classifier}
\label{tab:class_ranksum-ft-sent}
%\scalebox{0.8}{
\begin{tabular}{ll|ll}
\hline\noalign{\smallskip}
\multirow{2}{*}{\textbf{Classifier}}& \textbf{Accuracy} & \multirow{2}{*}{\textbf{Classifier}}& \textbf{$\bm{F_1}$-macro} \\
& \textbf{avg. rank pos.} & & \textbf{avg. rank pos.}\\
\noalign{\smallskip}\hline\noalign{\smallskip}
\rule{0pt}{3ex}
MLP&14.83&LR&14.41\\
\rule{0pt}{3.0ex}
LR&15.73&MLP&15.28\\
\rule{0pt}{3.0ex}
SVM&22.42&SVM&19.77\\
\rule{0pt}{3.0ex}
XGB&25.70&XGB&28.13\\
\rule{0pt}{3.0ex}
RF&36.32&RF&37.41\\
\noalign{\smallskip}
\hline
\end{tabular}
%}
\end{table}

To evaluate the effectiveness of fine-tuning the Transformer-based models using tweets from sentiment datasets, we present a comparison among all fine-tuning strategies assessed in this study for each language model. Specifically, we compare the fine-tuned models presented in this section, by using strategies InData, LOO, and AllData, against the best fine-tuned models identified in Section~\ref{sec:finetuning-exp}, i.e., BERT-250K, RoBERTa-50K, and BERTweet-5K. Tables~\ref{tab:emb_ranksum-ft-context-bert-sent},~\ref{tab:emb_ranksum-ft-context-rob-sent}, and~\ref{tab:emb_ranksum-ft-context-bt-sent} report these results in terms of the average rank position for BERT, RoBERTa, and BERTweet, respectively.

Regarding BERT, as shown in Table~\ref{tab:emb_ranksum-ft-context-bert-sent}, note that all fine-tuning strategies using tweets from sentiment datasets achieved better overall results than using the sample of 250K generic tweets. Moreover, strategy InData appears at the top of the ranking as the best fine-tuning strategy. It is worth mentioning that strategy InData uses only the tweets from the target dataset itself to adjust the language model. This means that the strategy InData used a number of tweets much smaller than the 250K tweets contained in the sample.

On the other hand, as we can see in Tables~\ref{tab:emb_ranksum-ft-context-rob-sent} and~\ref{tab:emb_ranksum-ft-context-bt-sent}, strategy InData did not achieve meaningful results for RoBERTa and BERTweet models. Nevertheless, for these models, strategies AllData and LOO, which also use tweets from sentiment datasets, achieved rather comparable performances and were ranked as the top two best fine-tuning strategies.

%As we intent with this Section to conclude if a fine-tuning process with sentiment-dataset brings benefits to sentiment classification task, we compare separated each sentiment tuned model with the best preview fine-tuned models (BERT-250K, RoBERTa-50K and BERTweet-5K). By Tables~\ref{tab:emb_ranksum-ft-context-bert-sent},~\ref{tab:emb_ranksum-ft-context-rob-sent}, and~\ref{tab:emb_ranksum-ft-context-bt-sent}, analyzing the average Average rank position, we can conclude the fine-tuning process with dataset fill with sentiment really brings benefits for all language models performance. In almost all three models, the generic fine-tuned model is the second-worst or the worst performance.

\begin{table}%[!htbp]
\caption{Average rank position results achieved for BERT models}
\label{tab:emb_ranksum-ft-context-bert-sent}
%\scalebox{0.8}{
\begin{tabular}{ll|ll}
\hline\noalign{\smallskip}
\multirow{2}{*}{\textbf{Model}}& \textbf{Accuracy} & \multirow{2}{*}{\textbf{Model}}& \textbf{$\bm{F_1}$-macro} \\
& \textbf{avg. rank pos.} & & \textbf{avg. rank pos.}\\
\noalign{\smallskip}\hline\noalign{\smallskip}
\rule{0pt}{3ex}
BERT-InData&8.04&BERT-InData&8.10\\
\rule{0pt}{3.0ex}
BERT-AllData&10.61&BERT-AllData&10.66\\
\rule{0pt}{3.0ex}
BERT-LOO&11.24&BERT-LOO&11.29\\
\rule{0pt}{3.0ex}
BERT-250K&12.12&BERT-250K&11.95\\
\noalign{\smallskip}
\hline
\end{tabular}
%}
\end{table}

\begin{table}%[!htbp]
\caption{Average rank position results achieved for RoBERTa models}
\label{tab:emb_ranksum-ft-context-rob-sent}
%\scalebox{0.8}{
\begin{tabular}{ll|ll}
\hline\noalign{\smallskip}
\multirow{2}{*}{\textbf{Model}}& \textbf{Accuracy} & \multirow{2}{*}{\textbf{Model}}& \textbf{$\bm{F_1}$-macro} \\
& \textbf{avg. rank pos.} & & \textbf{avg. rank pos.}\\
\noalign{\smallskip}\hline\noalign{\smallskip}
\rule{0pt}{3ex}
RoBERTa-AllData&8.81&RoBERTa-AllData&8.85\\
\rule{0pt}{3.0ex}
RoBERTa-LOO&9.89&RoBERTa-LOO&9.84\\
\rule{0pt}{3.0ex}
RoBERTa-50K&11.61&RoBERTa-50K&11.52\\
\rule{0pt}{3.0ex}
RoBERTa-InData&11.68&RoBERTa-InData&11.80\\
\noalign{\smallskip}
\hline
\end{tabular}
%}
\end{table}

\begin{table}%[!htbp]
\caption{Average rank position results achieved for BERTweet models}
\label{tab:emb_ranksum-ft-context-bt-sent}
%\scalebox{0.8}{
\begin{tabular}{ll|ll}
\hline\noalign{\smallskip}
\multirow{2}{*}{\textbf{Model}}& \textbf{Accuracy} & \multirow{2}{*}{\textbf{Model}}& \textbf{$\bm{F_1}$-macro} \\
& \textbf{avg. rank pos.} & & \textbf{avg. rank pos.}\\
\noalign{\smallskip}\hline\noalign{\smallskip}
\rule{0pt}{3ex}
BERTweet-AllData&9.00&BERTweet-AllData&9.11\\
\rule{0pt}{3.0ex}
BERTweet-LOO&9.76&BERTweet-LOO&9.76\\
\rule{0pt}{3.0ex}
BERTweet-5K&10.19&BERTweet-5K&10.20\\
\rule{0pt}{3.0ex}
BERTweet-InData&13.05&BERTweet-InData&12.93\\
\noalign{\smallskip}
\hline
\end{tabular}
%}
\end{table}

To acknowledge the effectiveness of fine-tuning the Transformer-based models using tweets from sentiment datasets, we present an overall comparison among all fine-tuning strategies and all 47 models previously assessed in this study. Tables~\ref{tab:top10_ranksum_all-models-class} and~\ref{tab:tail10_ranksum_all-models-class} present, respectively, the 10 best and the 10 worst overall combination of model and classifier, assessing the average rank position of all 280 (56 models and five classifier) model and classifier combinations. We note that BERTweet tuned with tweets from sentiment datasets and combined with LR and MLP had the four best results, in terms of accuracy, and the two best results, in terms of $F_1$-macro. These combinations were followed by BERTweet tuned with generic tweets. More specifically, combinations with the strategy AllData and LOO achieved better overall results. Independently of the language model, LR and MLP were the most frequent classifier in the top 10 results. Conversely, all the ten worst combinations are static representations combined with RF, which was unanimous in the worst model and classifiers combinations.

Assessing only the different kinds of embeddings, Tables~\ref{tab:top-emb_ranksum-all-models} and~\ref{tab:tail-emb_ranksum-all-models} present, respectively, the best and the worst average rank position comparing all 56 representations (the nine models tuned with sentiment datasets and the 47 previous representations). This analysis confirms the good performance of fine-tuning the Transformer-based models using tweets from sentiment datasets. More specifically, the strategies AllData and LOO obtained the two best results. It is possible to notice that tuning BERTweet with generic tweets also brings performance improvement to BERTweet. Regarding the worst behaviors, presented in Table~\ref{tab:tail-emb_ranksum-all-models}, it is possible to note that all the ten strategies are again static representations.

Lastly, regarding research question RQ4, we can highlight that fine-tuning Transformer-based models using tweets from sentiment datasets seems to boost classification performance in Twitter sentiment analysis. As a matter of fact, the strategies AllData and LOO exploited in this section, which use a collection of sentiment tweets to adjust a language model, achieved better overall results than using samples of unlabeled, or generic unlabeled, tweets. Although we do not use the labels of those tweets in the fine-tuning procedure, they may carry a lot of sentiment information as compared to the tweets from the Edinburgh corpus, which originated the samples of generic unlabeled tweets used in the experiments. Furthermore, BERTweet embeddings fine-tuned with strategy AllData seems to be very effective in determining the sentiment expressed in tweets, especially when used to train LR, MLP, and SVM classifiers.

%Regarding the RQ4, with all the above analysis, we can conclude fine-tuning Transformer-based autoencoder models with tweets from sentiment datasets available in the literature brings benefits to the sentiment classification task in tweets. This fine-tuning overcome the previews fine-tuning performance. Concerning the most appropriate model, BERTweet-AllData appears as the more effective language model in tweets' sentiment classification task. We were also able to realize BERTweet-AllData's good performance occurs when combined with MLP and LR. Considering the entire classification system (Embeddings + classifier), our suggestion is to use the BERTweet language, tuned with a collection of tweets from sentiment datasets, and with LR or MLP classifier.

\begin{table}
\caption{Top 10 average rank position results achieved for each combination Model-Classifier by evaluating all assessed model in this study}
\label{tab:top10_ranksum_all-models-class}
\scalebox{0.8}{
\begin{tabular}{lll|lll}
\hline\noalign{\smallskip}
\multirow{2}{*}{\textbf{Model}} & \multirow{2}{*}{\textbf{Classifier}} & \textbf{Accuracy} & \multirow{2}{*}{\textbf{Model}} & \multirow{2}{*}{\textbf{Classifier}} & \textbf{$\bm{F_1}$-macro} \\
& & \textbf{avg. rank pos.} & & & \textbf{avg. rank pos.}\\
\noalign{\smallskip}\hline\noalign{\smallskip}
\rule{0pt}{3ex}
BERTweet-LOO&LR&17.30&BERTweet-LOO&LR&15.52\\
\rule{0pt}{3.0ex}
BERTweet-AllData&LR&18.16&BERTweet-AllData&LR&16.30\\
\rule{0pt}{3.0ex}
BERTweet-LOO&MLP&18.55&BERTweet-5K&LR&18.64\\
\rule{0pt}{3.0ex}
BERTweet-AllData&MLP&19.36&BERTweet-AllData&MLP&20.68\\
\rule{0pt}{3.0ex}
BERTweet-5K&LR&20.64&BERTweet-LOO&MLP&21.00\\
\rule{0pt}{3.0ex}
BERTweet-5K&MLP&23.14&BERTweet-25K&SVM&22.48\\
\rule{0pt}{3.0ex}
BERTweet-25K&MLP&24.30&BERTweet-10K&SVM&23.16\\
\rule{0pt}{3.0ex}
BERTweet-25K&LR&26.23&BERTweet-25K&LR&23.30\\
\rule{0pt}{3.0ex}
BERTweet-1K&MLP&26.91&BERTweet-10K&LR&23.86\\
\rule{0pt}{3.0ex}
BERTweet-50K&MLP&27.02&BERTweet-AllData&SVM&24.82\\
\noalign{\smallskip}
\hline
\end{tabular}
}
\end{table}

\begin{table}
\caption{Tail 10 average rank position results achieved for each combination Model-Classifier by evaluating all assessed model in this study}
\label{tab:tail10_ranksum_all-models-class}
\scalebox{0.83}{
\begin{tabular}{lll|lll}
\hline\noalign{\smallskip}
\multirow{2}{*}{\textbf{Model}} & \multirow{2}{*}{\textbf{Classifier}} & \textbf{Accuracy} & \multirow{2}{*}{\textbf{Model}} & \multirow{2}{*}{\textbf{Classifier}} & \textbf{$\bm{F_1}$-macro} \\
& & \textbf{avg. rank pos.} & & & \textbf{avg. rank pos.}\\
\noalign{\smallskip}\hline\noalign{\smallskip}
\rule{0pt}{3ex}
EWE&RF&247.23&DeepMoji&RF&252.86\\
\rule{0pt}{3.0ex}
W2V-Araque&RF&249.75&BERTweet-static&LR&253.14\\
\rule{0pt}{3.0ex}
W2V-GN&RF&250.00&EWE&RF&256.48\\
\rule{0pt}{3.0ex}
GloVe-WP&RF&253.68&W2V-Araque&RF&259.80\\
\rule{0pt}{3.0ex}
fastText&RF&255.75&W2V-GN&RF&261.70\\
\rule{0pt}{3.0ex}
BERT-static&RF&257.32&GloVe-WP&RF&263.11\\
\rule{0pt}{3.0ex}
RoBERTa-static&RF&259.70&fastText&RF&266.95\\
\rule{0pt}{3.0ex}
BERT-static&LR&263.34&BERT-static&RF&267.75\\
\rule{0pt}{3.0ex}
BERTweet-static&RF&265.91&RoBERTa-static&RF&269.93\\
\rule{0pt}{3.0ex}
BERTweet-static&LR&274.43&BERTweet-static&RF&275.43\\
\noalign{\smallskip}
\hline
\end{tabular}
}
\end{table}

\begin{table}%[!htbp]
\caption{Top 10 average rank position results achieved comparing all assessed models in this study}
\label{tab:top-emb_ranksum-all-models}
\scalebox{1}{
\begin{tabular}{ll|ll}
\hline\noalign{\smallskip}
\multirow{2}{*}{\textbf{Model}}& \textbf{Accuracy} & \multirow{2}{*}{\textbf{Model}}& \textbf{$\bm{F_1}$-macro} \\
& \textbf{avg. rank pos.} & & \textbf{avg. rank pos.}\\
\noalign{\smallskip}\hline\noalign{\smallskip}
\rule{0pt}{3ex}
BERTweet-AllData&53.83&BERTweet-AllData&60.67\\
\rule{0pt}{3.0ex}
BERTweet-LOO&56.52&BERTweet-LOO&62.59\\
\rule{0pt}{3.0ex}
BERTweet-5K&60.56&BERTweet-5K&66.21\\
\rule{0pt}{3.0ex}
BERTweet-25K&65.38&BERTweet-25K&71.07\\
\rule{0pt}{3.0ex}
BERTweet-10K&65.51&BERTweet-10K&71.17\\
\rule{0pt}{3.0ex}
BERTweet-1K&68.59&BERTweet-1K&73.32\\
\rule{0pt}{3.0ex}
BERTweet-50K&72.31&BERTweet-50K&77.50\\
\rule{0pt}{3.0ex}
BERTweet-250K&78.13&BERTweet-250K&82.90\\
\rule{0pt}{3.0ex}
BERTweet-InData&83.80&BERTweet-InData&87.93\\
\rule{0pt}{3.0ex}
BERTweet-500K&86.10&BERTweet-500K&90.85\\
\noalign{\smallskip}
\hline
\end{tabular}
}
\end{table}

\begin{table}%[!htbp]
\caption{Tail 10 average rank position results achieved comparing all assessed models in this study}
\label{tab:tail-emb_ranksum-all-models}
\scalebox{1}{
\begin{tabular}{ll|ll}
\hline\noalign{\smallskip}
\multirow{2}{*}{\textbf{Model}}& \textbf{Accuracy} & \multirow{2}{*}{\textbf{Model}}& \textbf{$\bm{F_1}$-macro} \\
& \textbf{avg. rank pos.} & & \textbf{avg. rank pos.}\\
\noalign{\smallskip}\hline\noalign{\smallskip}
\rule{0pt}{3ex}
SSWE&209.69&W2V-GN&204.38\\
\rule{0pt}{3.0ex}
GloVe-TWT&215.62&GloVe-TWT&207.22\\
\rule{0pt}{3.0ex}
DeepMoji&217.13&DeepMoji&208.08\\
\rule{0pt}{3.0ex}
EWE&217.46&EWE&208.43\\
\rule{0pt}{3.0ex}
TF-IDF&220.83&GloVe-WP&215.45\\
\rule{0pt}{3.0ex}
BERT-static&224.61&fastText&218.05\\
\rule{0pt}{3.0ex}
GloVe-WP&225.94&BERT-static&218.40\\
\rule{0pt}{3.0ex}
fastText&227.80&w2v-Araque&222.85\\
\rule{0pt}{3.0ex}
W2V-Araque&230.56&TF-IDF&224.34\\
\rule{0pt}{3.0ex}
BERTweet-static&244.21&BERTweet-static&237.01\\
\noalign{\smallskip}
\hline
\end{tabular}
}
\end{table}

%%%%%%%%%%%%%%%%%%%%%%%% Seção 8 %%%%%%%%%%%%%%%%%%%

\section{Conclusions and future works}
\label{sec:conclusions_future}

In this article, we presented an extensive assessment of modern and classical word representations when used for the task of Twitter sentiment analysis. Specifically, we assessed the classification performance of 14 static representations, the most recent Transformer-based autoencoder models, including BERT, RoBERTa, and BERTweet, as well as different fine-tuning strategies of the language representation tasks in such models. All models were evaluated in the context of Twitter sentiment analysis using a rich set of 22 datasets and five classifiers from distinct natures. The main focus of this study was on identifying the most appropriate word representations for the sentiment analysis of English tweets.

Based on the results of the experiments performed in this study, we can highlight the following conclusions:

\begin{itemize}
    \item Considering the static representations in limited resource scenario, we could note that Emo2Vec, w2v-Edin, and RoBERTa models seem to be well-suited representations for determining the sentiment expressed in tweets. The good performance achieved by Emo2Vec and w2v-Edin indicates that being trained from scratch with tweets can boost the classification performance of static representations when applied in Twitter sentiment analysis. Although RoBERTa was not trained from stratch with tweets, it is a Transformer-based autoencoder model, which holds state-of-the-art performance in several NLP tasks. Regarding the classifiers, we could see that SVM and MLP achieved the best overall performances, especially when used to train RoBERTa's static embeddings.
    
    \item Regarding the Transformer-based models, we could observe that BERTweet is the most appropriate language model to be used in the sentiment classification of tweets. Specifically, the particular vocabulary tweets contain, combined with a language model that was trained focused on learning their intrinsic structure, can effectively improve the performance of the Twitter sentiment analysis task. Considering the combination of language models and classifiers, we can point out that BERTweet achieved the best overall results when combined with LR and MLP. Furthermore, by comparing the Transformer-based models and the static representations, we could notice that the adaptation of the tokens' embeddings to the context they appear performed by the Transformer-based models benefits the sentiment classification task.
    
    \item When fine-tuning the Transformer-based models with a large set of English unlabeled tweets we could note that although it improves the classification performance, using as many tweets as possible does not necessarily means better results. In this context, we presented an extensive evaluation of sets of tweets with different sizes, varying from 0.5K to 1.5M. These results have shown that while BERT and RoBERTa achieved better predictive performances when tuned with sets of 250K and 50K tweets, respectively, BERTweet outperformed all fine-tuned models using only 5K tweets. This result indicates that models trained from scratch with tweets, such as BERTweet, needs less tweets to have its performance improved. Moreover, by comparing all fine-tuned models taking into account the classifiers, BERTweet combined with MLP, LR, and SVM achieved the best overall performances.
    
    \item Analyzing the fine-tuning of the language model based on Transformers autoencoders with sentiment analysis datasets, i.e., with tweets that express polarity, we can see that the tuned models' performance is better than when tuned with generic tweets. All fine-tuning strategies with sentiment analysis datasets performed better than the best-tuned models adjusted with generic tweets. We conclude then that it is worth fine-tuning a model based on Transformer autoencoders using a set of sentiment tweets. Among the fine-tuning strategies -- using sentiment analysis tweets -- explored in the study, it was possible to perceive that each Transformer model presented a better performance with different adjustment methods. The use of only the target dataset, for example, was a good option to be used with BERT. For RoBERTa and BERTweet, the combination of the target dataset with a set of tweets from other datasets presented a good strategy for fine-tuning the language model. In a general comparison, we noticed that BERTweet tuned with the union of the target dataset and the set of sentiment analysis tweets (BERTweet\_22Dt) performed better than the other adjusted models. Besides, we could observe that BERTweet\_22Dt presented a good performance when combined with LR and MLP classifiers.
    
    \item  After answering our research questions, we can briefly state that: (i) Transformer-based autoencoder models perform better than static representation, (ii) Transformer autoencoder models fine-tuned with English tweets behavior better than the respective original models and, finally, (iii) it is worth fine-tuning a language model originally trained with generic English tweets with tweets from sentiment analysis datasets. Considering all original and fine-tuned models, the best overall performance for the English tweets sentiment analysis task was achieved by the Transformer-Autoencoder model trained from scratch with generic tweets (BERTweet) when fine-tuned with tweets from a target sentiment dataset added by tweets from a large set of other sentiment datasets. This strategy was called BERTweet\_22Dt, which we consider a good suggestion for sentiment classification of English tweets, mainly when combined with MLP or LR classifiers.

\end{itemize}

For future work, we plan to investigate other methods for fine-tuning language models, mainly considering the polarity classification as the downstream tuning task. Transformer-Autoencoder pre-trained models, like BERT, RoBERTa and BERTweet, can have its weights adjusted looking for becoming more accurate in a specific task, like sentiment analysis. This adjustment is made by adding an extra classification layer in the top of the model and back-propagating the error in the final task through language models' weights. We intend then to compare the best results obtained in this study with the ones achieved by this specific-task category of fine-tuning.

%\begin{acknowledgements}
%\end{acknowledgements}

\section*{Acknowledgments}
The authors would like to thank the Brazilian Research agencies FAPERJ and CNPq for the financial support. 

\section*{Conflict of interest}

The authors declare that they have no conflict of interest.

\bibliographystyle{spmpsci}      % mathematics and physical sciences
\bibliography{ref.bib}   % name your BibTeX data base

%\begin{thebibliography}{}
% etc
%\end{thebibliography}

\end{document}